%% file: paper.tex
\pdfoutput=1

\documentclass[11pt]{article}

\usepackage[final]{acl}

\usepackage{times}
\usepackage{latexsym}

\usepackage[T1]{fontenc}

\usepackage[utf8]{inputenc}

\usepackage{microtype}

\usepackage{inconsolata}

\usepackage{graphicx}
\usepackage{caption}
\usepackage{subcaption}
\usepackage{tcolorbox}
\usepackage{booktabs}
\usepackage{multirow}
\usepackage{amsfonts}
\usepackage{amsmath}
\usepackage{xcolor}

\newtcolorbox{highlight}{
  colback=lightgray!20, 
  colframe=gray!60, 
  boxrule=1pt, 
  arc=0pt, 
  boxsep=0pt, 
  left=6pt, 
  right=6pt, 
  top=6pt, 
  bottom=6pt, 
}

\newtcolorbox{highlightgreen}{
  colback=green!20, 
  colframe=green!60, 
  boxrule=1pt, 
  arc=0pt, 
  boxsep=0pt, 
  left=6pt, 
  right=6pt, 
  top=6pt, 
  bottom=6pt, 
}

\newcommand{\mete}[1]{{#1}}

\title{Evaluating Morphological Compositional Generalization \\in Large Language Models}

\author{
 \textbf{Mete Ismayilzada\textsuperscript{1,2}},
 \textbf{Defne Circi\textsuperscript{*3}},
 \textbf{Jonne Sälevä\textsuperscript{*4}},
 \textbf{Hale Sirin\textsuperscript{7}},
\\
 \textbf{Abdullatif Köksal\textsuperscript{5,6}},
 \textbf{Bhuwan Dhingra\textsuperscript{3}},
 \textbf{Antoine Bosselut\textsuperscript{1}},
 \textbf{Duygu Ataman\textsuperscript{\textdagger9}},
\\
 \textbf{Lonneke van der Plas \textsuperscript{\textdagger2,8}}
\\
\\
 \textsuperscript{1}EPFL,
 \textsuperscript{2}Idiap Research Institute,
 \textsuperscript{3}Duke University,
 \textsuperscript{4}Brandeis University,
 \\
 \textsuperscript{5}LMU Munich,
 \textsuperscript{6}University of Cambridge,
 \textsuperscript{7}Johns Hopkins University,
 \\
 \textsuperscript{8}Università della Svizzera Italiana,
 \textsuperscript{9}New York University
\\
 \small{
   \href{mailto:email@domain}{mahammad.ismayilzada@epfl.ch}
 }
}

\begin{document}
\maketitle

\begingroup
\renewcommand\thefootnote{\textsuperscript{*}}\footnotetext{Equal contribution}
\endgroup

\begingroup
\renewcommand\thefootnote{\textsuperscript{\textdagger}}\footnotetext{Equal supervision}
\endgroup

\input{00_abstract}
\input{01_introduction}
\input{02_related_work}
\input{03_methodology}
\input{05_experiments}
\input{06_analysis}

\input{07_conclusion}

\input{08_limitations}
\input{10_acknowledgements}

\bibliography{anthology, literature}

\newpage
\appendix

\input{11_appendix}

\end{document}

%% file: 00_abstract.tex
\begin{abstract}
Large language models (LLMs) have demonstrated significant progress in various natural language generation and understanding tasks. However, their linguistic generalization capabilities remain questionable, raising doubts about whether these models learn language similarly to humans. While humans exhibit compositional generalization and linguistic creativity in language use, the extent to which LLMs replicate these abilities, particularly in morphology, is under-explored. In this work, we systematically investigate the morphological generalization abilities of LLMs through the lens of compositionality. We define morphemes as compositional primitives and design a novel suite of generative and discriminative tasks to assess morphological productivity and systematicity. Focusing on agglutinative languages such as Turkish and Finnish, we evaluate several state-of-the-art instruction-finetuned multilingual models, including GPT-4 and Gemini. Our analysis shows that LLMs struggle with morphological compositional generalization particularly when applied to novel word roots, with performance declining sharply as morphological complexity increases. While models can identify individual morphological combinations better than chance, their performance lacks systematicity, leading to significant accuracy gaps compared to humans.
\end{abstract}

%% file: 01_introduction.tex
\section{Introduction}
\label{sec:introduction}

Large language models (LLMs) have recently achieved remarkable advances in the broad domain of natural language generation and understanding tasks \cite{geminiteam2024gemini, zhao2023survey, bubeck2023sparks, wei2022emergent, brown2020language}. However, these models have also been shown to lack strong linguistic generalization capabilities \cite{weissweiler-etal-2023-counting, mccoy-etal-2023-much, goldman-etal-2022-un, wilson-etal-2023-abstract, Linzen2020HowCW, Baroni2019LinguisticGA}. This discrepancy casts doubt on whether language models learn a language the same way as humans do. When learning a language which is essentially a finite set of words and rules, humans exhibit linguistic creativity \cite{chomsky1965aspects, bergs2019linguistic} and compositional generalization \mete{through \textit{productivity} and \textit{systematicity}} \cite{FODOR19883, chomsky1957syntactic}. \mete{These abilities allow humans respectively to produce and understand novel combinations of familiar grammar units.} While compositional generalization abilities of language models have been extensively studied \cite{lake2018generalization, Keysers2019MeasuringCG, kim-linzen-2020-cogs}, the extent to which language models employ this ability in morphology however remains largely under-explored. Recent works evaluating morphological generalization in language models have only focused on the productivity aspect with a limited coverage of inflectional forms \cite{weissweiler-etal-2023-counting, anh-etal-2024-morphology}.

\begin{figure}[t]
  \includegraphics[width=\columnwidth,trim={0 0cm 6cm 0},clip]{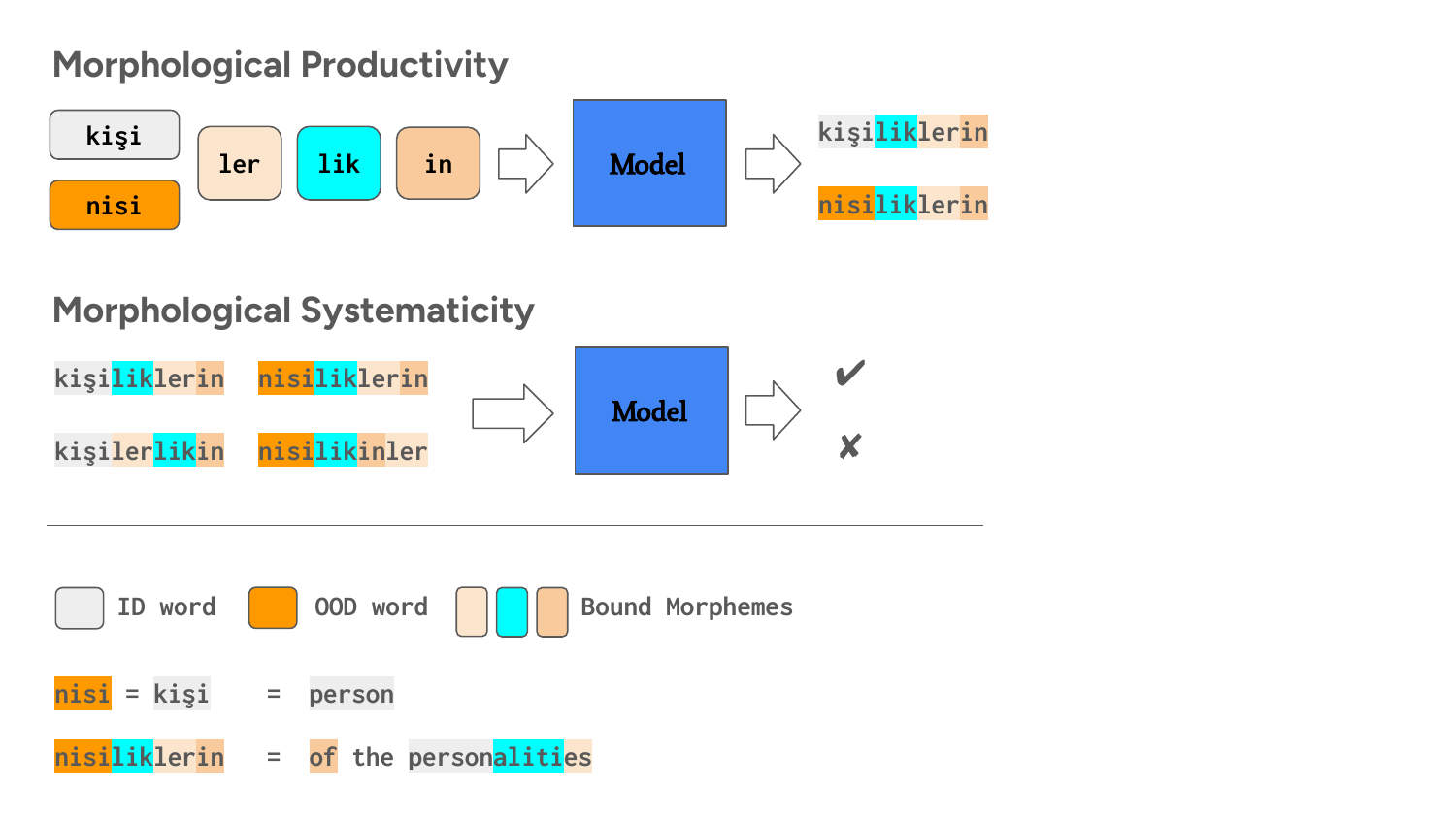}
  \caption{\textbf{Our morphological generalization tasks illustrated with an example in Turkish.} ID and OOD refer to in-distribution and out-of-distribution respectively. English translations are not part of the task and only shown here for illustrative purposes.}
  \label{fig:introduction}
\end{figure}

In this work, we address this gap by systematically investigating the morphological generalization abilities of LLMs through the lens of compositionality. Following \citet{Keysers2019MeasuringCG}, we define the morphemes (smallest meaningful units in a language)\footnote{In linguistics, a distinction is made between \textit{free} morphemes that can stand alone such as words "cat" and "come" and \textit{bound} morphemes that can only appear as part of a larger expression (e.g. "cats", "coming") such as affixes "s" and "ing".} as the compositional primitives and design a novel suite of generative and discriminative language tasks based on the morphological combinations of these primitives. These tasks aim to test \textbf{morphological productivity} (ability to produce novel well-formed combinations of morphemes) and \textbf{systematicity} (ability to systematically understand novel combinations) respectively. Figure \ref{fig:introduction} illustrates an example of both tasks.

We evaluate several state-of-the-art instruction-finetuned large multilingual models on these tasks: GPT-4, Gemini-1.5, Aya-23 and Qwen-2.5.
To ensure our findings are not language-specific, we experiment with two morphologically rich (i.e. characterized by a large number of inflectional and derivational forms) languages: Turkish and Finnish. Both languages share typological features (e.g. agglutination) despite being unrelated.

We find that LLMs lack human-like morphological compositional generalization ability in agglutinative languages despite their high performance on various tasks in these languages. Our analysis shows that morphological productivity, especially when applied to novel word roots is highly challenging for LLMs. Moreover, as the morphological complexity of words increases the model performance sharply decreases (to nearly zero) while human performance is not consistently affected. On the systematicity task, while models perform much better than chance in identifying the validity of \textit{individual} morphological combinations, however, this behaviour is not robust or systematic i.e. models fail to \textit{consistently} determine validity of several compositions made up of the same set of morphemes.

In summary, our contributions are as follows: \textbf{1)} We design novel morphological generalization tasks that require compositional processing. \textbf{2)} We prepare specific test suites in both Turkish and Finnish to measure morphological generalization and make these available for future research\footnote{\url{https://github.com/mismayil/morph-gen}}. \textbf{3)} Using our novel tasks and test suites, we conduct a systematic analysis of morphological compositional generalization abilities of LLMs. \textbf{4)} Our findings reveal a systematic gap in LLM's ability compared to humans concerning morphological generalization \mete{in agglutinative languages} that also requires compositionality.


%% file: 02_related_work.tex
\section{Related Work}
\label{sec:related_work}
\subsection{Compositional Generalization}
Compositional generalization is the capacity to understand and produce novel compositions of seen primitives and is typically characterized by systematicity and productivity \cite{FODOR19883, Keysers2019MeasuringCG}. Systematicity refers to the ability to understand different combinations that are made up of the same known components such as \textit{John loves Mary} and \textit{Mary loves John}. Productivity, on the other hand, is the ability to produce potentially infinite novel combinations of a finite number of known building blocks such as using conjunctions to construct sentences \textit{Mary knows that John loves Mary} and \textit{John heard that Mary knows that John loves Mary}. Past work has developed several benchmarks to measure compositional generalization abilities of neural models both in fine-tuning and in-context learning settings and has shown this task to be highly challenging \cite{yang2024exploring, lake2018generalization, Keysers2019MeasuringCG, kim-linzen-2020-cogs, an-etal-2023-context, Dziri2023FaithAF}. These benchmarks have mainly focused on synthetic sequence matching, semantic parsing, question-answering and problem-solving tasks. Our work however, investigates compositional generalization in the context of morphology.

\subsection{Morphological Generalization}
Morphological generalization is the ability to understand words based on their constituent parts known as \textit{morphemes} and combine them to derive new words \cite{wysocki1987morph}. Morphemes are the smallest meaningful units of language that typically correspond to word roots and affixes (i.e. prefixes, infixes and suffixes). Composing these units to construct new words can be done through \textit{inflection} and \textit{derivation} tasks in morphology where derivation \mete{often} changes the syntactic category of the words and inflection does not. These tasks have gained considerable attention as part of the SIGMORPHON's shared tasks \cite{Cotterell2016TheS2, cotterell-etal-2018-conll, vylomova-etal-2020-sigmorphon, Kodner2022SIGMORPHONUniMorph2S, Goldman2023SIGMORPHONUniMorph2S} and efforts to create a universal morphology \cite{McCarthy2020UniMorph3U}. While transformer-based models have been shown to achieve near-perfect accuracy on these tasks \cite{Canby2020UniversityOI}, recent work has also found that these results are inflated due to lemma overlap pointing to a lack of generalization \cite{goldman-etal-2022-un}. Other works have recently investigated the morphological capabilities of LLMs using inflection tasks and reported similarly weak performance results \cite{anh-etal-2024-morphology, weissweiler-etal-2023-counting}. Similar to our study, both of these works use the popular Wug test \cite{berko1958} to evaluate the morphological generalization, however, they only focus on the productivity aspect, and their coverage of inflectional and derivational forms is limited. \mete{For example, \citet{weissweiler-etal-2023-counting} considers only a handful of specific inflectional forms (e.g. first person singular agreement and past tense, second person plural agreement etc.) for each language and \citet{anh-etal-2024-morphology} translates the original Wug test suite which is very small in size (23 samples) into different languages.} On the other hand, we cast the inflection and derivation tasks into the form of a compositional generalization task and evaluate models on both productivity and systematicity aspects. While focus of other works is breadth (languages from different families), we instead conduct an in-depth analysis of morphological generalization in typologically similar but unrelated languages with a large test suite covering a wide and diverse range of inflectional and derivational combinations.

%% file: 03_methodology.tex
\section{Methodology}
\label{sec:methodology}

\mete{
\subsection{Background}

The important role of compositional processing in language understanding and generation has been extensively studied \cite{carnap1947meaning, chomsky1965aspects, fodor1988connectionism, zadrozny1994compositional, Bauer_2001, aronoff2014}. Past works have shown that new word formation is often a multi-level process that requires identifying the correct order of primary and secondary morphemes \cite{kiparsky1982lexical, kiparsky1982word, hockett1954two}, and while humans might memorize some frequent words and phrases as a whole, most of the expressive language generation relies on productive rules of grammar \cite{o2015productivity}. However, not all languages are equally productive, and more productive languages (e.g. agglutinative) tend to have complex inflectional morphology \cite{cotterell2019complexity, ackerman2013}. Moreover, these languages have been shown to be harder to model for \textit{n}-gram and recurrent language models \cite{cotterell2018all, czarnowska-etal-2019-dont}. Inspired by these works, we focus our study on two highly agglutinative languages and compositional tasks which we describe in detail below.
}

\subsection{Tasks}

\mete{Similar to works studying compositional abilities of neural networks \cite{goodwin-etal-2020-probing, lake2018generalization, Keysers2019MeasuringCG},} we design two novel and simple compositional \mete{probing} tasks to test morphological abilities of models. First, a \textbf{morphological productivity task} which we define as a generative task where the model is given a word root, a list of affixes (not necessarily in the correct order) and is asked to derive a meaningful word by composing the root with the affixes in the correct order. Second, a \textbf{morphological systematicity task} which we define as a binary discriminative task where the model is again given a word root, a list of affixes and a word derived from the root using the given affixes (not necessarily a meaningful word) and is asked to determine the grammatical validity of the derived word. Figure \ref{fig:introduction} illustrates these tasks with an example in Turkish. 


Additionally, to measure the morphological generalization capabilities of LLMs, we take inspiration from Berko's Wug test \cite{berko1958} that is typically used to probe the inflectional and derivational morphological knowledge of children and design out-of-distribution (OOD) versions of our tasks using nonce word roots. More specifically, for each in-distribution (ID) word root in our test suite, we automatically generate a nonce word (i.e. word that does not exist in the given language) and use it in both tasks as the word root in place of the original one. However, since the model has never seen these words, to make sure the model understands the meaning of this new word, we provide the model with the original word root as a definition of the novel word root. Our generation of nonce words relies on the underlying morphophonological features and the frequency of each letter in a given language to make sure these words are plausible and inflected in the same way as the original root. Further details on nonce word generation can be found in Appendix \ref{sec:nonce-word-gen}.

\subsection{Data}
We focus our study on two highly agglutinative languages, Turkish and Finnish, and prepare test suites specific for our tasks in these languages. We particularly choose these languages because they are characterized by a large number of morphemes and hence require a high degree of compositional generalization ability.

\paragraph{Turkish}
Turkic languages are well-known to be highly agglutinative where the word is composed of several morphemes in addition to a root. We select Turkish as a representative of this language family in our study. To prepare our test suite we use the Bilkent Turkish Writings Dataset\footnote{\url{https://github.com/selimfirat/bilkent-turkish-writings-dataset}} as our base corpus which contains $6,844$ creative writings of Turkish 101 and Turkish 102 courses between 2014-2018 and hence, is full of morphologically complex words. Data statistics can be found in Appendix Table \ref{tab:tr-data-stats}. We preprocess this dataset to extract words and the sentences they are found in. Then we employ a morphological analyzer for Turkish \cite{ozturel2019} to segment these words into a root and surface-level morphemes. To create a diverse and balanced test suite, we sample $\approx 150$ examples per morpheme length $1$ to $7$ while maximizing the number of unique roots and morphemes (in total $1,049$ samples). Finally, we automatically generate a nonce word for each word in our test suite by relying on the fact that surface realizations of morphemes in Turkish are characterized by deterministic morphophonological processes such as vowel harmony, consonant assimilation and elision. Final data statistics and examples can be found in Appendix Tables \ref{tab:tr-final-data-stats} and \ref{tab:tr-final-data-examples} respectively. Further details on data collection can be found in Appendix \ref{sec:app-data}.

\paragraph{Finnish}

We first collect a $\sim$1,000,000 sentence subsample of the Finnish mC4 corpus \citep{xue-etal-2021-mt5}. 
We then extract unique words from the text and morphologically segment them using \texttt{omorfi} \citep{pirinen2015development} and UralicNLP \citep{Hamalainen2019}. After excluding words that analyzers did not cover, we manually annotate the segmentations to identify prefixes, lemmas, and affixes among the segments. We then perform stratified sampling based on the number of affixes to ensure an even range of morphological complexity in our data set. Finally, we extract sentences corresponding to each analyzed word from mC4 and validate whether they make sense. In a significant portion of cases, we notice that the raw sentences are noisy; in these cases, we opt to generate synthetic sentences using ChatGPT, which we \mete{(authors)} then \mete{manually} validate to be grammatical.
Final data statistics and examples can be found in Appendix Tables \ref{tab:fi-final-data-stats} and \ref{tab:fi-final-data-examples}.

%% file: 05_experiments.tex
\section{Experiments}
\label{sec:experiments}

\begin{figure*}[t]
\begin{subfigure}[b]{0.33\textwidth}
    \centering
    \includegraphics[width=\textwidth]{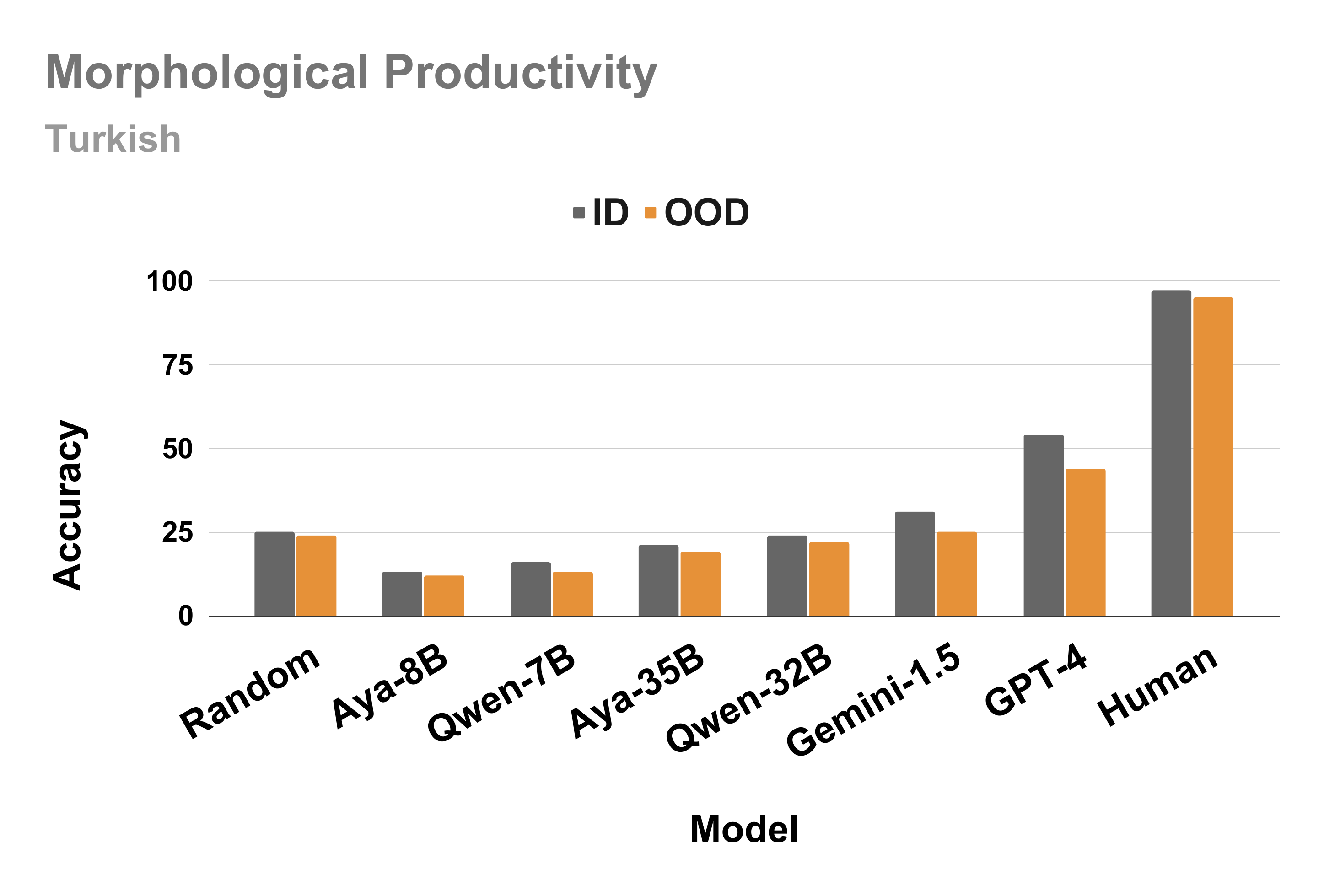}
\end{subfigure}
\begin{subfigure}[b]{0.33\textwidth}
    \centering
    \includegraphics[width=\textwidth]{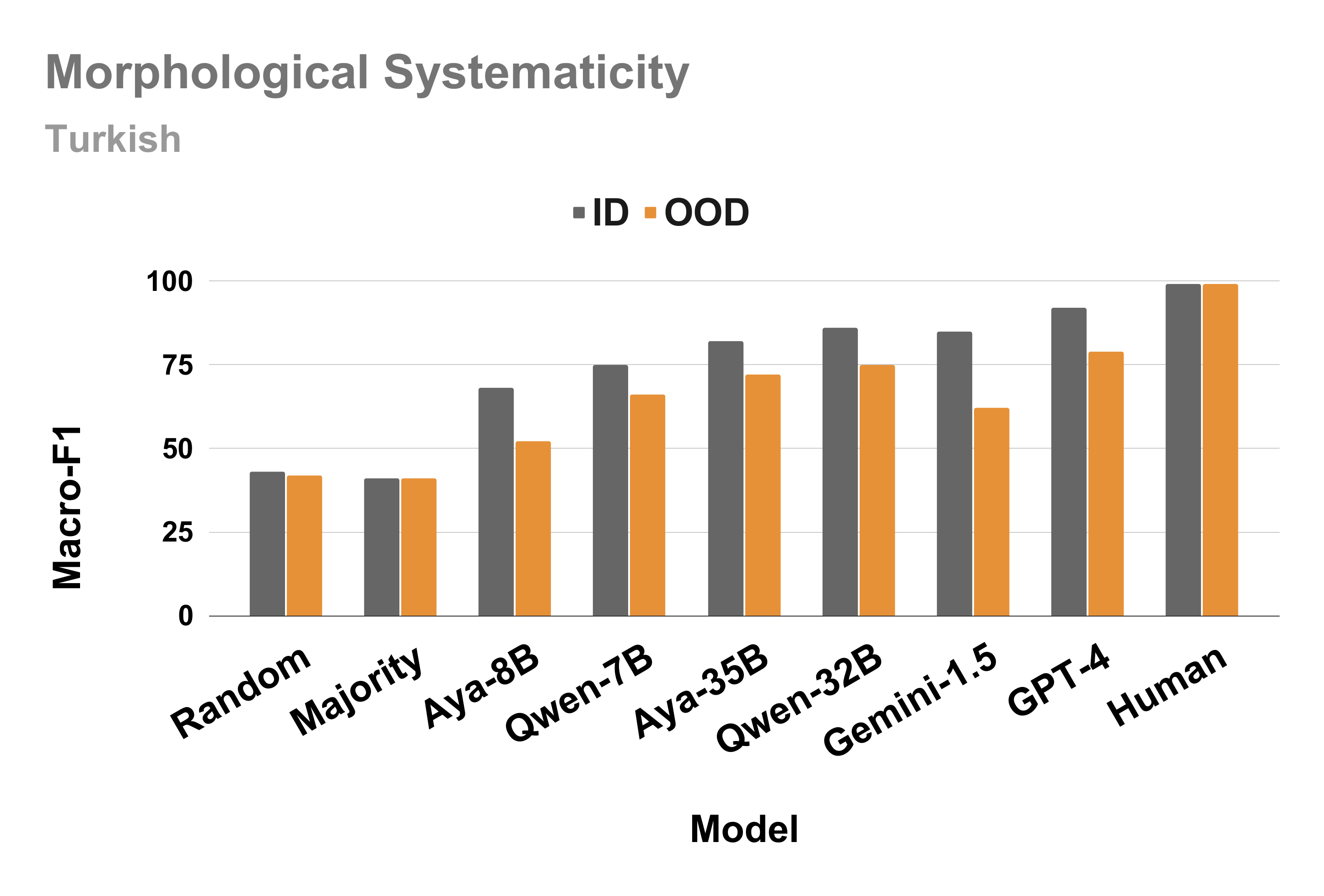}
\end{subfigure}
\begin{subfigure}[b]{0.33\textwidth}
    \centering
    \includegraphics[width=\textwidth]{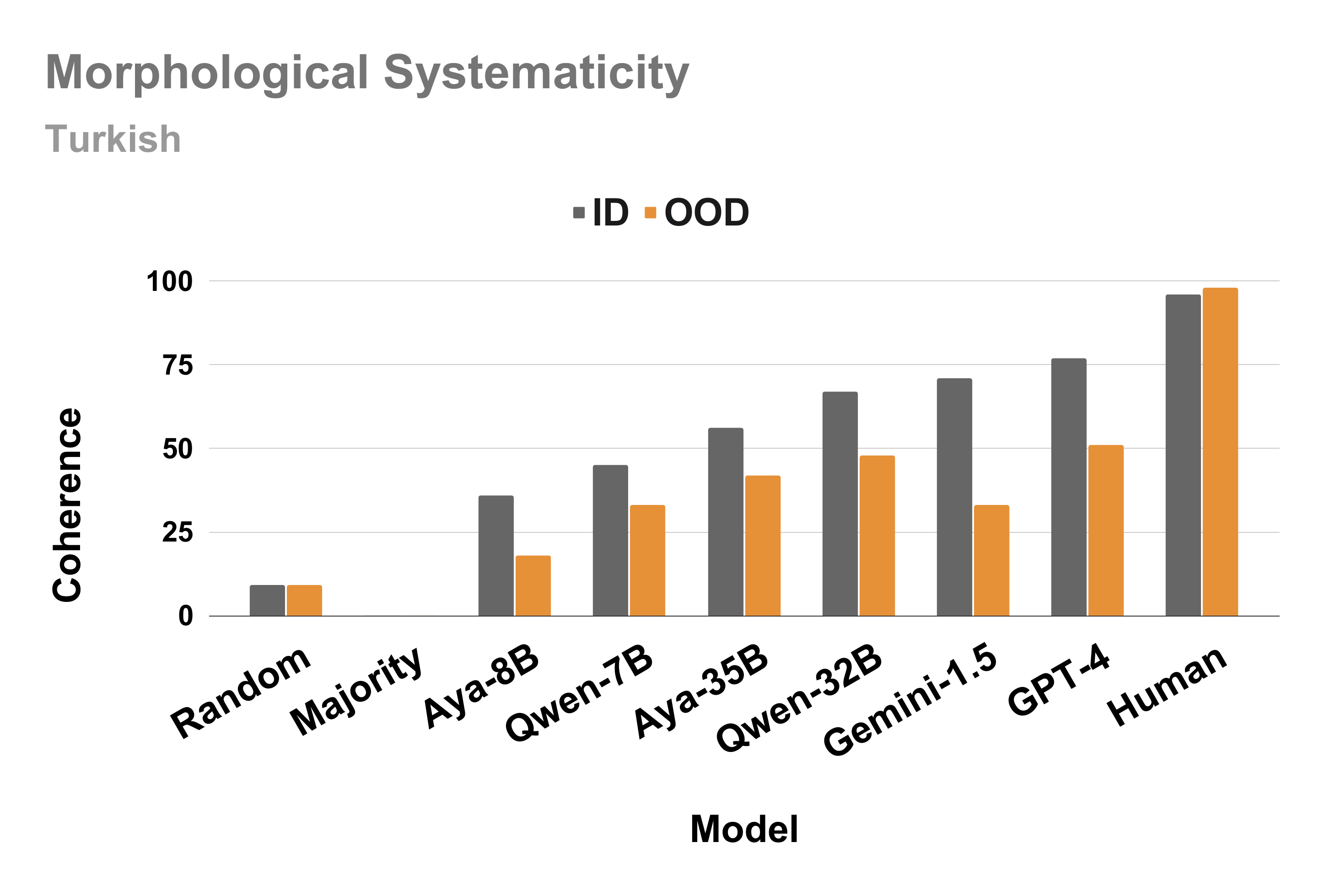}
\end{subfigure}
\caption{\textbf{Morphological productivity and systematicity task results for Turkish}. Detailed results for all shots are in Appendix Table \ref{tab:results-tr-en-default}.}
\label{fig:main-results-tr}
\end{figure*}

\begin{figure*}[t]
\begin{subfigure}[b]{0.33\textwidth}
    \centering
    \includegraphics[width=\textwidth]{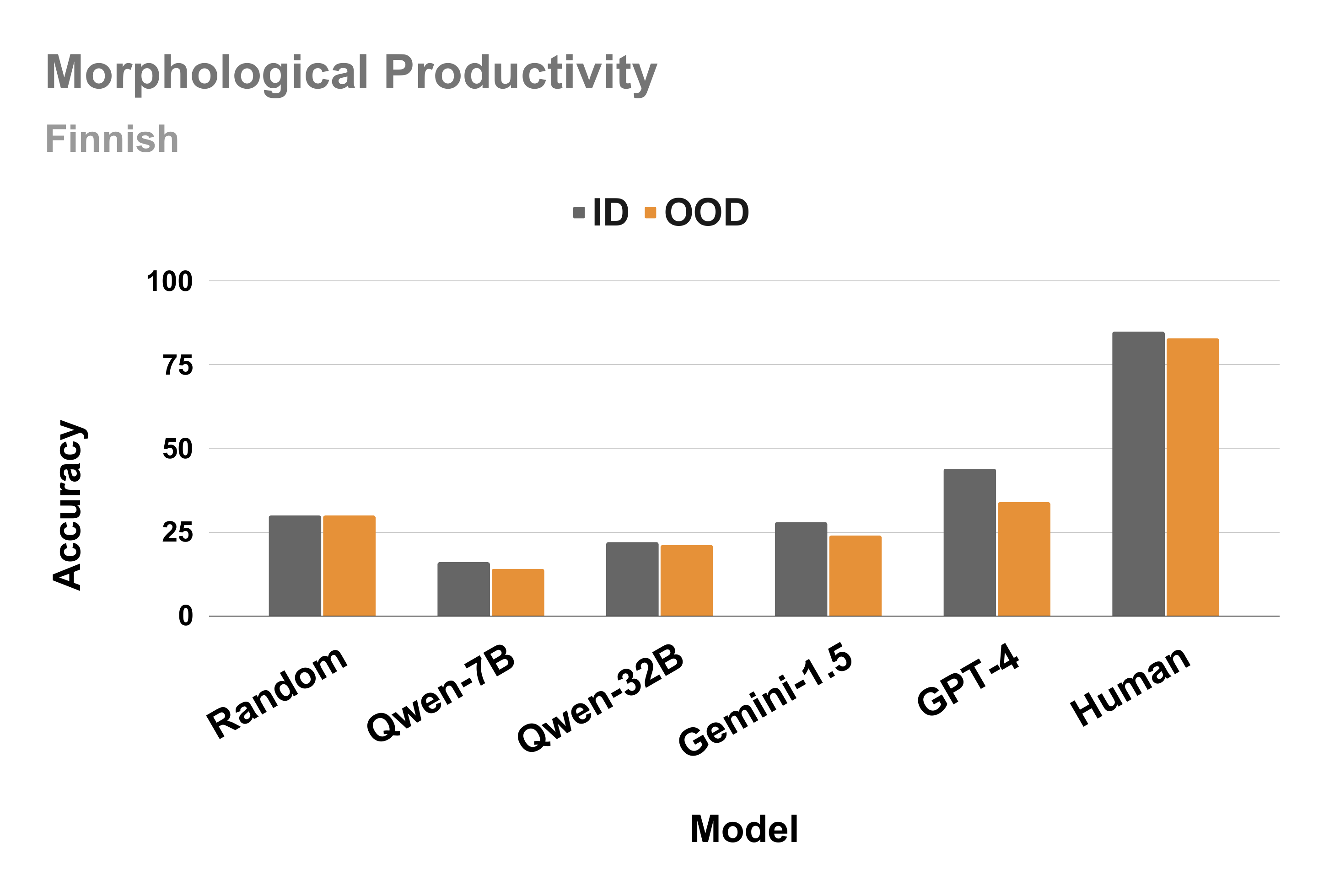}
\end{subfigure}
\begin{subfigure}[b]{0.33\textwidth}
    \centering
    \includegraphics[width=\textwidth]{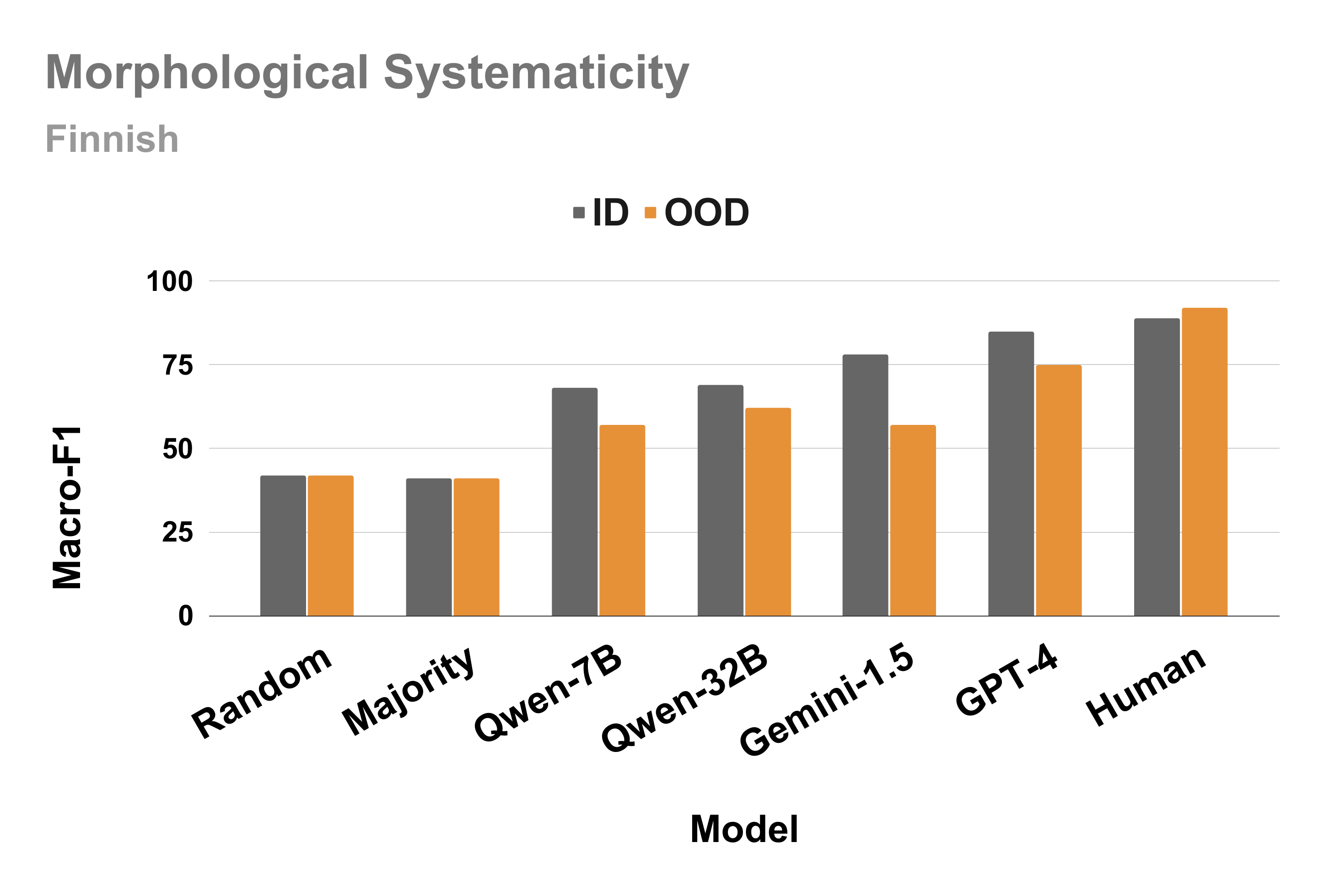}
\end{subfigure}
\begin{subfigure}[b]{0.33\textwidth}
    \centering
    \includegraphics[width=\textwidth]{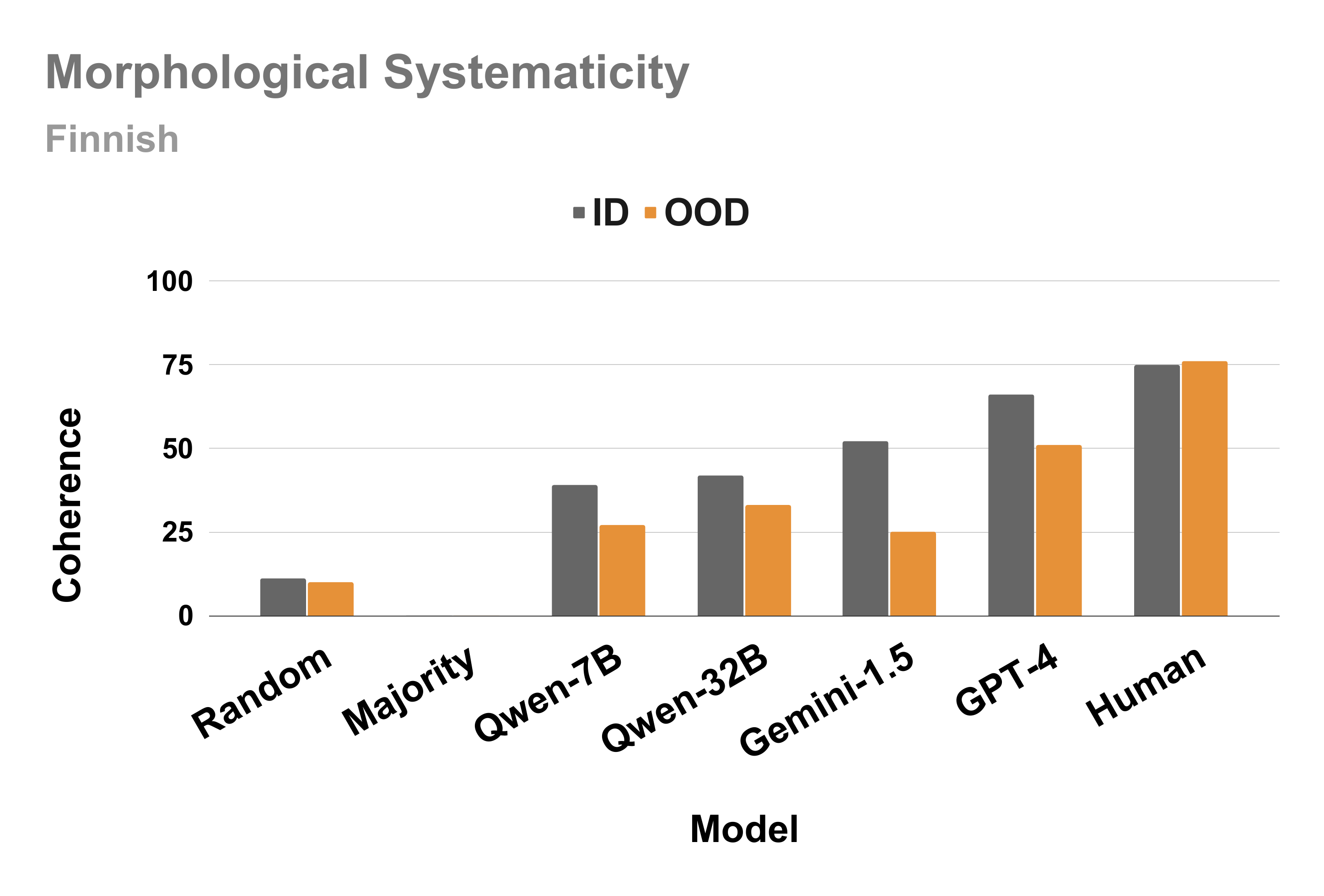}
\end{subfigure}
\caption{\textbf{Morphological productivity and systematicity task results for Finnish}. Detailed results for all shots are in Appendix Table \ref{tab:results-fi-en-default}.}
\label{fig:main-results-fi}
\end{figure*}

\paragraph{Setup}
We treat the productivity task as an open-ended task in which the model is asked to derive a word from the given root and affixes and the systematicity task as a binary classification task in which the model is asked to determine whether the given derivation is grammatically correct. For the systematicity task, we generate negative examples by producing all the combinations of morphemes attached to the same root and choosing the top four compositions (two for morpheme lengths of 1 and 2)\footnote{For 1-morpheme words, we manually annotate a negative morpheme to generate one negative option.}  that are closest to the original valid combination measured by the Levenshtein distance. We do this to ensure our incorrect combinations are challenging enough for the model as they will be deceptively close to a plausible derivation. We also experiment with other negative example selection strategies such as random selection and a heuristic selection based on the linguistic characteristics of the given language. We describe these settings in more detail and compare the results in Section \ref{sec:effect-neg-selection}. Finally, we \mete{(authors)} manually verify all the generated negative examples and fix the label of false negatives.


\paragraph{Models}
We evaluate several state-of-the-art multilingual instruction-finetuned LLMs, namely, two open-weights models, \textbf{Aya-23} \cite{aryabumi2024aya23openweight} and \textbf{Qwen-2.5} \cite{qwen2.5}, and two closed-source models, \textbf{Gemini-1.5} \cite{geminiteam2024gemini} and \textbf{GPT-4} \cite{openai2024gpt4technicalreport}. We evaluate all models on all languages except for Aya-23 which officially supports Turkish, but not Finnish.\footnote{We also experimented with recent LLMs that are instruction-finetuned specifically on Finnish such as Poro-34B \cite{luukkonen2024poro34bblessingmultilinguality} and Ahma model series that are Llama \cite{touvron2023llamaopenefficientfoundation} models fine-tuned on Finnish, however, we omitted them from our analysis as they failed to follow our task prompts in both English and Finnish templates.}. We also report the performance of a \textit{random} baseline that generates a derivation with a random combination of given morphemes (productivity task) and randomly decides whether the derivation is grammatically correct, and a \textit{majority} baseline, which selects the most frequent label (in our case \textit{"No"}) for the systematicity task (not applicable for the productivity task). All models are evaluated using few-shot (1, 3, and 5) in-context learning and greedy decoding since our tasks are deterministic by nature\footnote{\mete{We also experiment with other decoding strategies, however, find no significant difference in performance. Results for different decoding strategies can be found in Appendix \ref{sec:decoding-strategies}}}. Unless otherwise specified, prompt instructions are in English, and number of shots is set to $5$ for reported results\footnote{\mete{We also experiment with paraphrased version of our prompt instructions, but find no significant difference in performance. Results for paraphrased prompt instruction can be found in Appendix \ref{sec:prompt-instructions}}}. Further details on model evaluation can be found in Appendix \ref{sec:app-models}.

\paragraph{Evaluation Metrics}
For the productivity task, we use \textbf{Exact Match} accuracy against the correct derivations. For the systematicity task, we report an average of \textbf{Macro-F1} scores for each sample and a \textbf{Coherence} score that measures whether the model correctly and consistently identifies the validity (or invalidity) of all derivations for a given set of morphemes. \mete{Hence, coherence is defined as a binary score where the model gets a score of $1$ for a given sample if and only if it correctly guesses the validity of all derivations pertinent to that sample, otherwise $0$.} We employ this stringent metric to test the robustness of model performance similar to \cite{storks-chai-2021-beyond-tip}.

\paragraph{Human Evaluation}
We evaluate human performance on both tasks using two native speakers\footnote{Annotators were recruited from a Turkish and Finnish researcher community and were not compensated as they volunteered} per language, who annotate $70$ and $60$ samples from the Turkish and Finnish test suites, respectively. To ensure our evaluation sample is a representative sample of the entire test suite, we randomly select $10$ examples per morpheme length for each test distribution. Human annotators follow the same task instructions used for model prompts and were shown five examples. We report almost perfect or substantial inter-annotator agreement measured by Cohen's kappa score \cite{cohen1960} for both tasks, languages, and test distributions (Appendix Tables \ref{tab:tr-kappa}, \ref{tab:fi-kappa}). Finally, for each task metric, we report the average score of annotators as the final human score.

\paragraph{Results}
Figure \ref{fig:main-results-tr} and  \ref{fig:main-results-fi} summarize all model results for both morphological productivity and systematicity tasks evaluated respectively on the Turkish and Finnish data. We see that on the productivity task, all models except GPT-4 barely crack the random performance. While GPT-4 performs the best for both languages, it significantly lags behind the human performance ($-43\%$ and $-51\%$ in Turkish and $-40.8\%$ and $-48.9\%$ for Finnish respectively for ID and OOD data). Moreover, the GPT-4 performance gap between the ID and OOD test suites for both languages is much larger than the human gap ($\approx 10\%$ vs. $3\%$ in Turkish and $1.7\%$ in Finnish). These results indicate that humans are much more compositionally productive in morphology and generalize more robustly to novel unseen words. 

From the systematicity task results, we see that models perform much better than random and majority baselines with GPT-4 again in lead, however, the performance gap compared to humans is still significant, especially, on robustness as measured by coherence score ($-19.1\%$ and $-46.5\%$ in Turkish and $-8.8\%$ and $-25.2\%$ in Finnish respectively for ID and OOD data). The ID and OOD performance gap is also significant for all models, especially when measured by coherence score (ranging from $-9.5\%$ to $-23.3\%$ in Macro-F1 and from $-13.5\%$ to $-37.2\%$ in Coherence) while this gap is very low ($\approx 2\%$) for humans when measured by both metrics. These results show that humans are much more compositionally systematic and consistent in discriminating between correct and incorrect morphological forms made up of the same set of morphemes.

%% file: 06_analysis.tex
\section{Analysis}
\label{sec:analysis}

\begin{figure*}[h]
\begin{subfigure}[b]{0.33\textwidth}
    \centering
    \includegraphics[width=\textwidth]{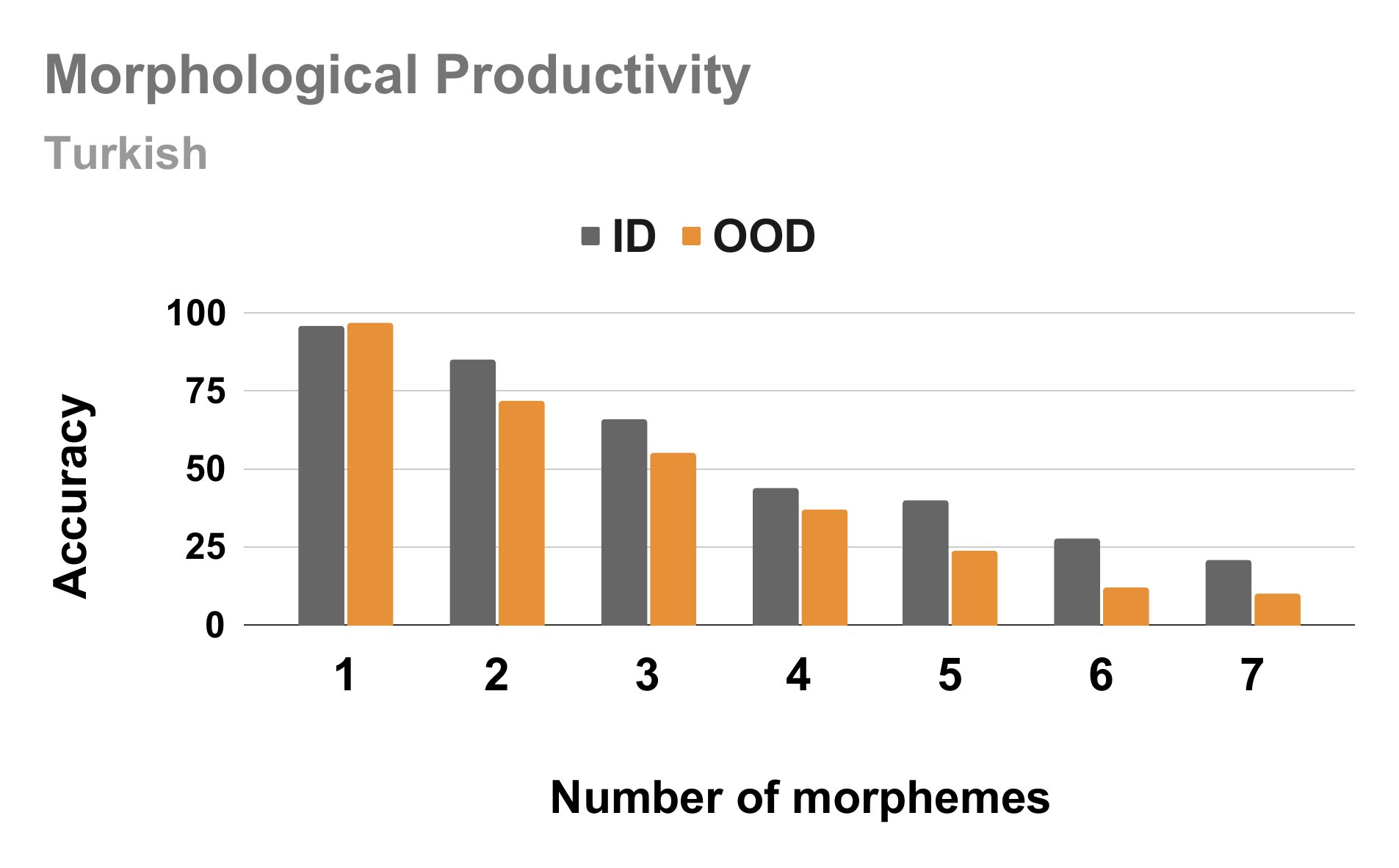}
\end{subfigure}
\begin{subfigure}[b]{0.33\textwidth}
    \centering
    \includegraphics[width=\textwidth]{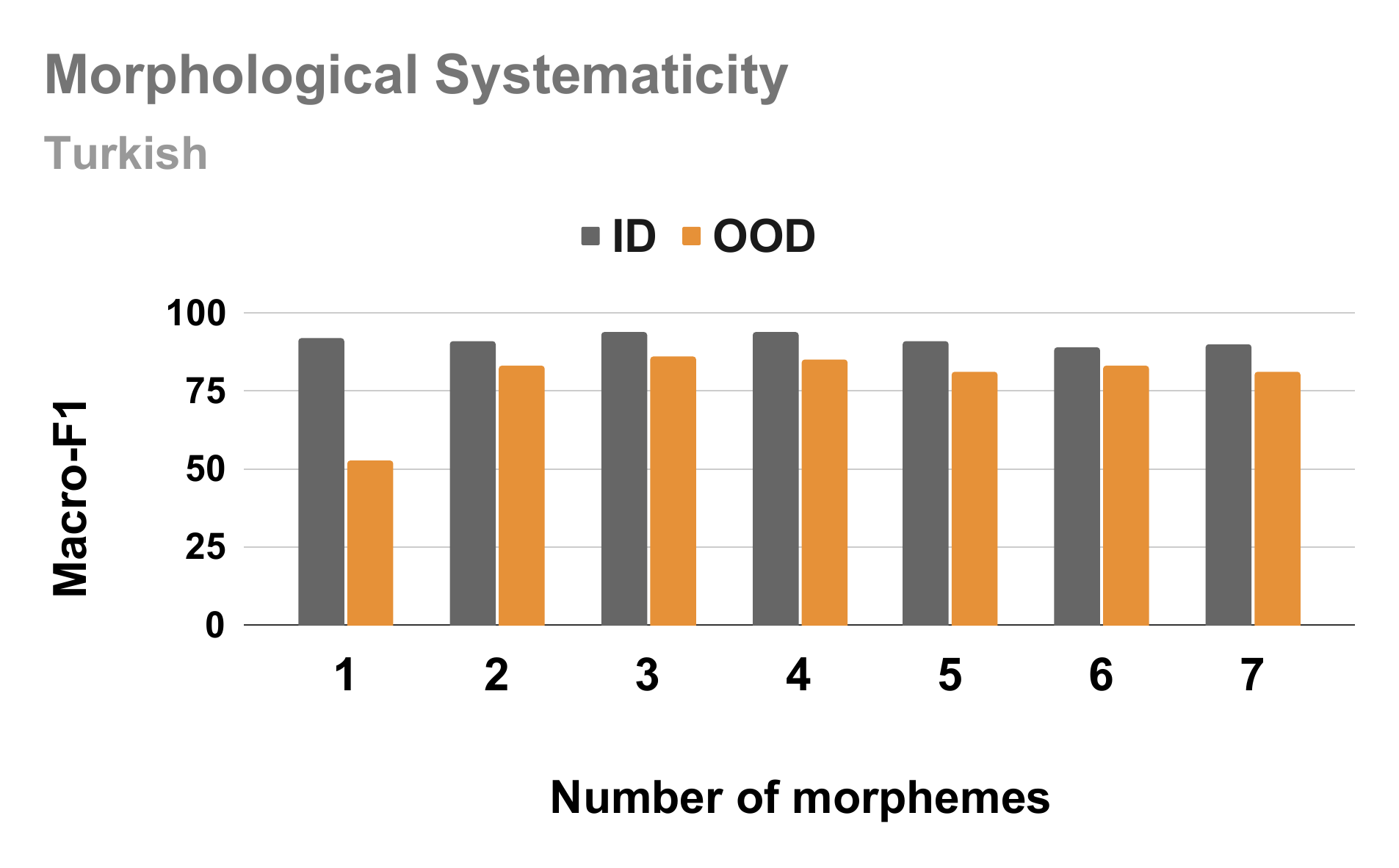}
\end{subfigure}
\begin{subfigure}[b]{0.33\textwidth}
    \centering
    \includegraphics[width=\textwidth]{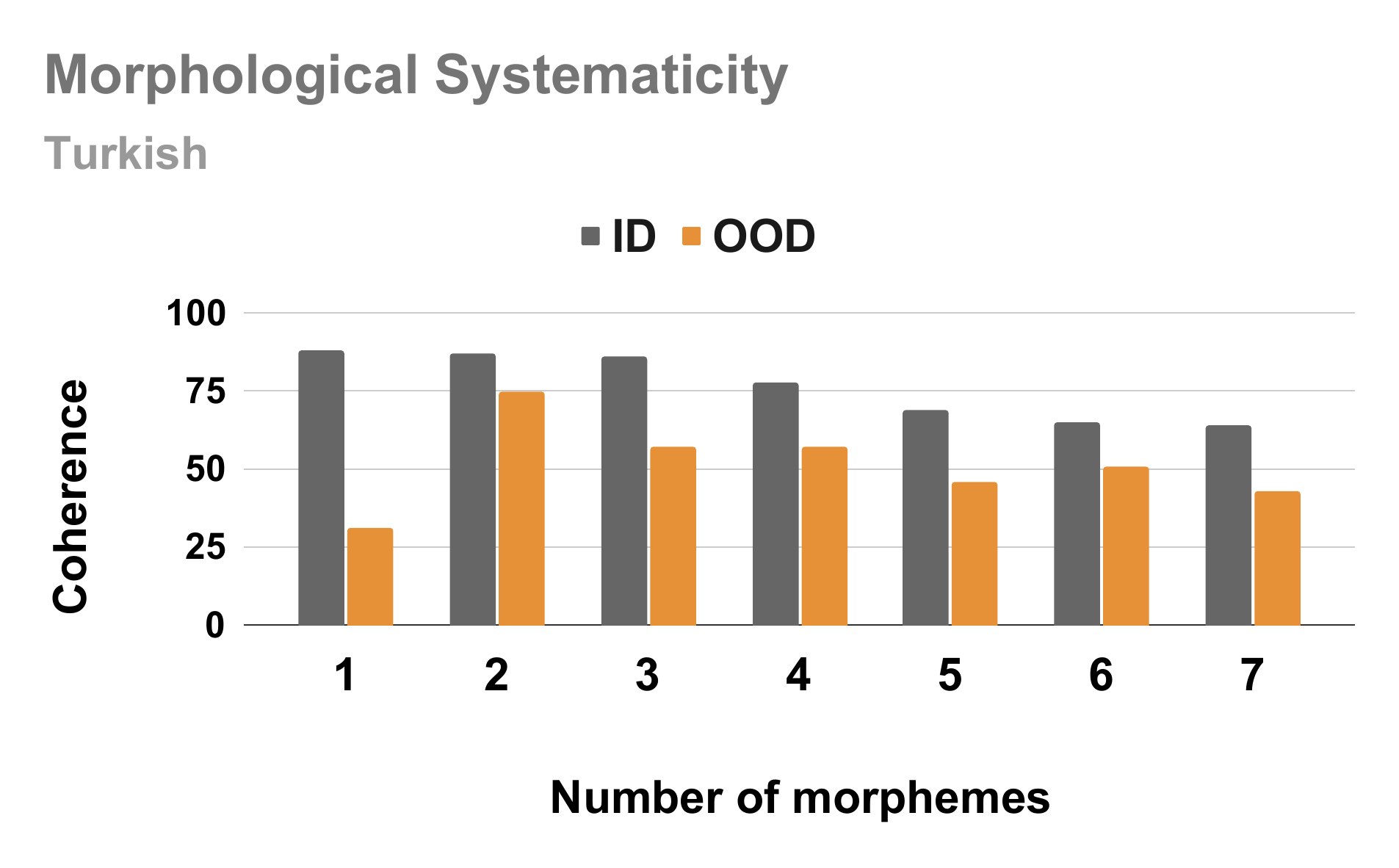}
\end{subfigure}
\caption{\textbf{GPT-4 morphological productivity and systematicity task results for Turkish stratified by number of bound morphemes}. Detailed results are in Appendix Tables \ref{tab:results-tr-en-by-affix-gen-acc-s5}, \ref{tab:results-tr-en-by-affix-disc-f1-s5}, \ref{tab:results-tr-en-by-affix-disc-coh-s5}. Finnish results are in Appendix Figure \ref{fig:analysis-by-affix-fi-en}.}
\label{fig:analysis-by-affix-tr-en}
\end{figure*}

\subsection{Effect of Morphological Complexity}
\label{sec:effect-num-morpheme}
Recent works have shown that morphological complexity plays a crucial role in the morphological generalization abilities of LLMs \cite{anh-etal-2024-morphology, czarnowska-etal-2019-dont, cotterell2018all}. Morphological complexity is typically categorized into \textit{integrative} (I-complexity) which refers to the predictability of inflected form and \textit{enumerative} (E-complexity) complexity which refers to the number of cases and inflectional paradigms in language grammar \cite{ackerman2013}. While both languages we study are morphologically complex, our test suites include inflectional and derivational forms of varying length in the number of morphemes (1-7 in Turkish and 1-6 in Finnish). This allows us to study the effect of within-language E-complexity on the performance of our models. Figure \ref{fig:analysis-by-affix-tr-en} summarizes the GPT-4 performance for both tasks stratified by the number of bound morphemes on the Turkish data. On the productivity task, we observe a sharp downward trend (plummeting to nearly zero) in performance as the number of morphemes increases for both ID and OOD test suites with a relatively constant gap between ID and OOD performance while humans exhibit no such dependence on complexity (Appendix Tables \ref{tab:results-tr-en-by-affix-gen-acc-s5}, \ref{tab:results-fi-en-by-affix-gen-acc-s5}). This shows that humans learn their native language robustly and can easily produce and identify long novel words while models are quite sensitive to the morphological (E-) complexity.

On the systematicity task, Macro-F1 scores for ID and OOD remain mostly unchanged as complexity increases, but coherence scores show a negative correlation with the increasing morphological complexity. We also observe a surprisingly low performance on 1-morpheme OOD words which we attribute to the varying number of negative options by morpheme length and potential shortcuts in longer morpheme words, as discussed in Appendix \ref{sec:app-effect-morph-comp}.

\subsection{Effect of Context}
\label{sec:effect-context}
While our core tasks are somewhat synthetic in nature, we do also experiment with more realistic versions where we provide the model a sentence as an additional context. Specifically, we frame them as sentence completion tasks where a sentence with a blank is provided and the model is asked to fill in the blank with the correct word derived from the given word root and affixes (productivity task) or determine if the given derivation is the correct option for the blank (systematicity task).

\begin{figure*}[h]
\begin{subfigure}[b]{0.33\textwidth}
    \centering
    \includegraphics[width=\textwidth]{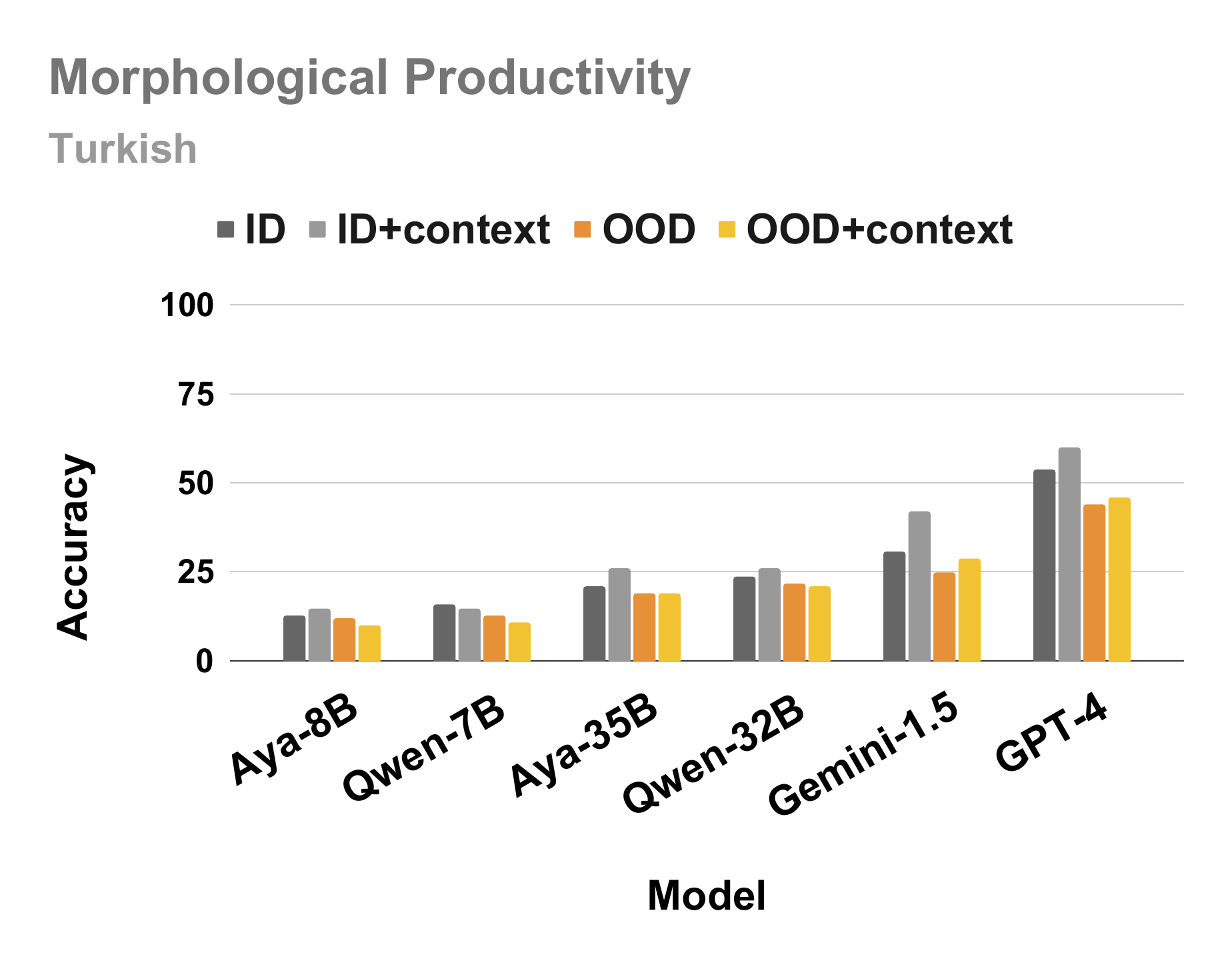}
\end{subfigure}
\begin{subfigure}[b]{0.33\textwidth}
    \centering
    \includegraphics[width=\textwidth]{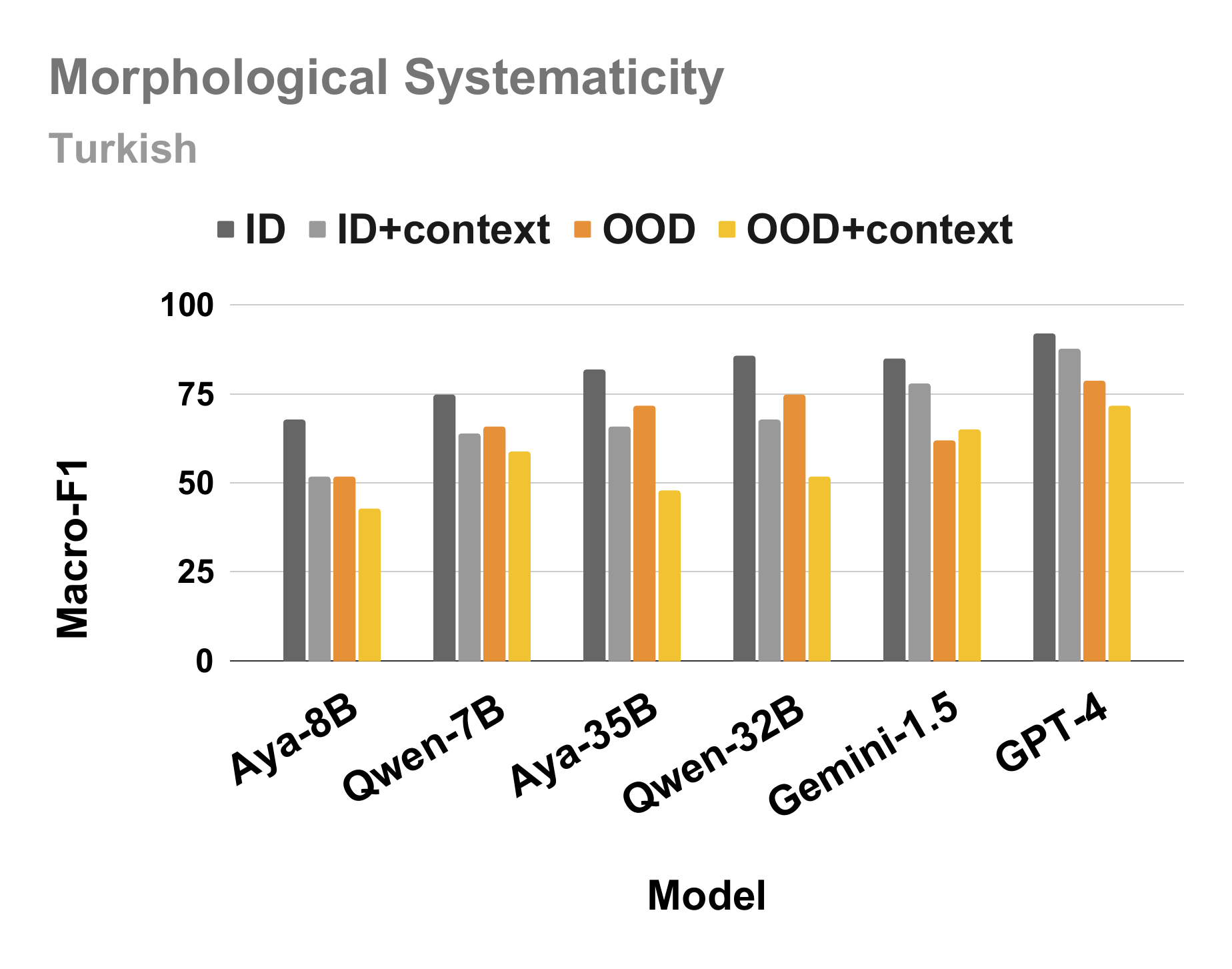}
\end{subfigure}
\begin{subfigure}[b]{0.33\textwidth}
    \centering
    \includegraphics[width=\textwidth]{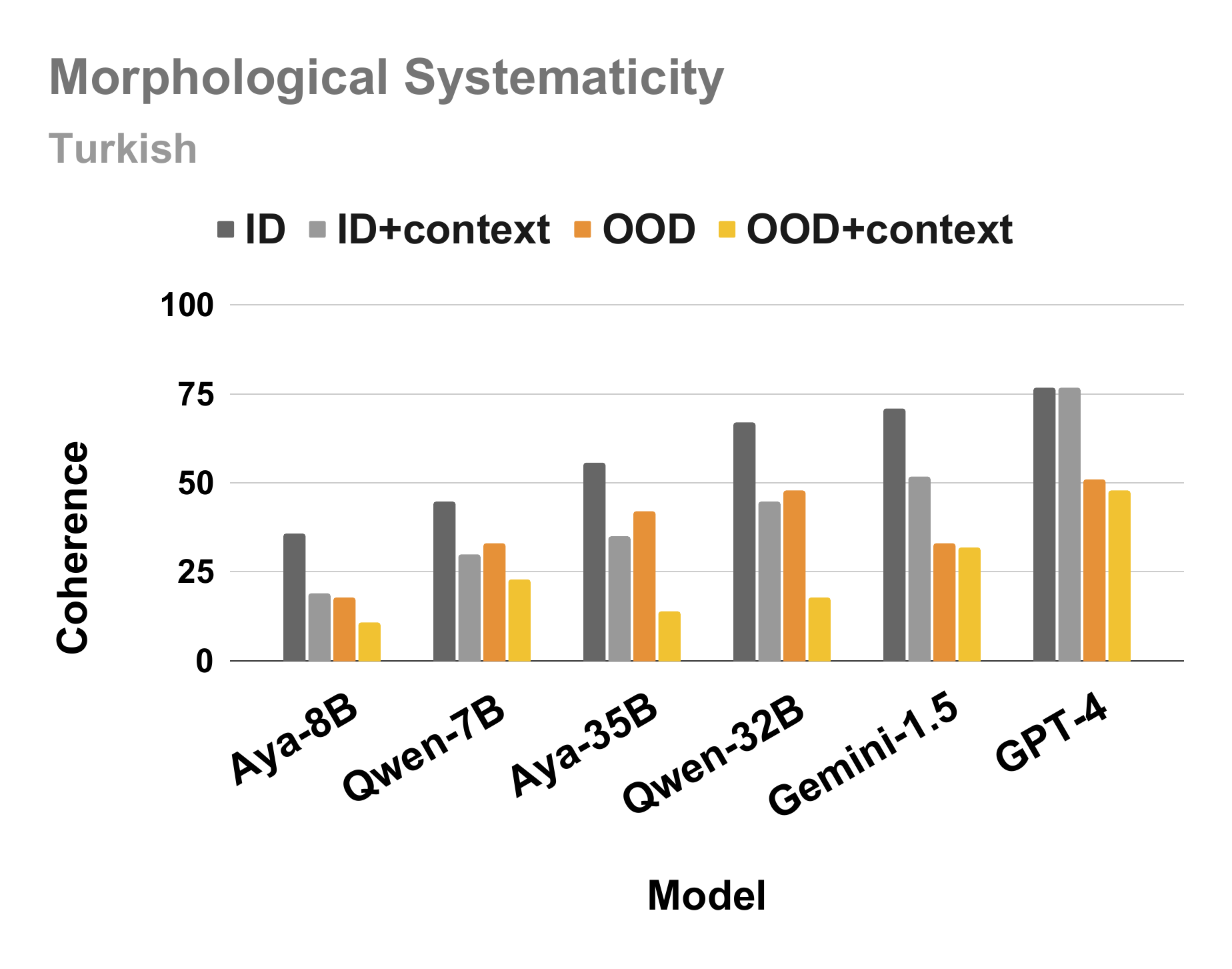}
\end{subfigure}
\caption{\textbf{Morphological productivity and systematicity task results for Turkish showing the effect of additional context}. Detailed results are in Appendix Table \ref{tab:results-tr-en-by-context-s5}. Results for Finnish are in Appendix Figure \ref{fig:analysis-by-context-fi-en}.}
\label{fig:analysis-by-context-tr-en}
\end{figure*}

Figure \ref{fig:analysis-by-context-tr-en} summarizes the results for both productivity and systematicity tasks evaluated on the Turkish data where we provide a sentence with a blank to the model as a context (i.e. sentence completion task). This results in some improvement on the productivity task, however, we observe significant decrease in performance on the systematicity task especially for smaller models such as Aya-23 and Qwen-2.5 series and in OOD setting. This could be due to the additional complexity introduced by the extra context, however, we should note that worse performance on this task implies even stronger generalization failure since this task is more real-world and closer to the next word prediction task compared to the original context-free setup.

\subsection{Effect of Tokenization}
\label{sec:effect-tok}
\begin{figure}[h]
\centering
\includegraphics[width=\columnwidth]{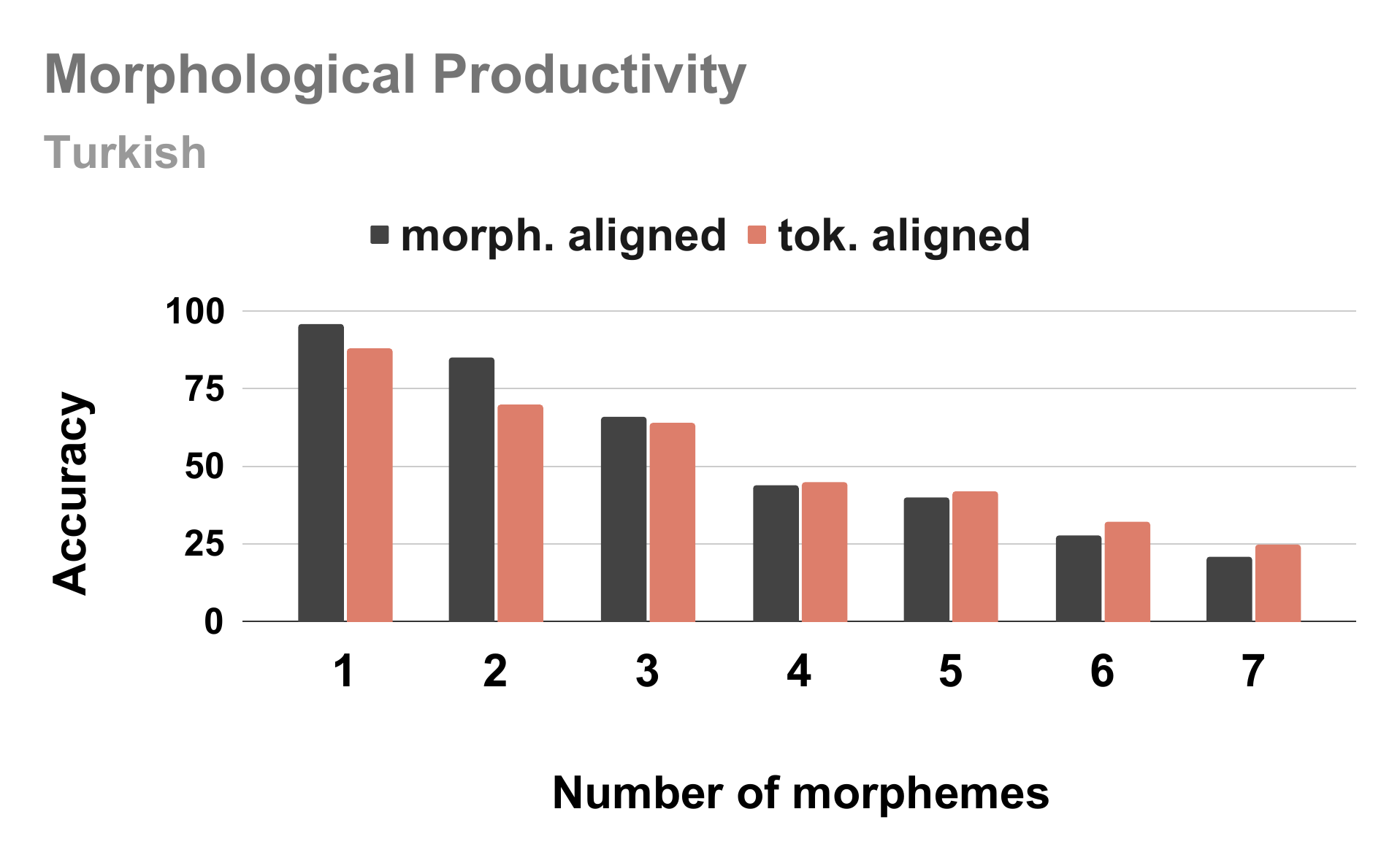}
\caption{\textbf{GPT-4 productivity task results on the ID test suite for Turkish stratified by number of bound morphemes showing the effect of tokenization}. Detailed results are in Appendix Table \ref{tab:results-tr-en-by-comp-gen-acc-s5}. }
\label{fig:analysis-by-comp-tr-en}
\end{figure}

Past work has shown that suboptimal tokenizers, especially byte-pair encoding \cite{sennrich-etal-2016-neural} used in GPT-4 have generally a negative effect on the morphological abilities of language models \cite{meyer-buys-2023-subword, bostrom-durrett-2020-byte, hofmann-etal-2021-superbizarre}. Whether the low performance of the model on the productivity task can be attributed to the suboptimal nature of the tokenization is of interest in particular because our tasks rely on the morphologically segmented morphemes while the model utilizes byte-level tokens that are mostly English. To measure the effect of the tokenization, we ran a version of the productivity task where the morphemes provided to the model are obtained by segmenting the final derivation based on the model's own tokenizer instead of the morphologically-aligned units.
Figure \ref{fig:analysis-by-comp-tr-en} compares the performance of the tokenizer-aligned morphemes with the morphologically-aligned morphemes on the ID test set.\footnote{Since we use the word root as a definition for the nonce root and the tokenizer tends to break the words into meaningless chunks, we skip this experiment on the OOD test set.} We see that the performance in both cases is very similar to each other which points to a possibility that tokenization may not be the underlying issue behind the low performance. This finding is also consistent with some past work on exploring morphological capabilities of ChatGPT \cite{weissweiler-etal-2023-counting}\footnote{\mete{We note that we perform this analysis only with subword-level tokenizers, but not character-level tokenizers for two reasons: 1) To the best of our knowledge, at the time of writing this paper, there were no instruction-tuned multilingual language models for Turkish and Finnish that uses character-level tokenizers; 2) Past work has shown that character-level tokenizers do not offer any significant advantages over subword-level tokenizers in morphological generalization \cite{Libovick2021WhyDP, Toraman2022ImpactOT}.}}

\begin{figure*}[h]
\begin{subfigure}[b]{0.33\textwidth}
    \centering
    \includegraphics[width=\textwidth]{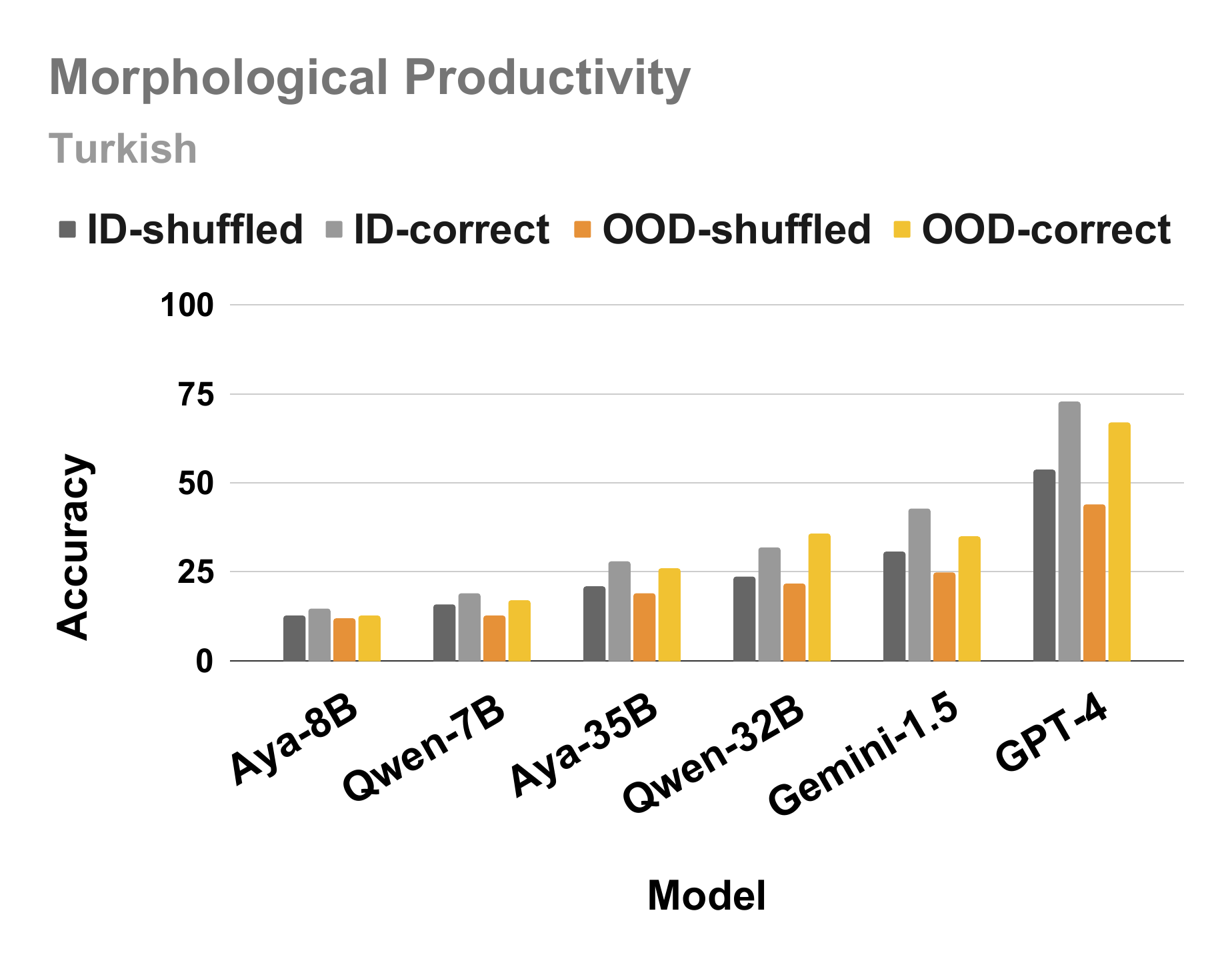}
\end{subfigure}
\begin{subfigure}[b]{0.33\textwidth}
    \centering
    \includegraphics[width=\textwidth]{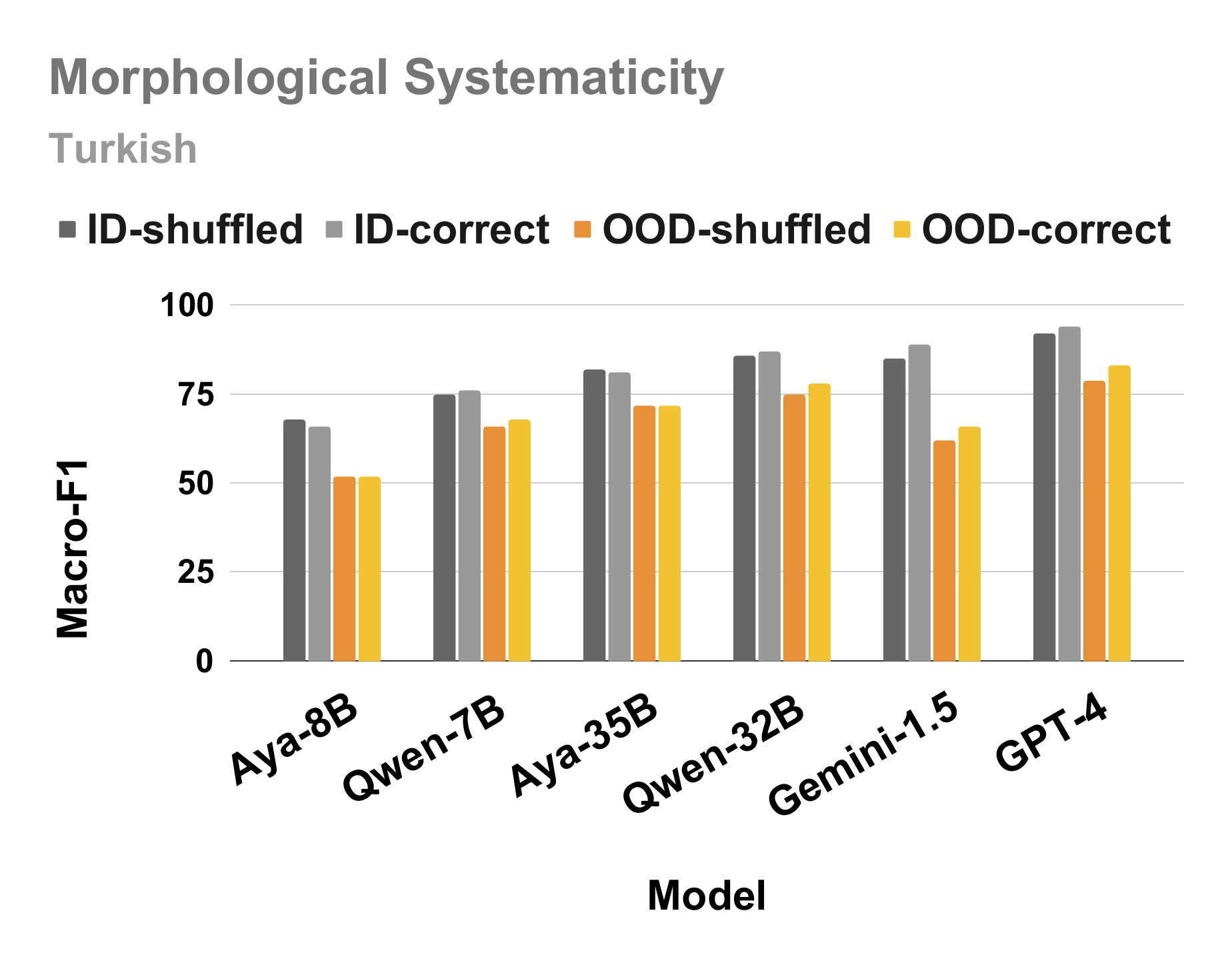}
\end{subfigure}
\begin{subfigure}[b]{0.33\textwidth}
    \centering
    \includegraphics[width=\textwidth]{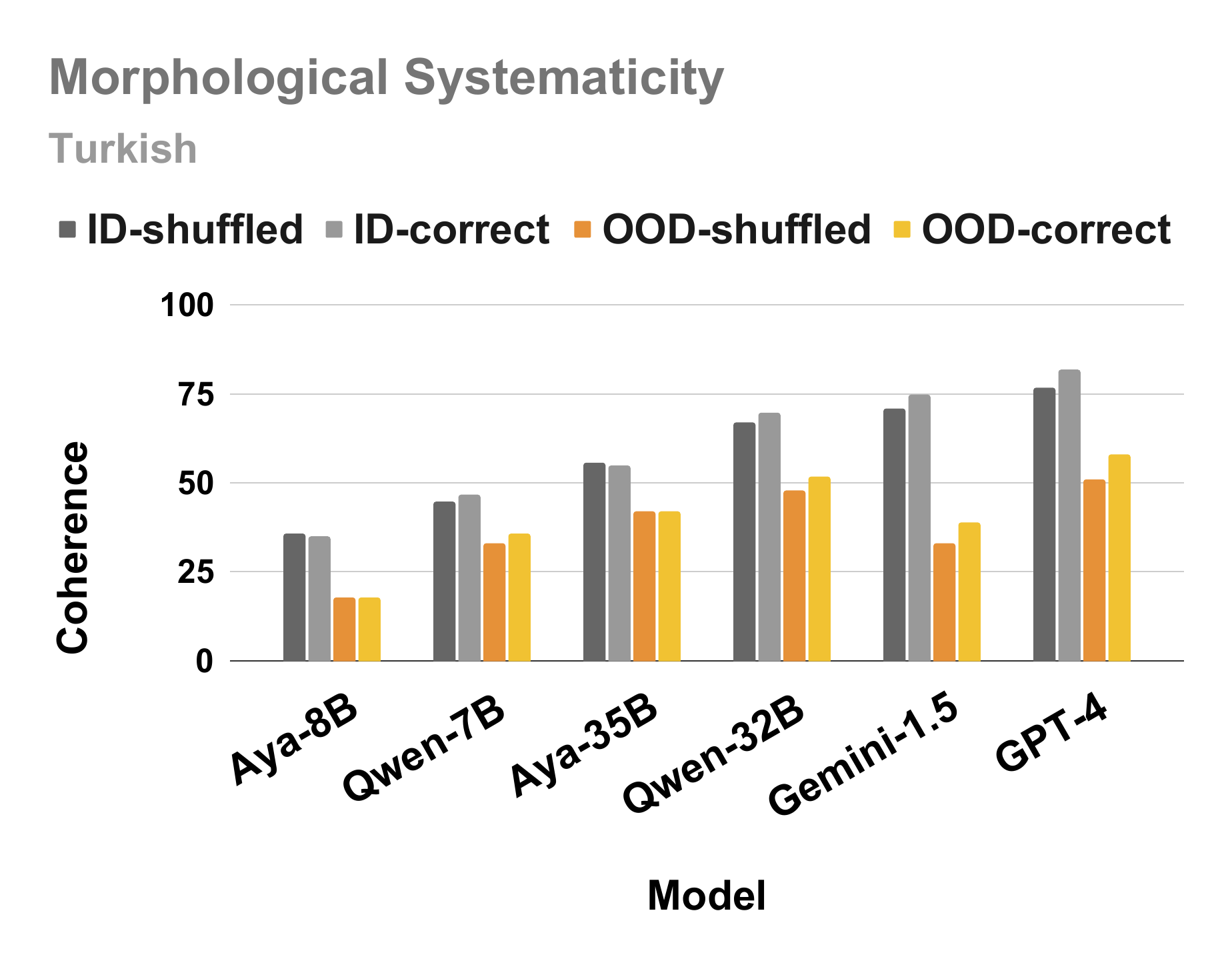}
\end{subfigure}
\caption{\textbf{Morphological productivity and systematicity task results for Turkish showing the effect of the morpheme order}. Detailed results are in Appendix Table \ref{tab:results-tr-en-by-order-s5}. Results for Finnish are in Appendix Figure \ref{fig:analysis-by-order-fi-en}.}
\label{fig:analysis-by-order-tr-en}
\end{figure*}

\subsection{Effect of Morpheme Order}
\label{sec:effect-morpheme-order}
Since our goal is to study the ability of LLMs to combine the morphological units in the correct order, in all of our experiments we shuffle the order of the units in the prompts. However, given that models are sensitive to small prompt changes \cite{pezeshkpour2023largelanguagemodelssensitivity, zhu2024promptrobustevaluatingrobustnesslarge, wang2023adversarialdemonstrationattackslarge, zhao2021calibrateuseimprovingfewshot}, we also analyze the effect of changing the morpheme order on the performance of the model. To this end, we run our main experiments with all the morphemes in their correct order and report the results in Figure \ref{fig:analysis-by-order-tr-en}. We can see that this small change improves the performance across both tasks and models and especially, in the productivity task, the improvement can be up to $20\%$. This shows that models understand the tasks and can provide a correct answer by simply copying the morphemes when they are given in the correct order, however, they struggle to compose the correct order themselves. This further indicates that LLMs lack the necessary robust compositional generalization in morphology.

\subsection{Effect of Negative Sample Selection}
\label{sec:effect-neg-selection}
\begin{figure}[h]
\begin{subfigure}[b]{\columnwidth}
    \centering
    \includegraphics[width=\textwidth]{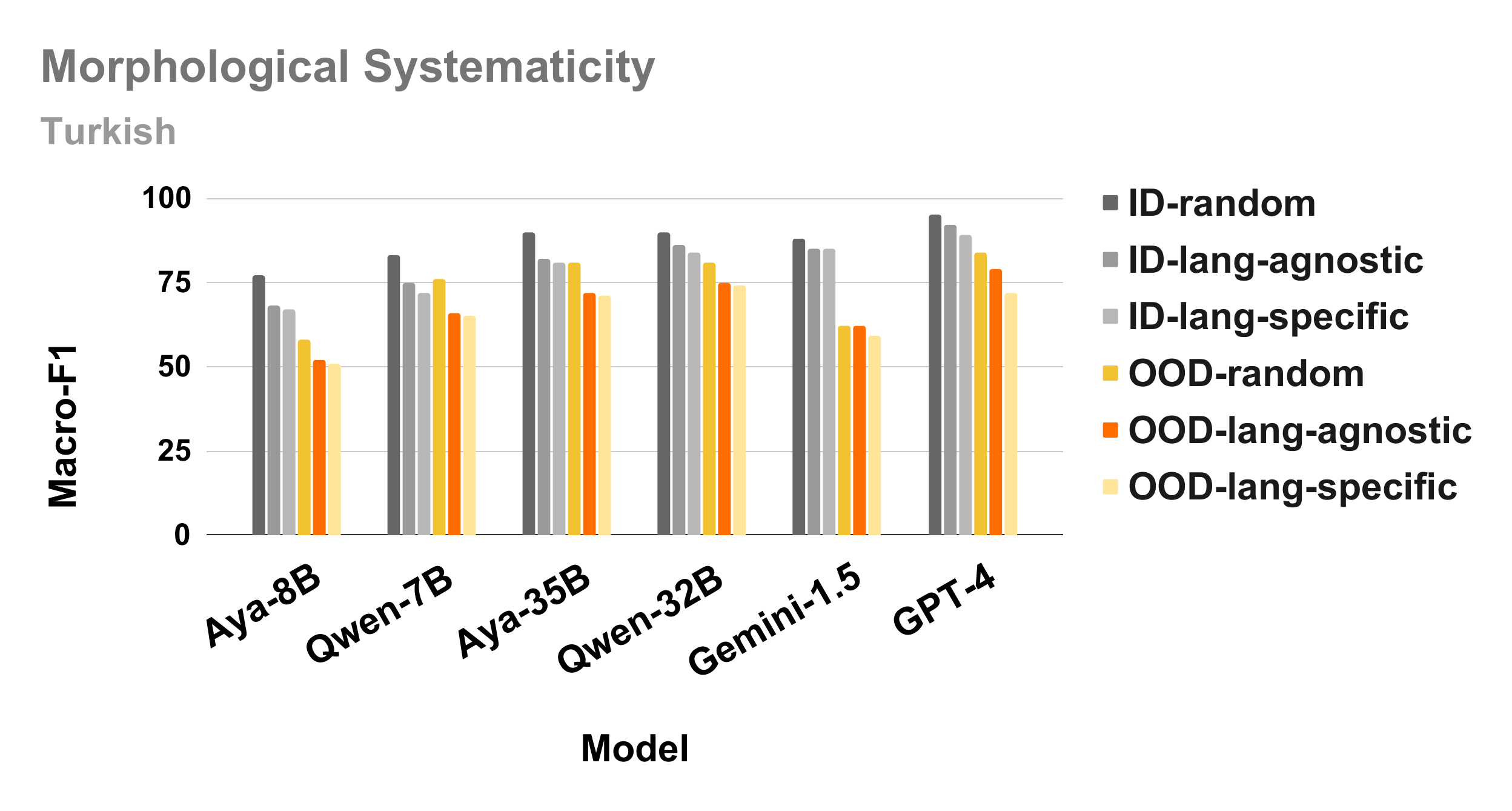}
\end{subfigure}
\begin{subfigure}[b]{\columnwidth}
    \centering
    \includegraphics[width=\textwidth]{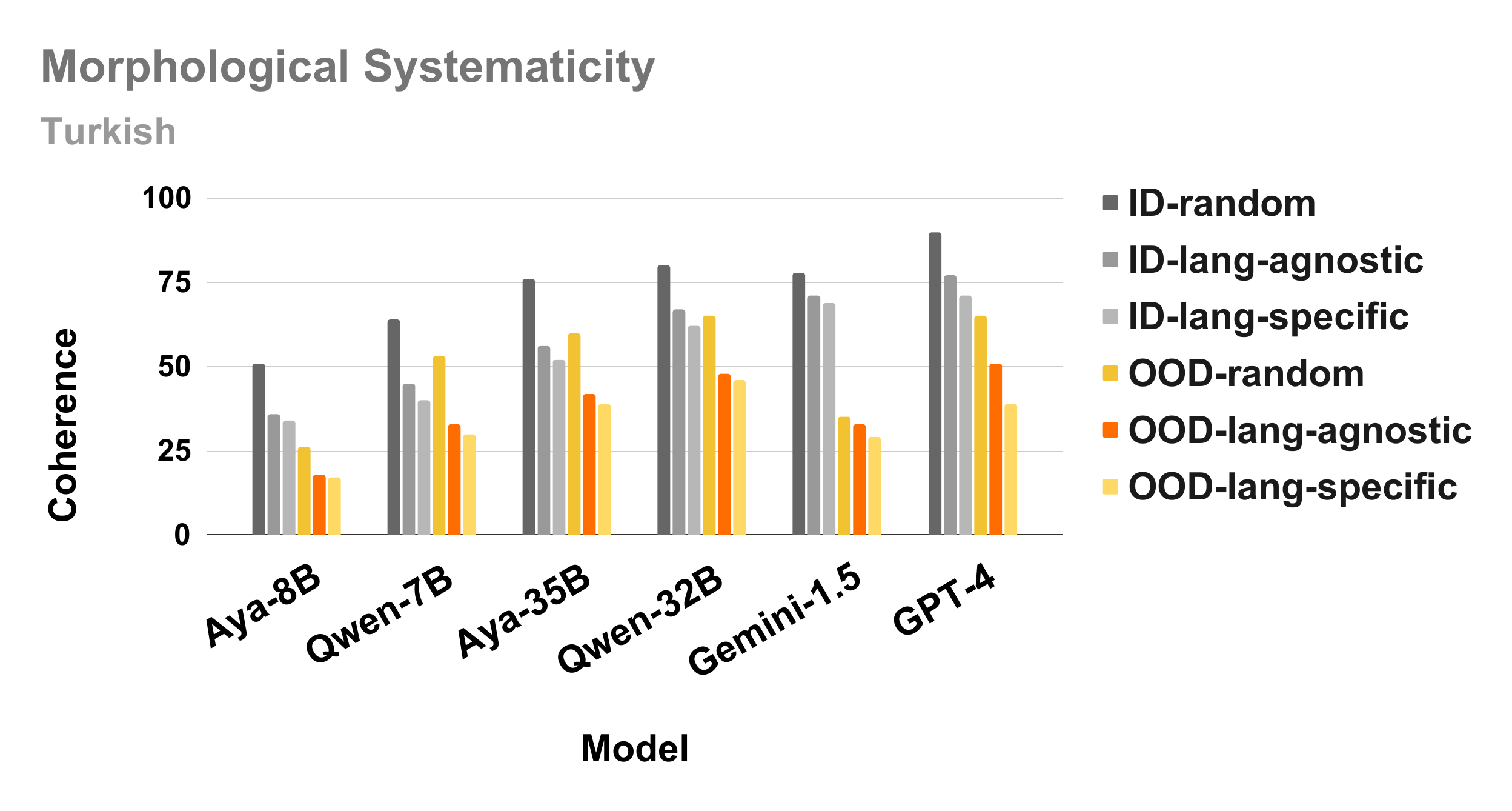}
\end{subfigure}
\caption{\textbf{Morphological systematicity task results for Turkish showing the effect of different negative sample selection strategies}. Detailed results are in Appendix Table \ref{tab:results-tr-en-by-negsel-s5}.}
\label{fig:analysis-by-negsel-tr-en}
\end{figure}

In our systematicity task, we generate negative samples (i.e. derived combinations that are not grammatically correct) by permuting the order of morphemes attached to the root. While the number of permutations is manageable for 2 or 3 morphemes (e.g., 2!=2, 3!=6), it grows rapidly with more morphemes (e.g., 6!=720). Evaluating all permutations would be ideal for robust systematicity testing, but this is infeasible due to high computational costs. Instead, we can select a subset of reasonable size to be a representative sample of all possible negative options. However, the strategy for which samples and how many to select can be somewhat arbitrary. Therefore, we experiment with three different selection strategies, and set the number of selections to four for simplicity: 1) \textbf{random} where we randomly select four negative options; 2) \textbf{language-agnostic heuristic} where we select the top four negative options that are closest to the positive option measured by Levenshtein distance (our default strategy); and 3) \textbf{language-specific heuristic} where we employ linguistic features of the tested language to filter out options that may be "too easy" for the model. We found one such heuristic for Turkish test suite based on the fact that Turkish phonology does not allow two adjacent vowels in morpheme combinations which we describe in Appendix \ref{sec:app-negsel-tr}. We report the results of these different negative sample selection experiments in Figure \ref{fig:analysis-by-negsel-tr-en}. We see that the random selection has the highest performance on both ID and OOD test sets, followed by the language-agnostic and language-specific strategies. This implies that all our previous model results might be an upper bound and the true performance gap compared to humans is even larger than what we observe.

\subsection{Error Analysis}
In order to understand the limitations of language models on our tasks, we manually analyze $30$ Turkish word derivations for each morpheme combination length (1-7) and for both productivity ID and OOD test sets resulting in a total of $178$ and $185$ derivations from GPT-4 that are incorrect. We annotate each generation on three criteria: 1) whether the generation is an \textbf{invalid} word (i.e. grammatically incorrect word) 2) whether the generation is \textbf{unfaithful} (i.e. generation does not follow the productivity task constraints) and 3) whether the generation includes any \textbf{hallucinations} (i.e. whether the generation has extra morphemes not mentioned in the task prompt). Our analysis shows that while on the OOD test set, GPT-4 generates a grammatically incorrect word most of the time ($79\%$), this proportion is significantly lower for the ID test set ($31\%$). However, on the ID test set, we observe a high unfaithfulness and hallucination ratio ($91\%$ and $67\%$) meaning that most of the valid generations do not follow the task constraints. On the other hand, we see lower unfaithfulness and hallucination ratios on the OOD test ($75\%$ and $52\%$ respectively) which points to a \textit{real word bias} also reported by \cite{weissweiler-etal-2023-counting} where the model is biased toward generating frequent words for word roots existing in a given language irrespective of the underlying task. In other words, OOD setting forces the model to perform the true morphological generalization task which it fails as indicated by the higher percentage of invalid derivations. To identify the root causes of some of these errors, we analyze the GPT-4 chain-of-thought answers on the Turkish data and reveal several failure modes such as \textbf{sequential dependency errors}, \textbf{semantic misinterpretations}, \textbf{lack of grammatical knowledge}, and \textbf{unfaithful reasoning}, all of which we detail with examples in Appendix \ref{sec:error-analysis}. Finally, we also analyze the few errors human annotators made and find that these errors are either trivial typos or failure to notice an extra letter in a long word.

%% file: 07_conclusion.tex
\section{Conclusion}
\label{sec:conclusion}
In this paper, we proposed a novel experimental paradigm to test morphological generalization abilities of large language models through compositionality. Our tasks target measuring morphological productivity and systematicity in a given language. We applied these tasks on the morphologically complex languages of Turkish and Finnish and evaluated morphological compositional generalization abilities of several state-of-the-art large language models. Our experimental results and analysis reveal a significant gap in the performance of LLMs compared to humans with respect to generalization in morphology of \mete{agglutinative languages}.

%% file: 08_limitations.tex
\section*{Limitations}
\label{sec:limitations}
While our novel tasks are language, dataset, and model-independent, our study only focused on two agglutinative languages and a few large language models. Therefore, the applicability of our findings in other languages and models should be further studied. We also mainly focused on the grammatical validity of the words, whereas it would be equally interesting to study the capacity of LLMs to produce and understand novel semantically and pragmatically valid derivations. While we have also optimized our prompts to be as simple and maximally instructive and tested in multiple languages and in chain-of-thought setting, whether a different set of prompts would produce the same results is not clear. \mete{Finally, we mainly evaluate models using greedy decoding due to the deterministic nature of our tasks and additionally only experiment with temperature and top-p sampling, however, the effect of different decoding strategies needs to be explored.}

%% file: 10_acknowledgements.tex
\section*{Acknowledgments}
We gratefully acknowledge the support of the Swiss National Science Foundation (grant 205121\_207437: C - LING) and the Microsoft Accelerating Foundation Models Research Program. Defne would also like to acknowledge the support of the National Science Foundation under grant DGE-2022040. We also thank Mammad Hajili, Osman Batur Ince, Omer Goldman, members of the NLP Lab at EPFL and the CCL Group at Idiap Research Institute for their valuable feedback in the early stages of this project and Raghav Mantri for his help with the Gemini experiments.

%% file: 11_appendix.tex
\section{Additional Analysis}
\subsection{Effect of Instruction Language}
\label{sec:effect-instruct-lang}
\begin{figure*}[h]
\begin{subfigure}[b]{0.33\textwidth}
    \centering
    \includegraphics[width=\textwidth]{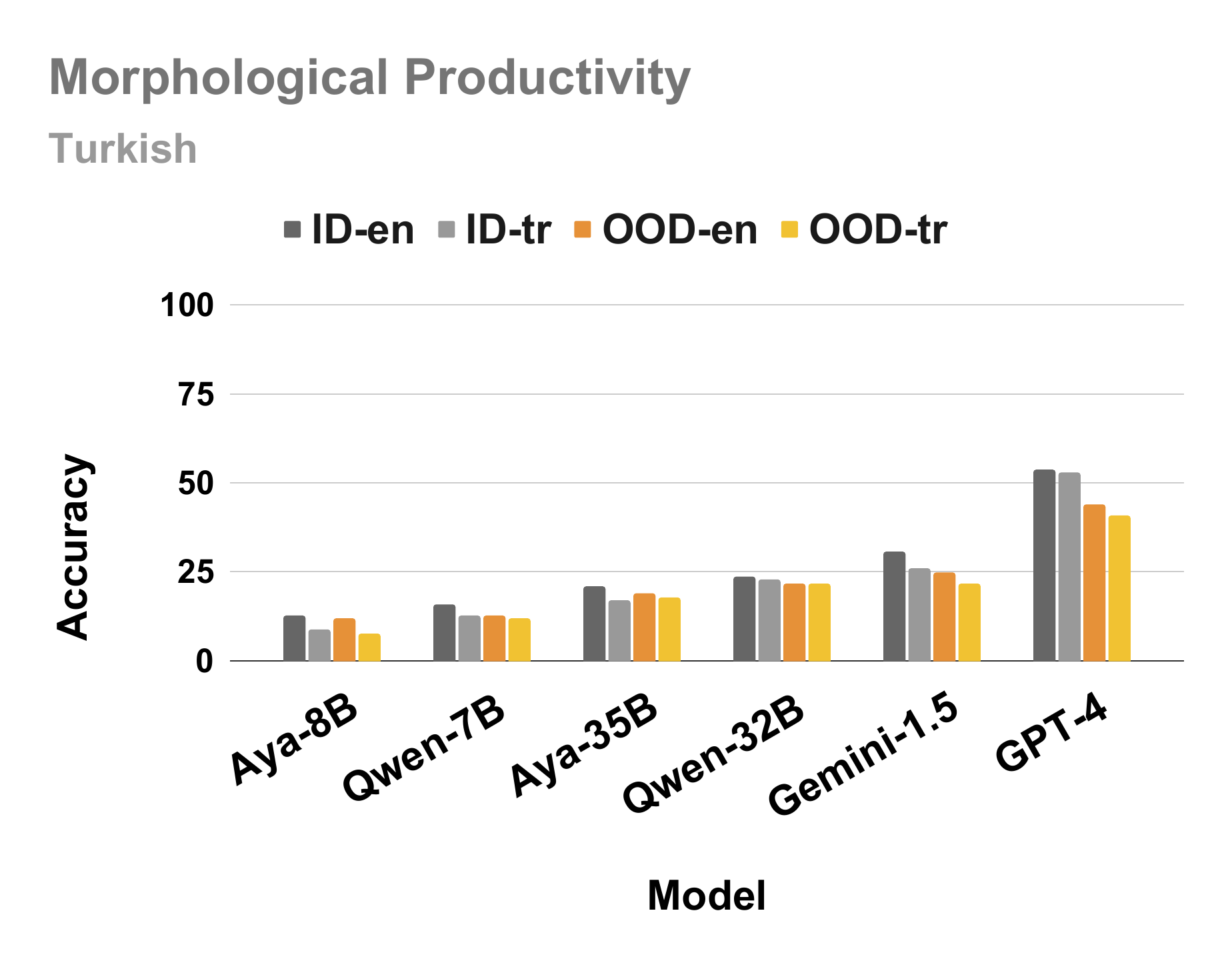}
\end{subfigure}
\begin{subfigure}[b]{0.33\textwidth}
    \centering
    \includegraphics[width=\textwidth]{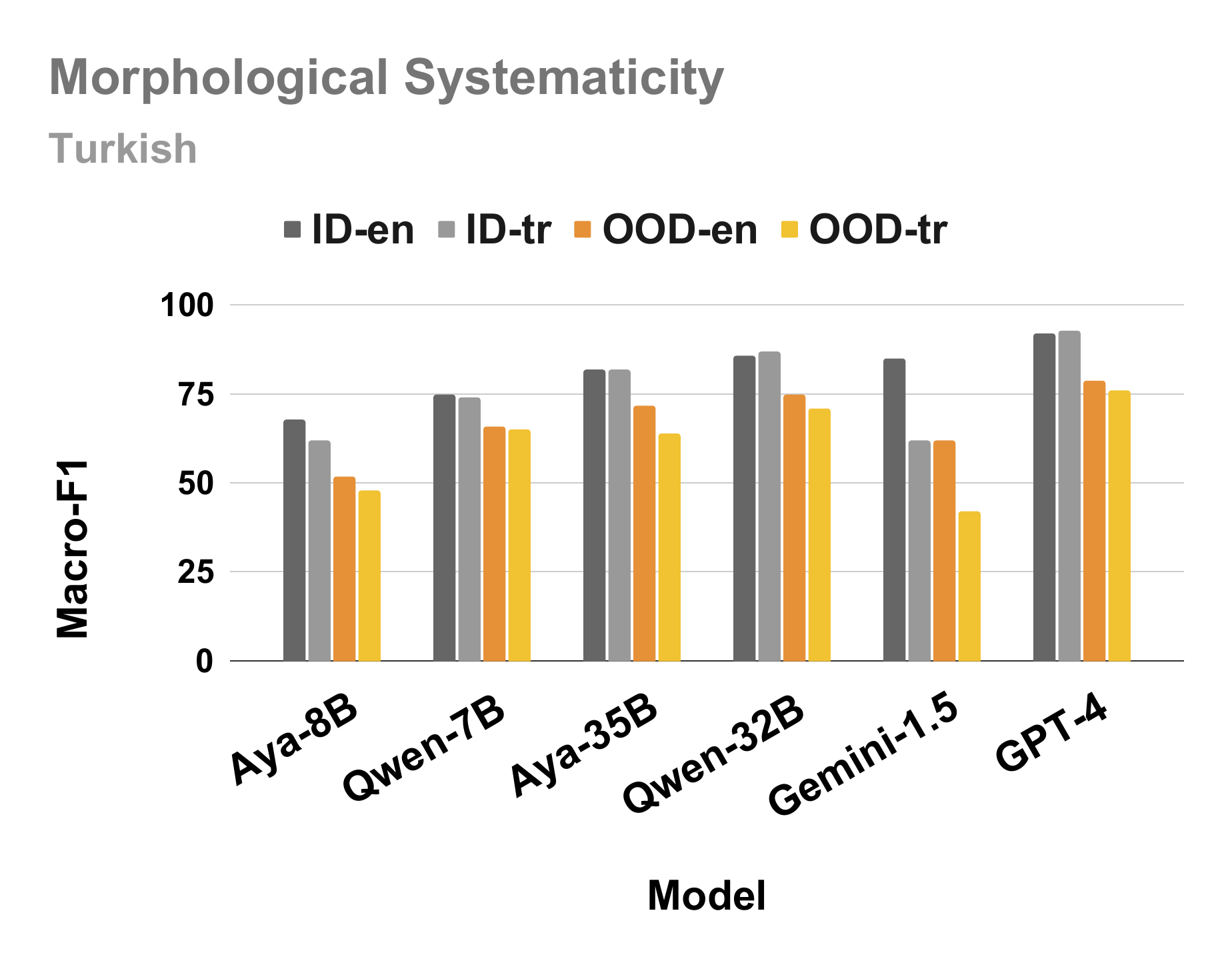}
\end{subfigure}
\begin{subfigure}[b]{0.33\textwidth}
    \centering
    \includegraphics[width=\textwidth]{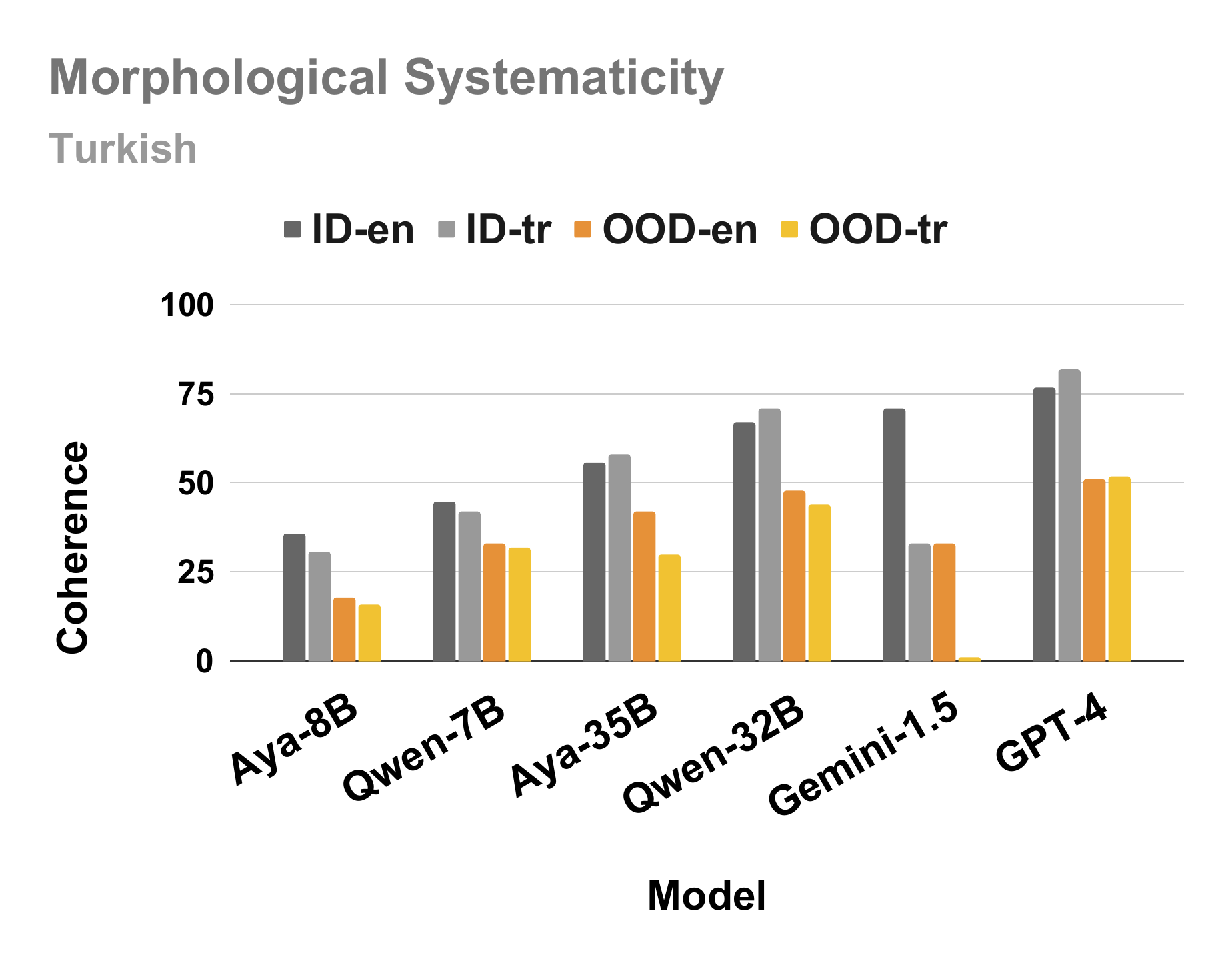}
\end{subfigure}
\caption{\textbf{Morphological productivity and systematicity task results for Turkish showing the effect of the instruction language}. Detailed results are in Table \ref{tab:results-tr-by-lang-s5}. Results for Finnish can be found in Figure \ref{fig:analysis-by-lang-fi}.}
\label{fig:analysis-by-lang-tr}
\end{figure*}

Since most LLMs are pre-trained on significantly more instruction data in English than other languages, we base most of our results on experiments where we use English as the prompt instruction language. However, as our data is in a different language, this results in a code-switched language which has been shown to be a challenge for large language models \cite{zhang-etal-2023-multilingual}. To measure the effect of the instruction language on the morphological generalization tasks, we run our experiments with Turkish and Finnish as the instruction language and report results for both tasks in Figure \ref{fig:analysis-by-lang-tr}. We mostly observe a drop or no change in performance when the instruction language is other than English. 

\subsection{Effect of Chain-of-thought Reasoning}
\label{sec:effect-cot}
\begin{figure*}[h]
\begin{subfigure}[b]{0.33\textwidth}
    \centering
    \includegraphics[width=\textwidth]{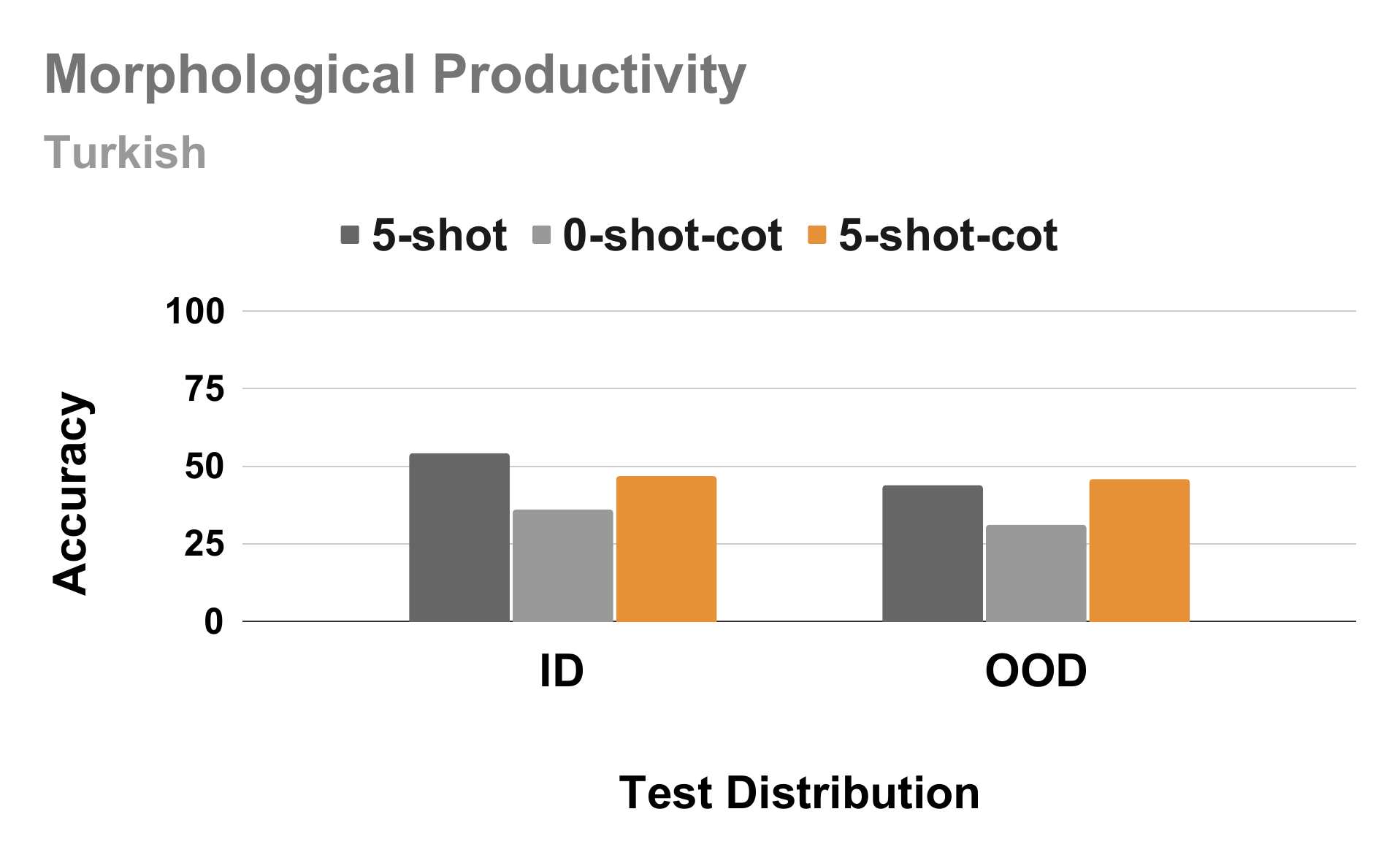}
\end{subfigure}
\begin{subfigure}[b]{0.33\textwidth}
    \centering
    \includegraphics[width=\textwidth]{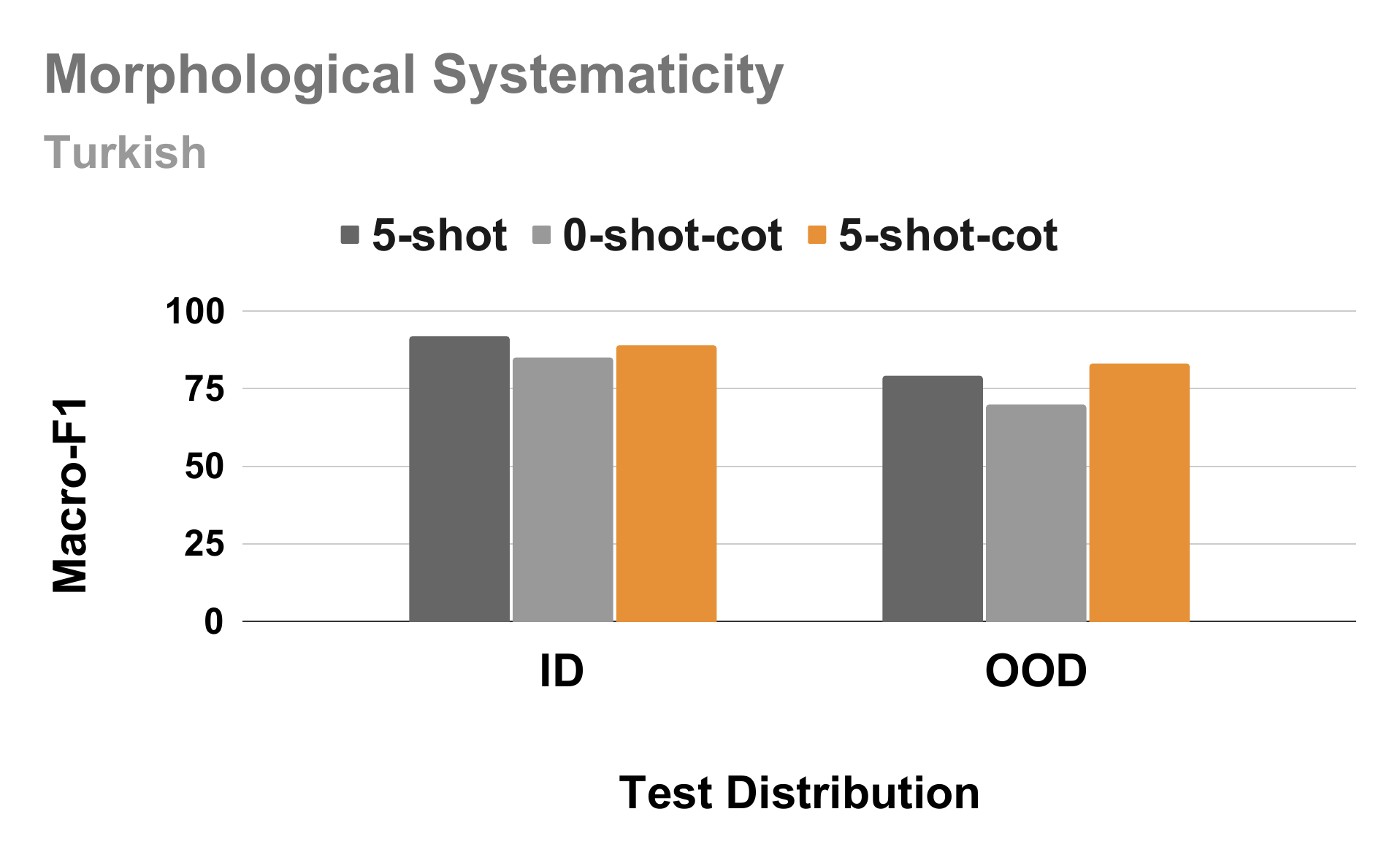}
\end{subfigure}
\begin{subfigure}[b]{0.33\textwidth}
    \centering
    \includegraphics[width=\textwidth]{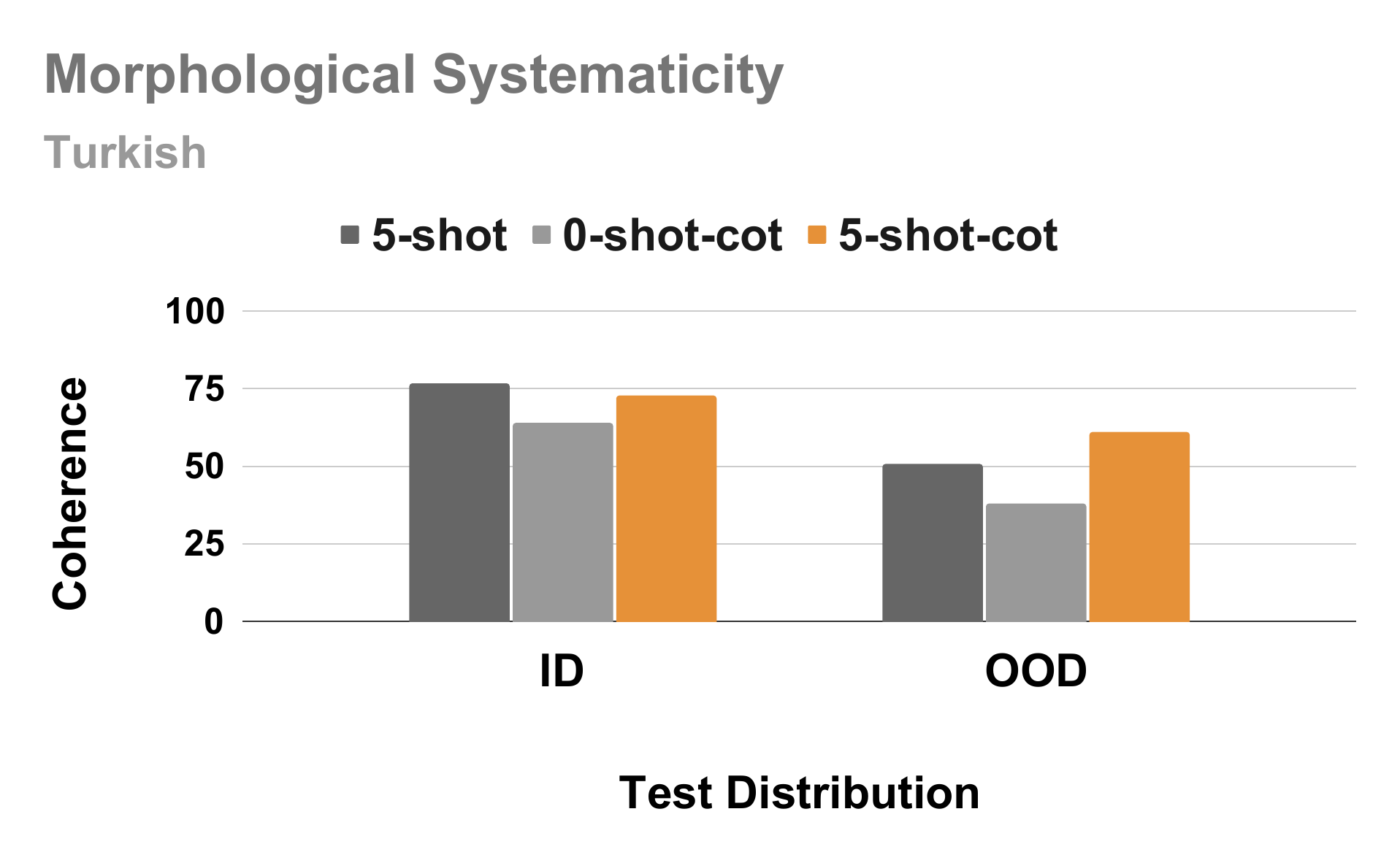}
\end{subfigure}
\caption{\textbf{GPT-4 morphological productivity and systematicity task results for Turkish showing the effect of chain-of-thought reasoning.} Detailed results are in Table \ref{tab:results-tr-en-by-cot-s505}.}
\label{fig:analysis-by-cot-tr-en}
\end{figure*}

Chain-of-thought prompting has been shown to be effective in eliciting strong reasoning capabilities from LLMs \cite{wei2023chainofthoughtpromptingelicitsreasoning}. In order to measure the effect of this reasoning technique on LLMs' performance on our tasks, we evaluate GPT-4 (the best performing model) on both productivity and systematicity tasks in zero-shot and 5-shot chain-of-thought settings. We report the results of these experiments compared with the 5-shot standard prompting in Figure \ref{fig:analysis-by-cot-tr-en}. We observe that while 5-shot chain-of-thought performance is better than the zero-shot chain-of-thought, it is slightly worse than or similar to the 5-shot standard prompting. To identify the causes of these errors, we manually analyze the several chain-of-thought answers which we describe in Appendix \ref{sec:error-analysis}.

\subsection{Further details on the effect of morphological complexity}
\label{sec:app-effect-morph-comp}
In Figure \ref{fig:analysis-by-affix-tr-en}, we observe a surprisingly low performance ($\approx 40\%$ drop from ID performance) on the 1-morpheme OOD words, but we attribute this behaviour to the varying number of negative options available for each morpheme length and possible presence of shortcuts in larger morpheme words. We should note that we have different number of total options to discriminate for a given sample depending on the number of morphemes (for 1 and 2 morphemes, we have 2 options, for 3-7, we have 5 options). Hence, a single mistake is penalized more in the former case than in the latter. However, within the former category, we see a much higher performance for 2-morpheme examples which might seem surprising, however, we hypothesize that this could be due to the presence of potential shortcuts for the model to exploit in the 2-morpheme case. Indeed, if we analyze the proportion of errors in both cases, we find that in the 1-morpheme case, a significant portion of errors ($64\%$) is false negative i.e. the model identifies a nonce root with a valid morpheme as grammatically incorrect, while this is not the case in the 2-morpheme case. However, in the 2-morpheme case, the model might be exploiting the correct order of morphemes as sole evidence for the validity of the derivation while in the 1-morpheme case, there is no such shortcut and the model should understand the applicability of the given morpheme to the given word root.

\subsection{Chain-of-thought Error Analysis}
\label{sec:error-analysis}
We randomly sample $10$ examples from the 5-shot chain-of-thought experiments on the Turkish evaluation data (per morpheme length and test distribution) where GPT-4 made an error and manually analyze its answers across both tasks. Our analysis reveals the following primary types of errors:
\begin{enumerate}
    \item \textbf{Sequential Dependency Errors}
    
    One common error we observe in the productivity task is due to the sequential processing of the given affixes by GPT-4. It typically starts applying the given affixes in the order they are given, however, since the affixes are typically given in shuffled order, this often results in an invalid word early on. The model, however, does not seem to realize its mistake and continues with the generation often confidently assigning meaning to the intermediate erroneous words. For example, given the word root "hedef" and affixes "-in", "-diğ" and "-le", it considers the affixes sequentially in this order by first producing "hedefin" which is valid, then "hedefindiğ" which is invalid, however, it interprets the generation as "which is the target" and finally produces "hedefindiğle" which it interprets as "with what is the target".
    \item \textbf{Semantic Misinterpretations}
    
    Another set of errors stems from GPT-4 misinterpreting the meaning of the individual morphemes or the whole derivation. For instance, in one example, where the given morphemes are "bağır" ("to shout"), "-sa" and "-k" and GPT-4 is asked to determine the validity of the combination "bağırsak", it misinterprets this derivation as meaning "intestine" (which is also written as "bağırsak") and argues that this derivation can not be made up of the given affixes. While this reasoning is correct, the model misses the other plausible meaning of this derivation ("as if we shout") that can be derived from the given morphemes. In another example, where the model is given the morphemes "oyna" and "-sana" and asked to produce a valid word, it misinterprets the meaning of the morpheme "-sana" as "to you" and argues that it can not be applied to the root "oyna" whereas "-sana" is a valid suffix added to verbs.
    \item \textbf{Lack of Grammatical Knowledge}
    
    Another common pattern we see can be attributed to the lack of proper grammatical knowledge. In one example, the model is given the morphemes "uyum", "-suz", "-luk" and "-ta" and asked to determine the validity of the derivation "uyumluktasuz" which is invalid, however, the model assesses the validity of each morpheme and concludes that the combination should also be valid. In another example, it tries to add a verb suffix to a noun ("yargıyoruz"). Yet in other examples, it argues that valid affixes do not exist in the language or a valid morphological combination is not possible.
    \item \textbf{Unfaithful Reasoning}
    
    Finally, we also observe a large set of reasoning errors due to inconsistent reasoning chains, hallucinations or unfaithful instruction following. For instance, in one example, GPT-4 concatenates the morphemes "unut" and "-alı" and yet derives "unutuluyor". In another example, it auto-corrects an invalid word ("kaldırınızdığda") to a valid word "kaldırdığınızda" and argues that the original derivation is correct.
\end{enumerate}

\mete{
\subsection{Effect of Decoding Strategies}
\label{sec:decoding-strategies}
We mainly experiment with greedy decoding (e.g. temperature is set to 0 and top\_p is set to 1) in all of our experiments as the nature of our tasks is deterministic. However, to check the sensitivity of our findings across diverse decoding settings, we additionally run our study with GPT-4 (the best performing model) on both tasks and languages with varying temperature and top\_p values and report the results in Tables \ref{tab:results-tr-en-by-decoding-temp}, \ref{tab:results-tr-en-by-decoding-topp}, \ref{tab:results-fi-en-by-decoding-temp} and \ref{tab:results-fi-en-by-decoding-topp} respectively. We find no significant or systematic differences across different decoding strategies which strengthens the robustness of our findings.

\subsection{Effect of Prompt Instructions}
\label{sec:prompt-instructions}
Due to the cost of LLM evaluation, we mainly experiment with one set of prompt instructions that we have found to be simple and effective through a moderate level of prompt engineering. However, to check the sensitivity of our findings across different prompt instructions, we additionally run our study with GPT-4 (the best performing model) on both tasks and languages with a paraphrased version of the original prompt instructions (found in Appendix \ref{sec:prompts}) and report the results in Tables \ref{tab:results-tr-en-by-prompt} and \ref{tab:results-fi-en-by-prompt}. We find no significant or systematic differences across different prompts which strengthens the robustness of our findings.
}

\section{Nonce word generation}
\label{sec:nonce-word-gen}
\paragraph{Turkish} To automatically generate novel nonce words in Turkish (out-of-distribution words that do not exist) that are inflected the same as the original word roots, we leverage the deterministic morphophonological features of Turkish. In particular, vowel harmony and consonant assimilation in Turkish completely determines which surface forms of the meta level morphemes would apply. Furthermore, these features depend only on the last vowel and the consonant. Hence, for a given word root in Turkish, we keep its last vowel and the consonant and randomly modify the other vowels and consonants with other vowels and consonants based on the frequency of each letter in Turkish to make sure we obtain words that would be plausible in this language. For example, if the given word root is \textit{"sanat"}, we keep the suffix \textit{"at"} as is and modify the prefix \textit{"san"} by randomly replacing each vowel in it with another vowel and consonant with another consonant. This makes sure that the words inflect the same and they are of the same length. However, if the word is too short (only two letters), and there is no prefix, we generate a random prefix of length three with vowels and consonants alternating (Turkish typically doesn't allow dense consonant clusters)
\paragraph{Finnish}
The Finnish nonce word generation is done similarly to the Turkish nonce word generation, where we alter only the word root. All consonants are replaced with other consonants and vowels with other vowels that conform to the rules of Finnish vowel harmony. 

\begin{table}
\centering
\begin{tabular}{ |l|l| } 
 \hline
\#words & 3,775,470 \\
\#unique words & 348,173 \\
\#unique roots & 9,576 \\ 
\#unique meta affixes & 103 \\
\#unique affixes & 372 \\
\#unique meta affix compositions & 21,930 \\
\#unique affix compositions & 37,853 \\
 \hline
\end{tabular}
\caption{Statistics of BTWD dataset in Turkish. Meta affixes refer to the bound morphemes that are not surface-realized.}
\label{tab:tr-data-stats}
\end{table}

\begin{table}
\centering
\begin{tabular}{ |l|l| } 
 \hline
\#samples & 1,049 \\
\#unique roots & 477 \\ 
\#unique meta affixes & 96 \\
\#unique affixes & 243 \\
\#unique meta affix compositions & 931 \\
\#unique affix compositions & 981 \\
 \hline
\end{tabular}
\caption{Statistics of our final test suite in Turkish. Meta affixes refer to the bound morphemes that are not surface-realized.}
\label{tab:tr-final-data-stats}
\end{table}

\begin{table}
\centering
\begin{tabular}{ |l|l|l| } 
 \hline
\textbf{Task} & \textbf{Test Distribution} & \textbf{$\kappa$} \\
 \hline
Productivity & ID & 0.94 \\
Productivity & OOD & 0.91 \\
Systematicity & ID & 0.94 \\
Systematicity & OOD & 0.99 \\
 \hline
\end{tabular}
\caption{Human inter-annotator agreement on Turkish test suite measured by Cohen's $\kappa$ score. We note that since the productivity task is an open-ended generative task, the chance agreement would be close to 0, hence $\kappa$ score is equal to the raw agreement.}
\label{tab:tr-kappa}
\end{table}

\begin{table}
\centering
\begin{tabular}{ |l|l|l| } 
 \hline
\textbf{Task} & \textbf{Test Distribution} & \textbf{$\kappa$} \\
 \hline
Productivity & ID & 0.77 \\
Productivity & OOD & 0.78 \\
Systematicity & ID & 0.75 \\
Systematicity & OOD & 0.84 \\
 \hline
\end{tabular}
\caption{Human inter-annotator agreement on Finnish test suite measured by Cohen's $\kappa$ score. We note that since the productivity task is an open-ended generative task, the chance agreement would be close to 0, hence $\kappa$ score is equal to the raw agreement.}
\label{tab:fi-kappa}
\end{table}

\section{Model Evaluation}
\label{sec:app-models}
We evaluate the following state-of-the-art multilingual instruction-finetuned LLMs:
\begin{itemize}
    \item \textbf{Aya-23} \cite{aryabumi2024aya23openweight} a powerful open-weights multilingual LLM serving 23 languages including Turkish. We evaluate both 8B and 35B sizes of this model series, but only on Turkish dataset as Aya-23 does not officially support Finnish yet.
    \item \textbf{Qwen-2.5} \cite{qwen2.5} recent open-weights multilingual LLM that has shown impressive results across various benchmarks and supports over 29 languages. We evaluate both 7B and 32B sizes of this model series in both languages.
    \item \textbf{Gemini-1.5} \cite{geminiteam2024gemini} a closed-source multilingual LLM that supports over 40 languages including Turkish and Finnish. We evaluate the \texttt{gemini-1.5-flash} version in both languages.
    \item \textbf{GPT-4} \cite{openai2024gpt4technicalreport} a closed-source multilingual LLM that supports many languages including Finnish and Turkish. We evaluate the \texttt{2024-02-15-preview} version in both languages.
\end{itemize}

Models are evaluated using in-context few-shot learning where number of shots take values in \{1,3,5\}. We make sure each shot has the same number of morphemes as its corresponding task example. By default, all our prompt templates are in English since LLMs are quite proficient in following instructions in this language \cite{wendler-etal-2024-llamas}, however, we also experiment with instruction templates in Turkish and Finnish which generally show worse performance (Appendix \ref{sec:effect-instruct-lang}). Similarly, while by default we use the standard prompting for all experiments, we also experiment with chain-of-thought prompting \cite{wei2023chainofthoughtpromptingelicitsreasoning}, but find very little difference in performance (Appendix \ref{sec:effect-cot}). Prompts for all tasks and languages can be found in Appendix \ref{sec:prompts}. 

\section{Data}
\label{sec:app-data}
\paragraph{Turkish}
Since the morphological analyzer we use to process the Turkish dataset \cite{ozturel2019} is based on a finite state machine relying on purely syntactic rules, it produces several alternative decompositions for some words \mete{(e.g. analyzer produces both decompositions ``an+la+dığ+ımız'' and ``anla+dığ+ımız'' for the word ``anladığımız'' )}. Hence, we further apply some language-specific heuristics to automatically filter out invalid decompositions. This preprocessing still leaves some words with multiple decompositions that can only be validated using semantics, hence, as a last step, we \mete{(authors)} manually verify and determine the final segmentation of a word.

\section{Heuristic Negative Sample Selection For Turkish}
\label{sec:app-negsel-tr}
Turkish phonology does not allow two vowels to occur together and typically employs "buffer" letters such as \textit{"y", "s"} in between these vowels, however, blindly permuting the order of Turkish morphemes inevitably results in negative samples where two vowels may occur next to each other. We hypothesized that models might easily identify these options by exploiting the "no-two-vowel" shortcut and without considering the semantic order of morphemes. To check this hypothesis, we counted the number of GPT-4 mistakes corresponding to options that both have and don't have two vowels occurring together and found that while the model makes a mistake in around $8\%$ (in-distribution) and $16\%$ (out-of-distribution) of all the negative options that do not have two vowels occurring together, these ratios are only $1\%$ and $4\%$ when we look at the negative options that have two adjacent vowels. Motivated by this discrepancy, we designed our third heuristic-based selection strategy for Turkish such that after ranking the options by their distance to the positive option, we select the top four negative options that do not have two adjacent vowels in their morpheme composition wherever possible.

\begin{table*}
\centering
\begin{tabular}{ l|l|l } 
 \hline
\textbf{ID root (OOD root)} & \textbf{Affixes} & \textbf{ID Derivations} \\
\hline
\multirow{2}{*}{sohbet (şakşet)} & \multirow{2}{*}{-ler or -yin} & sohbetler \checkmark \\
& & sohbetyin \\
 \hline
\multirow{2}{*}{sıra (yova)} & \multirow{2}{*}{-dan, -mış} & sıradanmış \checkmark \\
& & sıramışdan \\
 \hline
\multirow{5}{*}{değer (diser)} & \multirow{5}{*}{-len, -dir, -ip} & değerlendirip \checkmark \\
& & değeriplendir \\
 & & değerdirlenip \\
  & & değeripdirlen \\
  & & değerlenipdir \\
 \hline
 \multirow{5}{*}{endişe (ödlede)} & \multirow{5}{*}{-len, -dir, -me, -mek} & endişelendirmemek \checkmark \\
& & endişelendirmekme \\
 & & endişemelendirmek \\
  & & endişelenmedirmek \\
  & & endişemedirlenmek \\
 \hline
  \multirow{5}{*}{kişi (meşi)} & \multirow{5}{*}{-leş, -tir, -me, -si, -ne} & kişileştirmesine \checkmark \\
& & kişileştirnesime \\
 & & kişileştirmenesi \\
  & & kişileşsitirmene \\
  & & kişileşmetirsine \\
 \hline
   \multirow{5}{*}{hayal (rokal)} & \multirow{5}{*}{-ler, -im, -de, -ki, -ler, -i} & hayallerimdekileri \checkmark \\
& & hayalleriimdekiler \\
 & & hayalilerimdekiler \\
  & & hayallerimdeikiler \\
  & & hayallerimdekiiler \\
 \hline
    \multirow{5}{*}{sınıf (datıf)} & \multirow{5}{*}{-lan, -dır, -ıl, -ma, -lar, -ı, -nı} & sınıflandırılmalarını \checkmark \\
& & sınıflandırıılmalarnı \\
 & & sınıflardırılmalanını \\
  & & sınıflandırılmalarnıı \\
  & & sınıflandırılımalarnı \\
 \hline
\end{tabular}
\caption{Examples from our test suite in Turkish for each morpheme length from 1 to 7. OOD derivations can be obtained by replacing the ID root with the corresponding OOD root. Correct derivations are marked with \checkmark.}
\label{tab:tr-final-data-examples}
\end{table*}

\begin{table}
\centering
\begin{tabular}{ |l|l| } 
 \hline
\#samples & 480 \\
\#unique roots & 406 \\ 
\#unique affixes & 386 \\
\#unique affix compositions & 365 \\
 \hline
\end{tabular}
\caption{Statistics of our final test suite in Finnish.}
\label{tab:fi-final-data-stats}
\end{table}

\begin{table*}
\centering
\begin{tabular}{ l|l|l } 
 \hline
\textbf{ID root (OOD root)} & \textbf{Affixes} & \textbf{ID Derivations} \\
\hline
\multirow{2}{*}{yöpaikka (äydainca)} & \multirow{2}{*}{-nne or -ksi} & yöpaikkanne \checkmark \\
& & yöpaikkaksi \\
 \hline
\multirow{2}{*}{sano (tato)} & \multirow{2}{*}{-taan, -pas} & sanotaanpas \checkmark \\
& & sanopastaan \\
 \hline
\multirow{5}{*}{petoks (seloks)} & \multirow{5}{*}{-i, -ne, -en} & petoksineen \checkmark \\
& & petoksneien \\
 & & petoksneeni \\
  & & petoksienne \\
  & & petoksennei \\
 \hline
 \multirow{5}{*}{olosuhte (olanajke)} & \multirow{5}{*}{-kuvaus, -i, -lta, -an} & kuvausolosuhteiltaan \checkmark \\
& & kuvausolosuhteltaian \\
 & & kuvausolosuhteltaani \\
  & & kuvausolosuhteianlta \\
  & & kuvausolosuhteanilta \\
 \hline
  \multirow{5}{*}{palvelu (sapsevu)} & \multirow{5}{*}{-laina, -n, -välitys, -j, -a} & lainanvälityspalveluja \checkmark \\
& & lainanvälityspalveluaj \\
 & & nlainavälityspalveluja \\
  & & lainavälitysnpalveluja \\
  & & lainavälitysnpalveluaj \\
 \hline
   \multirow{5}{*}{salaisuuks (noraekauks)} & \multirow{5}{*}{-motivaatio, -n, -nostatus, -i, -a, -ni} & motivaationnostatussalaisuuksiani \checkmark \\
& & motivaationnostatussalaisuuksinia \\
 & & motivaationnostatussalaisuuksaini \\
  & & motivaationnostatussalaisuuksniai \\
  & & motivaationostatusnsalaisuuksiani \\
 \hline
\end{tabular}
\caption{Examples from our test suite in Finnish for each morpheme length from 1 to 6. OOD derivations can be obtained by replacing the ID root with the corresponding OOD root. Correct derivations are marked with \checkmark.}
\label{tab:fi-final-data-examples}
\end{table*}

\input{112_fi_figures}

\newpage

\input{results/tr/results-tr-en-default}
\input{results/fi/results-fi-en-default}


\input{results/tr/results-tr-en-by-affix/results-tr-en-by-affix_gen-acc_s1}

\input{results/tr/results-tr-en-by-affix/results-tr-en-by-affix_disc-f1_s1}

\input{results/tr/results-tr-en-by-affix/results-tr-en-by-affix_disc-coh_s1}

\input{results/tr/results-tr-en-by-affix/results-tr-en-by-affix_gen-acc_s3}

\input{results/tr/results-tr-en-by-affix/results-tr-en-by-affix_disc-f1_s3}

\input{results/tr/results-tr-en-by-affix/results-tr-en-by-affix_disc-coh_s3}

\input{results/tr/results-tr-en-by-affix/results-tr-en-by-affix_gen-acc_s5}

\input{results/tr/results-tr-en-by-affix/results-tr-en-by-affix_disc-f1_s5}

\input{results/tr/results-tr-en-by-affix/results-tr-en-by-affix_disc-coh_s5}


\input{results/fi/results-fi-en-by-affix/results-fi-en-by-affix_gen-acc_s1}

\input{results/fi/results-fi-en-by-affix/results-fi-en-by-affix_disc-f1_s1}

\input{results/fi/results-fi-en-by-affix/results-fi-en-by-affix_disc-coh_s1}

\input{results/fi/results-fi-en-by-affix/results-fi-en-by-affix_gen-acc_s3}

\input{results/fi/results-fi-en-by-affix/results-fi-en-by-affix_disc-f1_s3}

\input{results/fi/results-fi-en-by-affix/results-fi-en-by-affix_disc-coh_s3}

\input{results/fi/results-fi-en-by-affix/results-fi-en-by-affix_gen-acc_s5}

\input{results/fi/results-fi-en-by-affix/results-fi-en-by-affix_disc-f1_s5}

\input{results/fi/results-fi-en-by-affix/results-fi-en-by-affix_disc-coh_s5}

\input{results/tr/results-tr-by-lang/results-tr-by-lang_s1}

\input{results/tr/results-tr-by-lang/results-tr-by-lang_s3}

\input{results/tr/results-tr-by-lang/results-tr-by-lang_s5}

\input{results/fi/results-fi-by-lang/results-fi-by-lang_s1}

\input{results/fi/results-fi-by-lang/results-fi-by-lang_s3}

\input{results/fi/results-fi-by-lang/results-fi-by-lang_s5}

\input{results/tr/results-tr-en-by-context/results-tr-en-by-context_s1}

\input{results/tr/results-tr-en-by-context/results-tr-en-by-context_s3}

\input{results/tr/results-tr-en-by-context/results-tr-en-by-context_s5}

\input{results/fi/results-fi-en-by-context/results-fi-en-by-context_s1}

\input{results/fi/results-fi-en-by-context/results-fi-en-by-context_s3}

\input{results/fi/results-fi-en-by-context/results-fi-en-by-context_s5}


\input{results/tr/results-tr-en-by-comp/results-tr-en-by-comp_gen-acc_s1_id}

\input{results/tr/results-tr-en-by-comp/results-tr-en-by-comp_gen-acc_s3_id}

\input{results/tr/results-tr-en-by-comp/results-tr-en-by-comp_gen-acc_s5_id}


\input{results/tr/results-tr-en-by-cot/results-tr-en-by-cot_s505}

\input{results/tr/results-tr-en-by-order/results-tr-en-by-order_s1}

\input{results/tr/results-tr-en-by-order/results-tr-en-by-order_s3}

\input{results/tr/results-tr-en-by-order/results-tr-en-by-order_s5}


\input{results/fi/results-fi-en-by-order/results-fi-en-by-order_s1}

\input{results/fi/results-fi-en-by-order/results-fi-en-by-order_s3}

\input{results/fi/results-fi-en-by-order/results-fi-en-by-order_s5}

\newpage

\input{results/tr/results-tr-en-by-negsel/results-tr-en-by-negsel_s1}

\input{results/tr/results-tr-en-by-negsel/results-tr-en-by-negsel_s3}

\input{results/tr/results-tr-en-by-negsel/results-tr-en-by-negsel_s5}

\newpage
\input{111_prompts}

\input{results/tr/results-tr-en-by-decoding/results-tr-en-by-temp}
\input{results/tr/results-tr-en-by-decoding/results-tr-en-by-topp}

\input{results/fi/results-fi-en-by-decoding/results-fi-en-by-temp}
\input{results/fi/results-fi-en-by-decoding/results-fi-en-by-topp}

\input{results/tr/results-tr-en-by-prompt}

\input{results/fi/results-fi-en-by-prompt}

%% file: 112_fi_figures.tex
\begin{figure*}[h]
\begin{subfigure}[b]{0.33\textwidth}
    \centering
    \includegraphics[width=\textwidth]{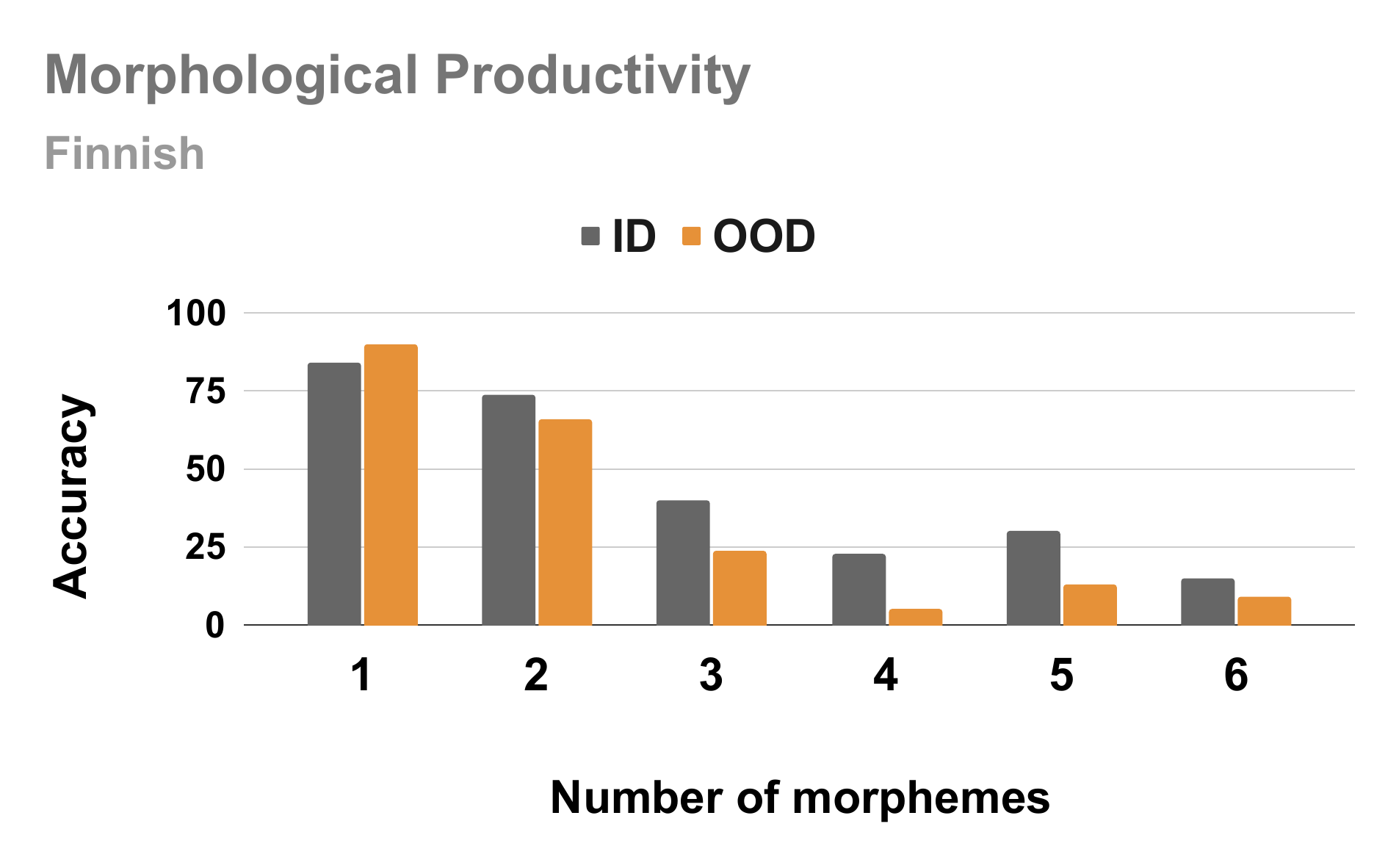}
\end{subfigure}
\begin{subfigure}[b]{0.33\textwidth}
    \centering
    \includegraphics[width=\textwidth]{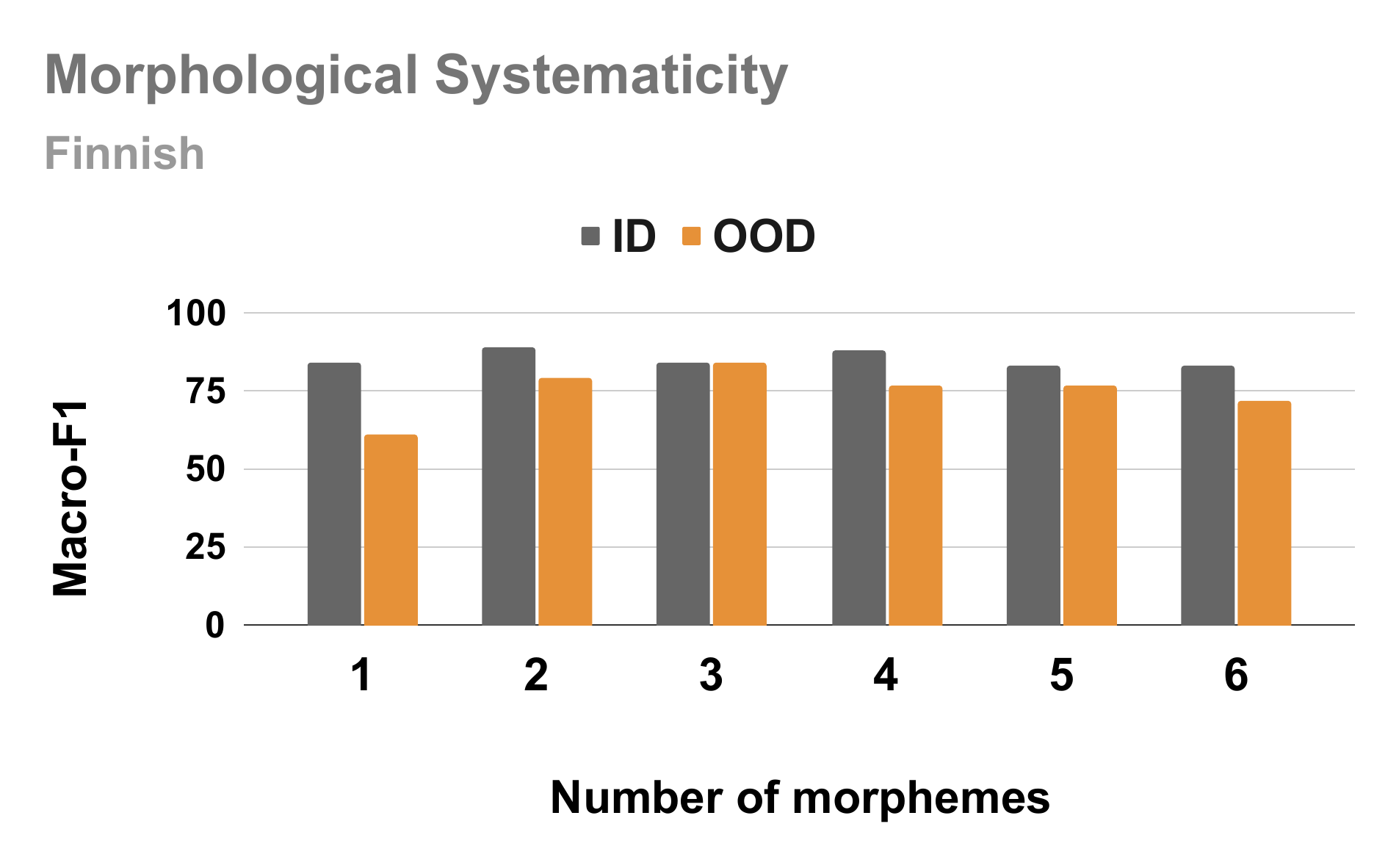}
\end{subfigure}
\begin{subfigure}[b]{0.33\textwidth}
    \centering
    \includegraphics[width=\textwidth]{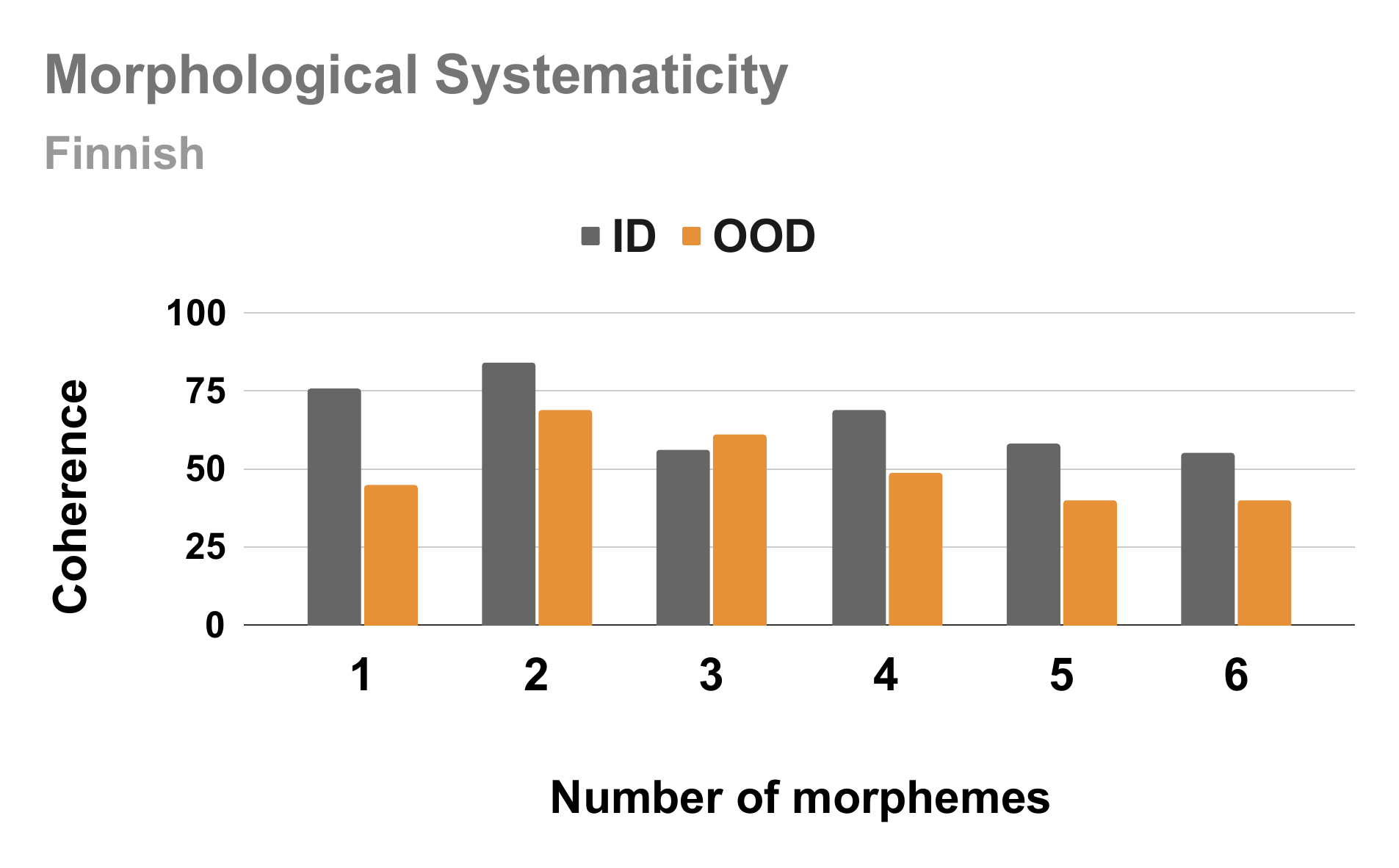}
\end{subfigure}
\caption{GPT-4 morphological productivity and systematicity task results for Finnish stratified by number of bound morphemes. Detailed results are in Tables \ref{tab:results-fi-en-by-affix-gen-acc-s5}, \ref{tab:results-fi-en-by-affix-disc-f1-s5}, \ref{tab:results-fi-en-by-affix-disc-coh-s5}.}
\label{fig:analysis-by-affix-fi-en}
\end{figure*}

\begin{figure*}[h]
\begin{subfigure}[b]{0.33\textwidth}
    \centering
    \includegraphics[width=\textwidth]{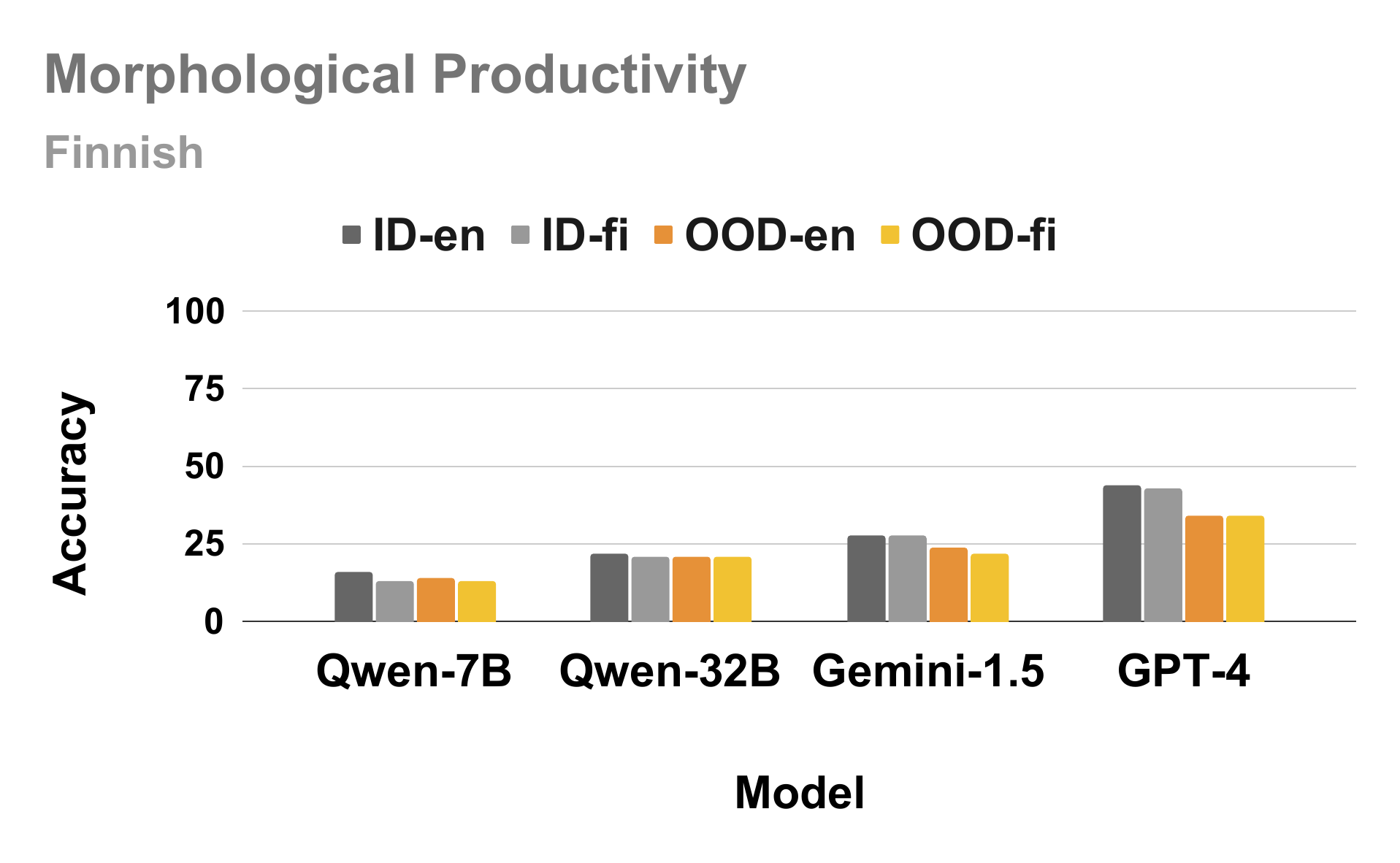}
\end{subfigure}
\begin{subfigure}[b]{0.33\textwidth}
    \centering
    \includegraphics[width=\textwidth]{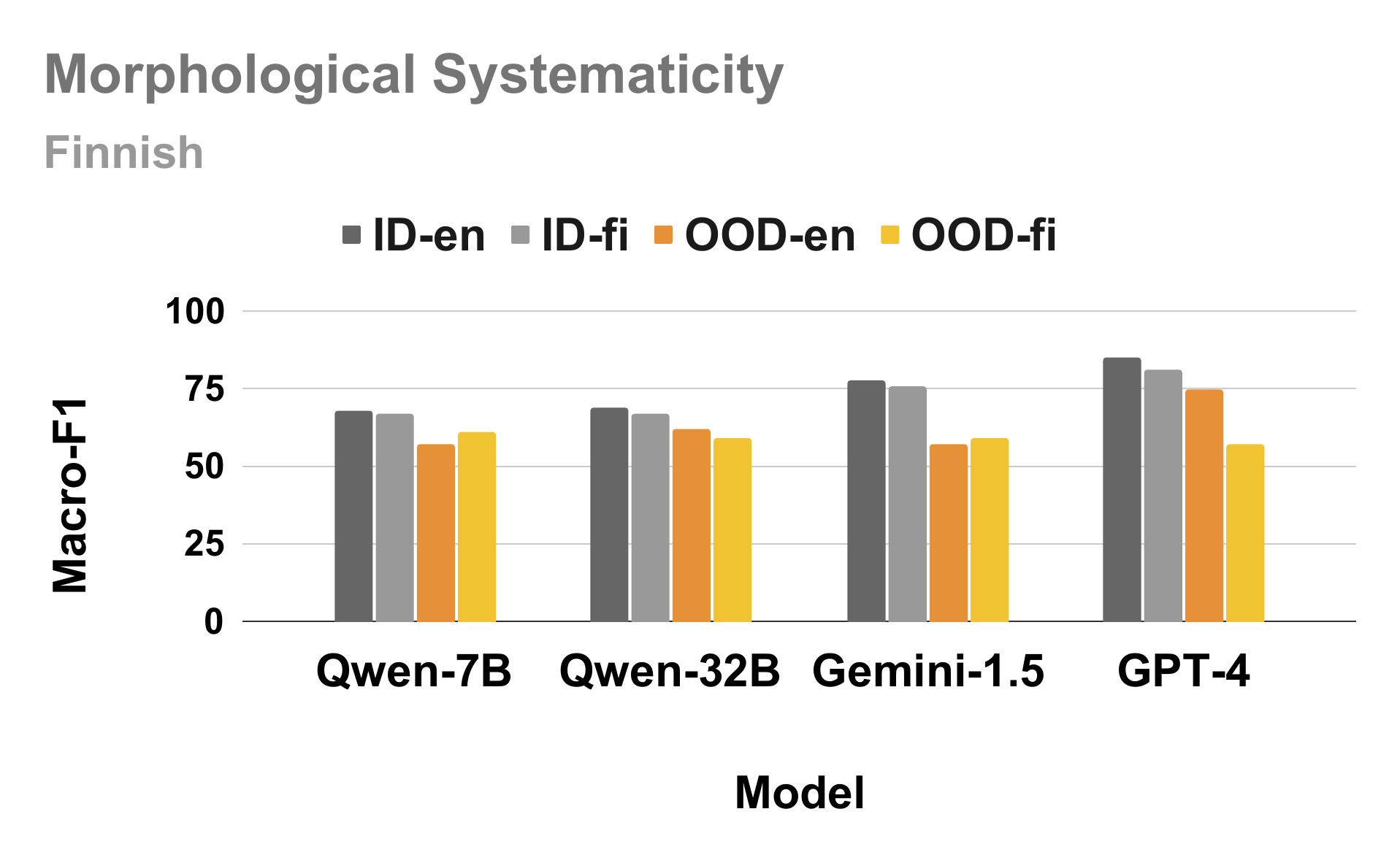}
\end{subfigure}
\begin{subfigure}[b]{0.33\textwidth}
    \centering
    \includegraphics[width=\textwidth]{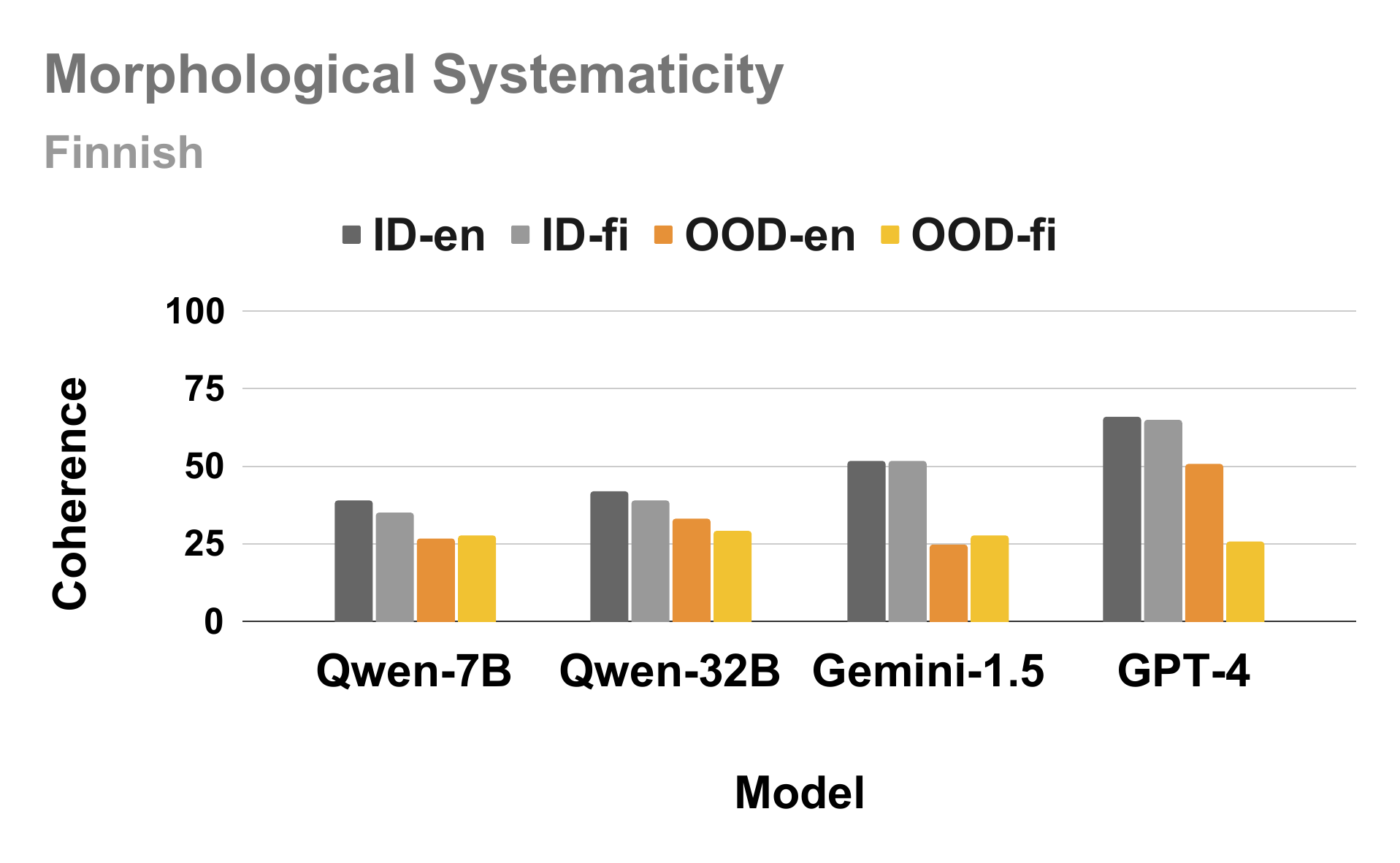}
\end{subfigure}
\caption{Morphological productivity and systematicity task results for Finnish showing the effect of the instruction language. Detailed results are in Tables \ref{tab:results-fi-by-lang-s5}.}
\label{fig:analysis-by-lang-fi}
\end{figure*}

\begin{figure*}[h]
\begin{subfigure}[b]{0.33\textwidth}
    \centering
    \includegraphics[width=\textwidth]{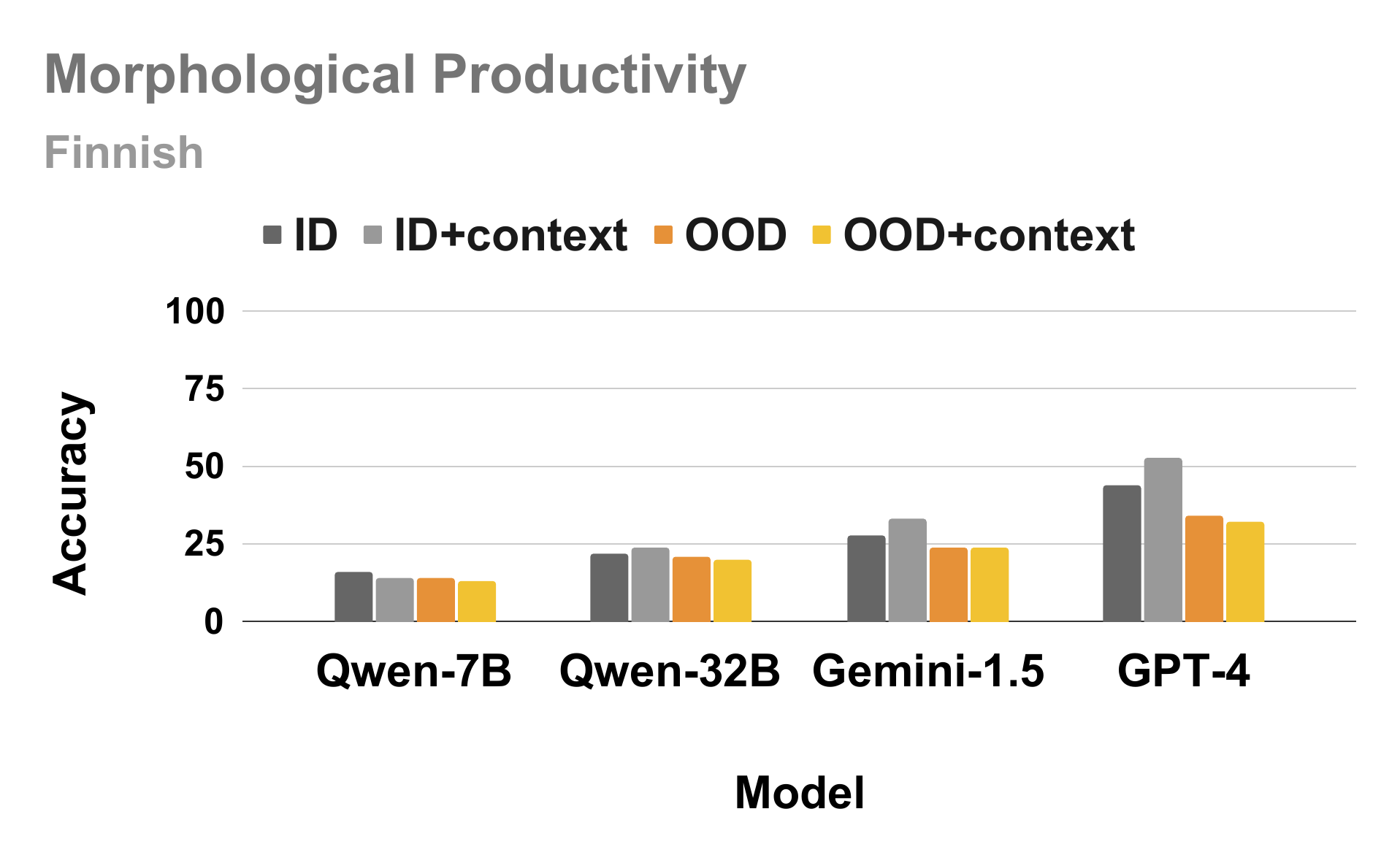}
\end{subfigure}
\begin{subfigure}[b]{0.33\textwidth}
    \centering
    \includegraphics[width=\textwidth]{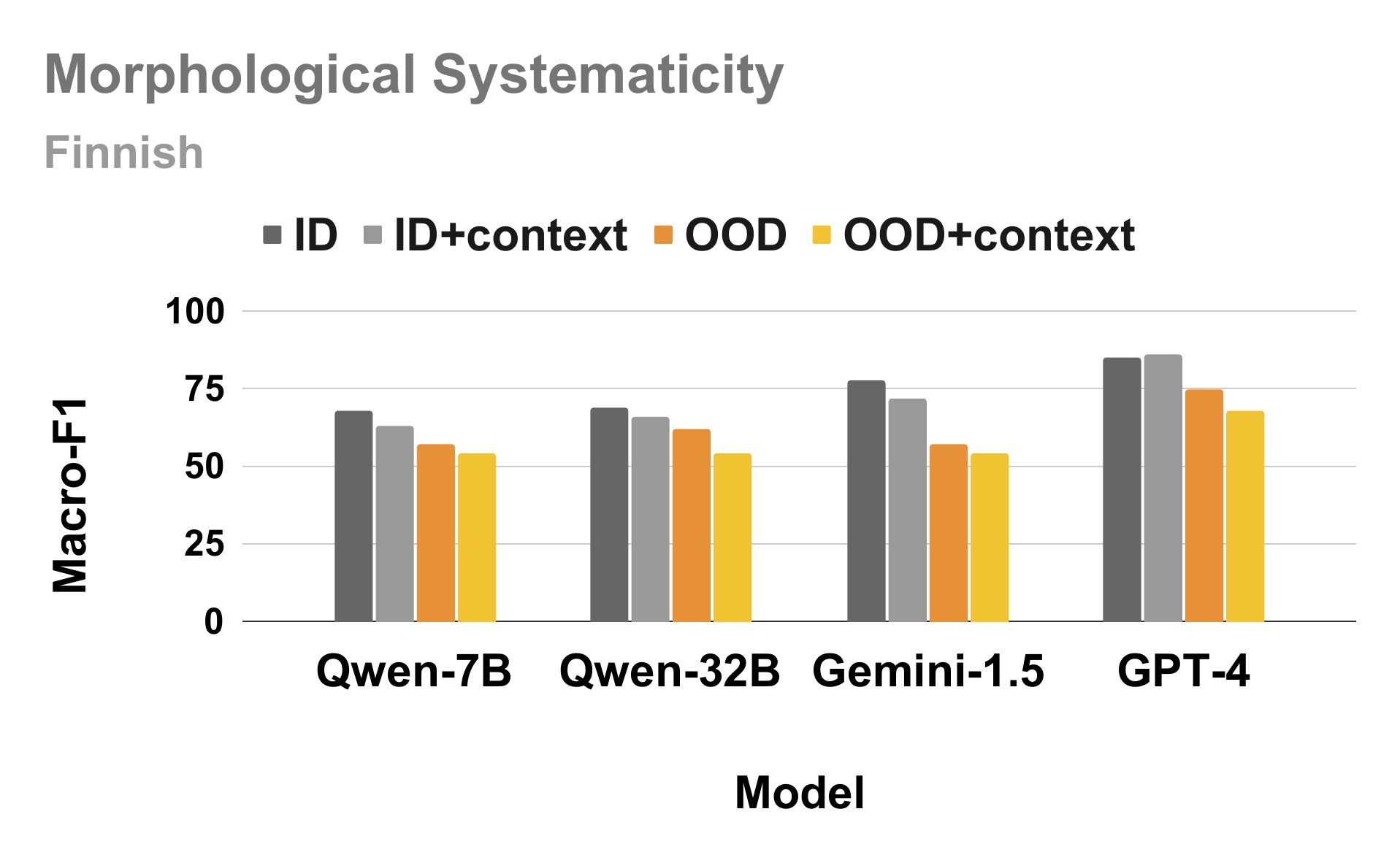}
\end{subfigure}
\begin{subfigure}[b]{0.33\textwidth}
    \centering
    \includegraphics[width=\textwidth]{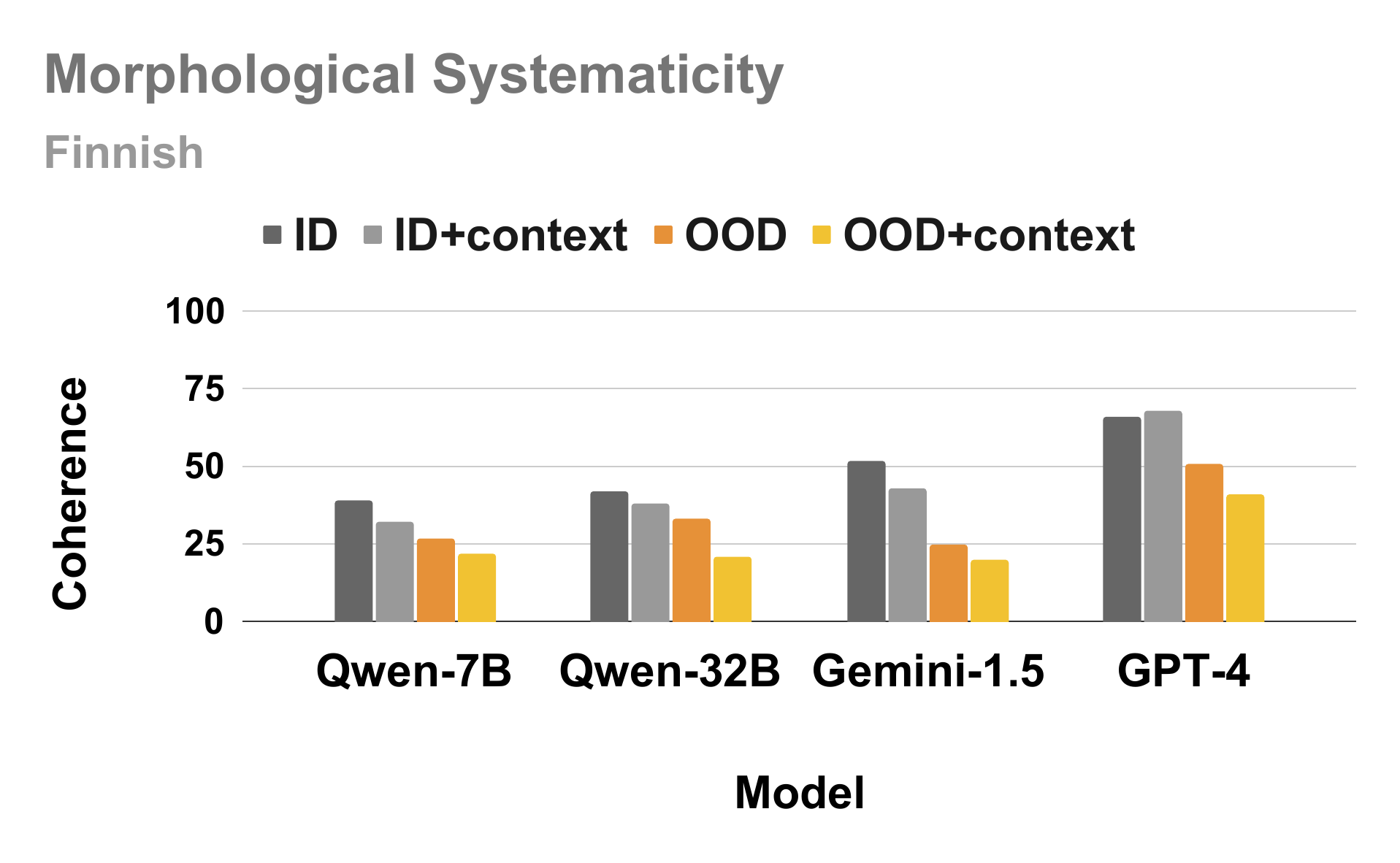}
\end{subfigure}
\caption{Morphological productivity and systematicity task results for Finnish showing the effect of additional context. Detailed results are in Table \ref{tab:results-fi-en-by-context-s5}.}
\label{fig:analysis-by-context-fi-en}
\end{figure*}

\begin{figure*}[h]
\begin{subfigure}[b]{0.33\textwidth}
    \centering
    \includegraphics[width=\textwidth]{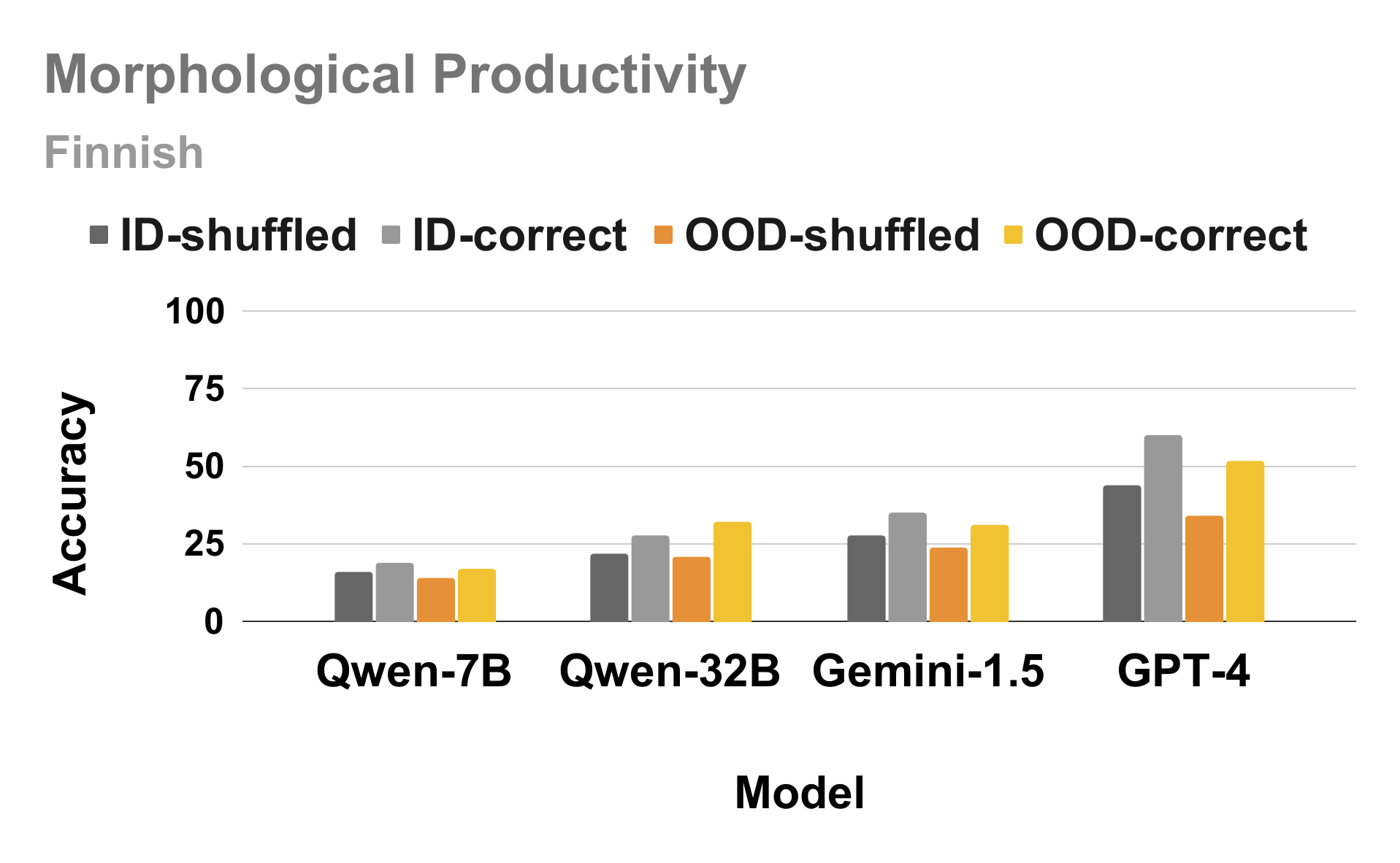}
\end{subfigure}
\begin{subfigure}[b]{0.33\textwidth}
    \centering
    \includegraphics[width=\textwidth]{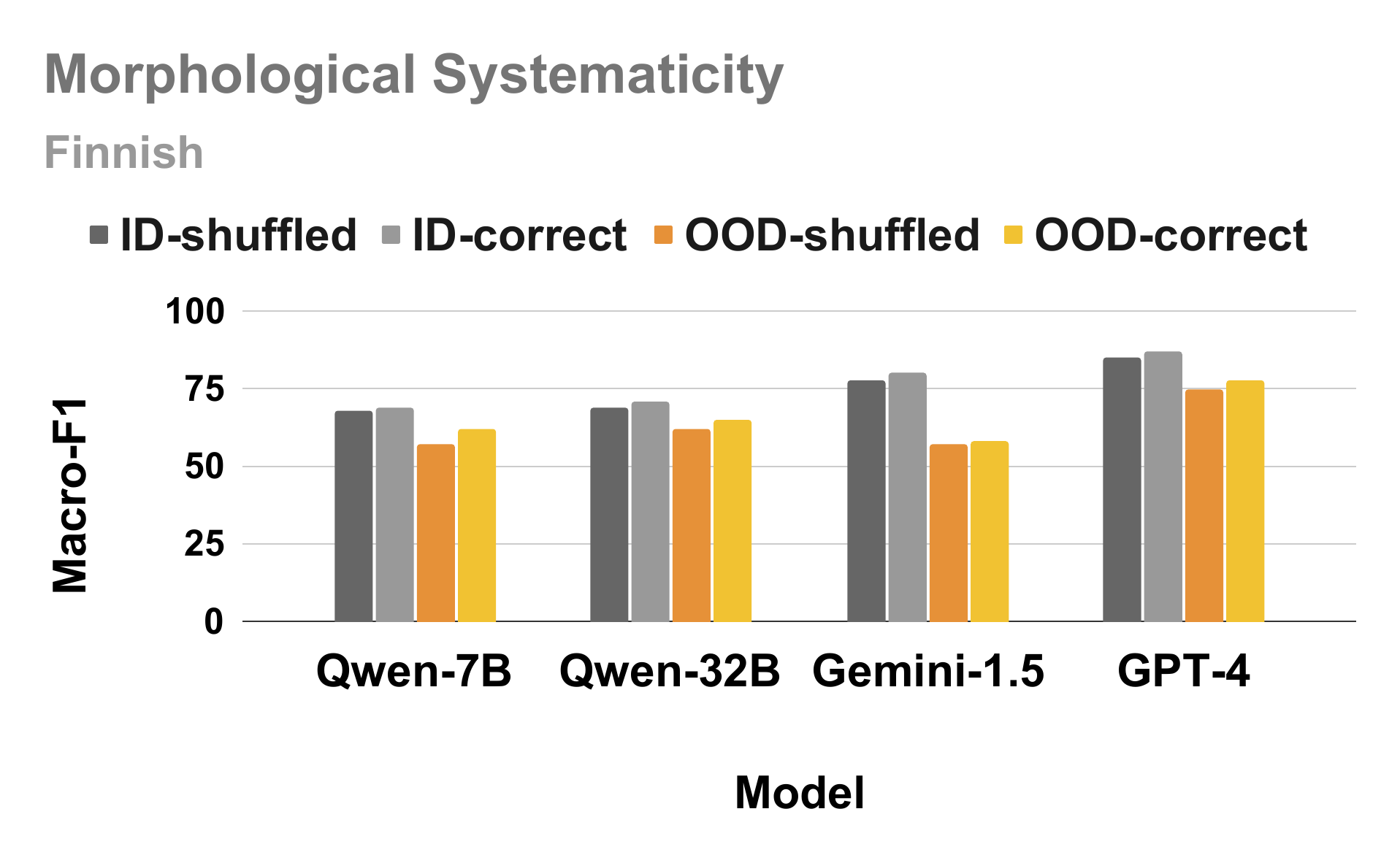}
\end{subfigure}
\begin{subfigure}[b]{0.33\textwidth}
    \centering
    \includegraphics[width=\textwidth]{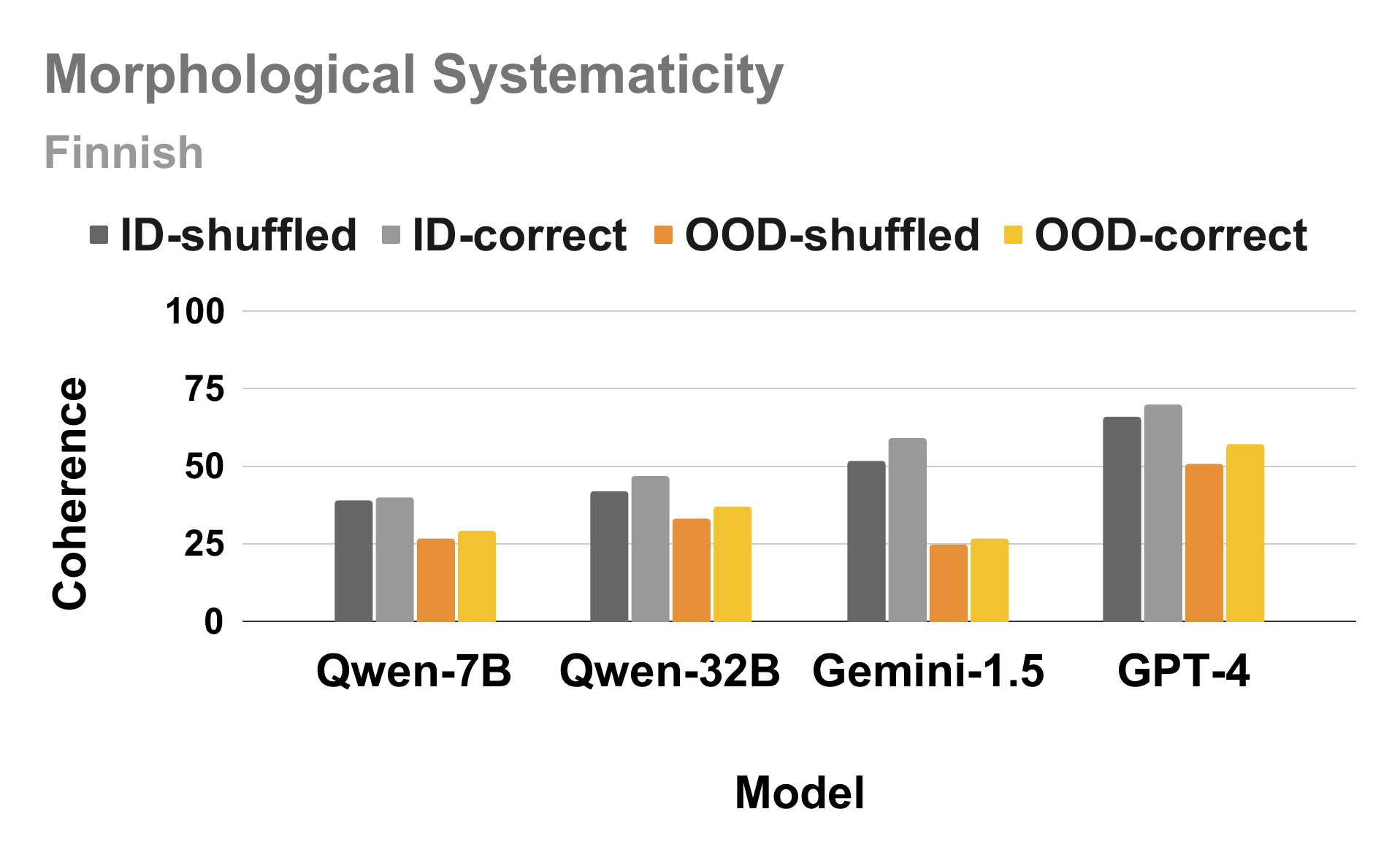}
\end{subfigure}
\caption{Morphological productivity and systematicity task results for Finnish showing the effect of the morpheme order. Detailed results are in Table \ref{tab:results-fi-en-by-order-s5}.}
\label{fig:analysis-by-order-fi-en}
\end{figure*}

%% file: results/tr/results-tr-en-default.tex
\begin{table*}[h]
\footnotesize
\centering
\scalebox{0.9}{
\centering
\begin{tabular}{lrrrrrr}
\toprule 
   \multirow{2}{*}{\textbf{Models}}
 & \multicolumn{2}{c}{{\textbf{Morph. Productivity (accuracy)}}} 
 & \multicolumn{2}{c}{{\textbf{Morph. Systematicity (macro-F1)}}}
 & \multicolumn{2}{c}{{\textbf{Morph. Systematicity (coherence)}}} \\
 
 & \multicolumn{1}{c}{ID} & \multicolumn{1}{c}{OOD} 
 & \multicolumn{1}{c}{ID} & \multicolumn{1}{c}{OOD} 
 & \multicolumn{1}{c}{ID} & \multicolumn{1}{c}{OOD} \\
 \midrule

\textbf{majority} & 0.0 / 0.0 / 0.0 & 0.0 / 0.0 / 0.0 & 41.2 / 41.2 / 41.2 & 41.2 / 41.2 / 41.2 & 0.0 / 0.0 / 0.0 & 0.0 / 0.0 / 0.0 \\
\textbf{random} & 24.6 / 25.0 / 25.0 & 24.7 / 24.6 / 24.2 & 41.8 / 41.8 / 43.5 & 43.0 / 42.3 / 42.3 & 9.1 / 9.0 / 9.4 & 8.5 / 9.6 / 9.0 \\
\midrule
\textbf{aya-23-8b} & 12.8 / 13.7 / 13.3 & 8.8 / 11.5 / 12.3 & 62.0 / 64.6 / 67.5 & 53.9 / 49.3 / 51.5 & 27.9 / 31.4 / 36.0 & 19.1 / 15.7 / 18.4 \\
\textbf{aya-23-35b} & 17.4 / 19.8 / 21.0 & 14.6 / 17.7 / 19.3 & 69.9 / 80.1 / 81.8 & 64.6 / 71.0 / 72.1 & 36.8 / 52.6 / 55.8 & 29.2 / 39.9 / 41.8 \\
\textbf{qwen-2.5-7b} & 15.0 / 14.9 / 15.8 & 13.2 / 12.9 / 12.9 & 71.1 / 73.6 / 74.6 & 65.7 / 66.8 / 66.0 & 40.5 / 44.3 / 45.1 & 33.5 / 33.9 / 33.1 \\
\textbf{qwen-2.5-32b} & 22.6 / 23.7 / 24.1 & 21.7 / 21.8 / 21.8 & 77.3 / 84.7 / 85.9 & 53.1 / 71.3 / 75.3 & 56.7 / 66.3 / 66.8 & 18.5 / 45.7 / 48.3 \\
\textbf{gemini-1.5-flash} & 28.8 / 30.5 / 30.7 & 24.9 / 25.7 / 25.1 & 60.8 / 80.8 / 85.4 & 41.4 / 52.8 / 62.1 & 32.2 / 63.6 / 70.7 & 0.4 / 19.3 / 33.3 \\
\textbf{gpt-4} & 49.0 / 52.1 / 54.2 & 36.7 / 40.5 / 43.9 & 85.5 / 90.2 / 91.6 & 61.9 / 77.7 / 78.8 & 71.4 / 76.8 / 76.6 & 33.5 / 55.9 / 51.4 \\

 \midrule
\textbf{human$^*$} & 97.1 & 95.0 & 98.8 & 99.1 & 95.7 & 97.9 \\
 
\bottomrule
\end{tabular}}
\caption{\textbf{1-shot} / \textbf{3-shot} / \textbf{5-shot} results for \textbf{Turkish} in \textbf{English} template for all examined models across tasks. $^*$Due to the cost of evaluation, our human study is only evaluated on $70$ randomly sampled instances per task and test distribution.}
\label{tab:results-tr-en-default}
\end{table*}

%% file: results/fi/results-fi-en-default.tex
\begin{table*}[h]
\footnotesize
\centering
\scalebox{0.9}{
\centering
\begin{tabular}{lrrrrrr}
\toprule 
   \multirow{2}{*}{\textbf{Models}}
 & \multicolumn{2}{c}{{\textbf{Morph. Productivity (accuracy)}}} 
 & \multicolumn{2}{c}{{\textbf{Morph. Systematicity (macro-F1)}}}
 & \multicolumn{2}{c}{{\textbf{Morph. Systematicity (coherence)}}} \\
 
 & \multicolumn{1}{c}{ID} & \multicolumn{1}{c}{OOD} 
 & \multicolumn{1}{c}{ID} & \multicolumn{1}{c}{OOD} 
 & \multicolumn{1}{c}{ID} & \multicolumn{1}{c}{OOD} \\
 \midrule

\textbf{majority} & 0.0 / 0.0 / 0.0 & 0.0 / 0.0 / 0.0 & 40.7 / 40.7 / 40.7 & 40.7 / 40.7 / 40.7 & 0.0 / 0.0 / 0.0 & 0.0 / 0.0 / 0.0 \\
\textbf{random} & 29.4 / 32.3 / 29.8 & 32.7 / 30.2 / 30.2 & 42.4 / 43.3 / 42.4 & 42.4 / 43.7 / 42.5 & 10.2 / 11.7 / 10.8 & 10.8 / 10.8 / 9.6 \\
\midrule

\textbf{qwen-2.5-7b} & 13.5 / 13.5 / 16.0 & 10.2 / 11.7 / 14.4 & 61.3 / 65.4 / 68.3 & 54.6 / 57.3 / 59.4 & 31.2 / 35.8 / 39.2 & 21.9 / 25.8 / 27.7 \\
\textbf{qwen-2.5-32b} & 22.5 / 21.9 / 22.3 & 19.2 / 19.8 / 21.3 & 52.0 / 65.9 / 69.0 & 43.6 / 54.7 / 62.2 & 19.0 / 39.8 / 42.5 & 5.2 / 22.1 / 33.1 \\
\textbf{gemini-1.5-flash} & 22.5 / 26.9 / 28.1 & 20.6 / 22.9 / 24.0 & 49.4 / 71.2 / 77.7 & 40.7 / 50.3 / 56.8 & 14.2 / 48.1 / 52.3 & 0.0 / 15.4 / 25.0 \\
\textbf{gpt-4} & 37.7 / 40.6 / 44.2 & 31.5 / 35.0 / 34.4 & 70.0 / 83.1 / 85.2 & 42.2 / 65.6 / 74.8 & 47.5 / 65.4 / 66.2 & 2.7 / 39.8 / 50.6 \\

 \midrule
\textbf{human$^*$} & 85.0 & 83.3 & 89.4 & 91.7 & 75.0 & 75.8 \\

\bottomrule
\end{tabular}}
\caption{\textbf{1-shot} / \textbf{3-shot} / \textbf{5-shot} results for \textbf{Finnish} in \textbf{English} template for all examined models across tasks. $^*$Due to the cost of evaluation, our human study is only evaluated on $60$ randomly sampled instances per task and test distribution.}
\label{tab:results-fi-en-default}
\end{table*}

%% file: results/tr/results-tr-en-by-affix/results-tr-en-by-affix_gen-acc_s1.tex
\begin{table*}[h]
\footnotesize
\centering
\scalebox{1.0}{
\centering
\begin{tabular}{lrrrrrrr}
\toprule 
   \multirow{2}{*}{\textbf{Models}}
 & \multicolumn{7}{c}{{\textbf{Number of morphemes (excl. root)}}} \\
 
 & \multicolumn{1}{c}{1} & \multicolumn{1}{c}{2} 
 & \multicolumn{1}{c}{3} & \multicolumn{1}{c}{4} 
 & \multicolumn{1}{c}{5} & \multicolumn{1}{c}{6} & \multicolumn{1}{c}{7} \\
 \midrule

\textbf{majority} & 0.0 / 0.0 & 0.0 / 0.0 & 0.0 / 0.0 & 0.0 / 0.0 & 0.0 / 0.0 & 0.0 / 0.0 & 0.0 / 0.0 \\
\textbf{random} & 100.0 / 100.0 & 48.0 / 46.7 & 20.0 / 18.0 & 3.3 / 6.7 & 0.0 / 1.3 & 0.7 / 0.0 & 0.0 / 0.0 \\
\midrule
\textbf{aya-23-8b} & 60.0 / 52.7 & 22.7 / 8.7 & 5.3 / 0.0 & 0.7 / 0.0 & 0.7 / 0.0 & 0.0 / 0.0 & 0.0 / 0.0 \\
\textbf{aya-23-35b} & 72.7 / 69.3 & 35.3 / 20.7 & 7.3 / 6.7 & 4.0 / 4.0 & 2.0 / 1.3 & 0.0 / 0.0 & 0.7 / 0.0 \\
\textbf{qwen-2.5-7b} & 63.3 / 64.0 & 26.7 / 22.0 & 13.3 / 6.0 & 0.0 / 0.7 & 0.0 / 0.0 & 0.7 / 0.0 & 0.7 / 0.0 \\
\textbf{qwen-2.5-32b} & 82.0 / 85.3 & 46.3 / 42.3 & 18.7 / 15.3 & 6.0 / 6.7 & 3.3 / 1.3 & 1.3 / 0.7 & 0.7 / 0.0 \\
\textbf{gemini-1.5-flash} & 86.7 / 80.0 & 52.7 / 44.7 & 36.0 / 30.7 & 12.0 / 10.7 & 7.3 / 7.3 & 4.7 / 0.7 & 2.0 / 0.0 \\
\textbf{gpt-4} & 95.3 / 96.7 & 80.7 / 65.3 & 62.7 / 43.3 & 43.8 / 31.3 & 27.3 / 17.6 & 19.3 / 0.7 & 13.8 / 2.1 \\

\bottomrule
\end{tabular}}
\caption{Morphological productivity 1-shot \textbf{ID} / \textbf{OOD} \textbf{accuracy} results for \textbf{Turkish} in \textbf{English} template for all examined models.}
\label{tab:results-tr-en-by-affix-gen-acc-s1}
\end{table*}

%% file: results/tr/results-tr-en-by-affix/results-tr-en-by-affix_disc-f1_s1.tex
\begin{table*}[h]
\footnotesize
\centering
\scalebox{1.0}{
\centering
\begin{tabular}{lrrrrrrr}
\toprule 
   \multirow{2}{*}{\textbf{Models}}
 & \multicolumn{7}{c}{{\textbf{Number of morphemes (excl. root)}}} \\
 
 & \multicolumn{1}{c}{1} & \multicolumn{1}{c}{2} 
 & \multicolumn{1}{c}{3} & \multicolumn{1}{c}{4} 
 & \multicolumn{1}{c}{5} & \multicolumn{1}{c}{6} & \multicolumn{1}{c}{7} \\
 \midrule

\textbf{majority} & 33.3 / 33.3 & 33.1 / 33.1 & 44.3 / 44.3 & 44.4 / 44.4 & 44.4 / 44.4 & 44.4 / 44.4 & 44.4 / 44.4 \\
\textbf{random} & 44.9 / 42.4 & 37.6 / 39.3 & 40.2 / 41.4 & 43.5 / 44.5 & 42.6 / 46.4 & 41.9 / 42.4 & 42.3 / 44.8 \\
\midrule
\textbf{aya-23-8b} & 72.9 / 54.7 & 68.0 / 48.7 & 66.1 / 50.0 & 58.8 / 54.8 & 59.0 / 56.5 & 54.3 / 55.3 & 55.1 / 57.5 \\
\textbf{aya-23-35b} & 70.4 / 60.2 & 82.7 / 70.2 & 83.1 / 73.0 & 65.0 / 58.9 & 63.5 / 65.5 & 61.5 / 61.9 & 63.2 / 62.6 \\
\textbf{qwen-2.5-7b} & 71.8 / 53.1 & 72.9 / 64.7 & 73.7 / 64.4 & 77.7 / 74.0 & 73.0 / 72.3 & 65.5 / 66.9 & 63.3 / 64.4 \\
\textbf{qwen-2.5-32b} & 65.6 / 34.2 & 57.0 / 35.1 & 75.6 / 57.6 & 87.4 / 61.5 & 86.8 / 62.1 & 84.2 / 57.3 & 84.7 / 63.9 \\
\textbf{gemini-1.5-flash} & 62.4 / 33.3 & 58.2 / 34.0 & 60.1 / 44.2 & 59.0 / 44.8 & 61.9 / 44.4 & 56.9 / 44.3 & 67.2 / 44.7 \\
\textbf{gpt-4} & 86.7 / 36.2 & 69.1 / 43.8 & 82.5 / 61.8 & 88.4 / 64.2 & 92.2 / 78.7 & 88.7 / 72.0 & 90.6 / 76.9 \\
 
\bottomrule
\end{tabular}}
\caption{Morphological systematicity 1-shot \textbf{ID} / \textbf{OOD} \textbf{macro-F1} results for \textbf{Turkish} in \textbf{English} template for all examined models.}
\label{tab:results-tr-en-by-affix-disc-f1-s1}
\end{table*}

%% file: results/tr/results-tr-en-by-affix/results-tr-en-by-affix_disc-coh_s1.tex
\begin{table*}[h]
\footnotesize
\centering
\scalebox{1.0}{
\centering
\begin{tabular}{lrrrrrrr}
\toprule 
   \multirow{2}{*}{\textbf{Models}}
 & \multicolumn{7}{c}{{\textbf{Number of morphemes (excl. root)}}} \\
 
 & \multicolumn{1}{c}{1} & \multicolumn{1}{c}{2} 
 & \multicolumn{1}{c}{3} & \multicolumn{1}{c}{4} 
 & \multicolumn{1}{c}{5} & \multicolumn{1}{c}{6} & \multicolumn{1}{c}{7} \\
 \midrule

\textbf{majority} & 0.0 / 0.0 & 0.0 / 0.0 & 0.0 / 0.0 & 0.0 / 0.0 & 0.0 / 0.0 & 0.0 / 0.0 & 0.0 / 0.0 \\
\textbf{random} & 28.0 / 25.3 & 18.7 / 22.0 & 2.0 / 1.3 & 4.7 / 2.0 & 2.7 / 2.7 & 4.0 / 2.7 & 4.0 / 3.4 \\
\midrule
\textbf{aya-23-8b} & 62.0 / 38.0 & 53.3 / 26.7 & 20.0 / 8.7 & 16.7 / 12.7 & 18.0 / 15.3 & 10.7 / 13.3 & 14.8 / 18.8 \\
\textbf{aya-23-35b} & 56.7 / 45.3 & 74.0 / 57.3 & 47.3 / 29.3 & 19.3 / 14.0 & 16.0 / 20.7 & 20.0 / 17.3 & 24.2 / 20.1 \\
\textbf{qwen-2.5-7b} & 60.7 / 36.0 & 60.7 / 49.3 & 39.3 / 29.3 & 42.0 / 34.7 & 36.7 / 34.7 & 21.3 / 24.0 & 22.8 / 26.2 \\
\textbf{qwen-2.5-32b} & 49.3 / 2.7 & 35.6 / 2.7 & 54.0 / 20.7 & 71.3 / 28.7 & 66.7 / 28.7 & 63.3 / 19.3 & 56.4 / 26.8 \\
\textbf{gemini-1.5-flash} & 44.0 / 0.0 & 38.7 / 1.3 & 26.7 / 0.0 & 24.7 / 0.7 & 30.7 / 0.0 & 22.7 / 0.0 & 38.3 / 0.7 \\
\textbf{gpt-4} & 80.0 / 4.7 & 54.0 / 16.0 & 66.7 / 30.0 & 76.0 / 34.7 & 82.7 / 58.7 & 68.7 / 41.3 & 71.8 / 49.0 \\
 
\bottomrule
\end{tabular}}
\caption{Morphological systematicity 1-shot \textbf{ID} / \textbf{OOD} \textbf{coherence} results for \textbf{Turkish} in \textbf{English} template for all examined models.}
\label{tab:results-tr-en-by-affix-disc-coh-s1}
\end{table*}

%% file: results/tr/results-tr-en-by-affix/results-tr-en-by-affix_gen-acc_s3.tex
\begin{table*}[h]
\footnotesize
\centering
\scalebox{1.0}{
\centering
\begin{tabular}{lrrrrrrr}
\toprule 
   \multirow{2}{*}{\textbf{Models}}
 & \multicolumn{7}{c}{{\textbf{Number of morphemes (excl. root)}}} \\
 
 & \multicolumn{1}{c}{1} & \multicolumn{1}{c}{2} 
 & \multicolumn{1}{c}{3} & \multicolumn{1}{c}{4} 
 & \multicolumn{1}{c}{5} & \multicolumn{1}{c}{6} & \multicolumn{1}{c}{7} \\
 \midrule

\textbf{majority} & 0.0 / 0.0 & 0.0 / 0.0 & 0.0 / 0.0 & 0.0 / 0.0 & 0.0 / 0.0 & 0.0 / 0.0 & 0.0 / 0.0 \\
\textbf{random} & 100.0 / 100.0 & 46.7 / 50.7 & 20.7 / 16.0 & 5.3 / 4.7 & 2.7 / 0.7 & 0.0 / 0.0 & 0.0 / 0.0 \\
\midrule
\textbf{aya-23-8b} & 58.0 / 64.7 & 29.3 / 13.3 & 5.3 / 2.0 & 2.0 / 0.0 & 1.3 / 0.7 & 0.0 / 0.0 & 0.0 / 0.0 \\
\textbf{aya-23-35b} & 73.3 / 84.0 & 43.3 / 29.3 & 12.0 / 6.7 & 5.3 / 4.0 & 1.3 / 0.0 & 1.3 / 0.0 & 2.0 / 0.0 \\
\textbf{qwen-2.5-7b} & 68.7 / 61.3 & 24.7 / 22.7 & 6.7 / 5.3 & 2.0 / 0.7 & 1.3 / 0.0 & 0.7 / 0.0 & 0.0 / 0.0 \\
\textbf{qwen-2.5-32b} & 84.7 / 80.7 & 45.0 / 38.9 & 21.3 / 18.7 & 6.7 / 11.3 & 4.7 / 3.3 & 0.7 / 0.0 & 2.7 / 0.0 \\
\textbf{gemini-1.5-flash} & 84.7 / 80.0 & 57.3 / 50.0 & 37.3 / 29.3 & 16.7 / 9.3 & 10.7 / 7.3 & 4.0 / 2.7 & 2.7 / 1.3 \\
\textbf{gpt-4} & 94.7 / 94.7 & 81.3 / 68.7 & 64.0 / 45.2 & 49.3 / 34.2 & 30.7 / 17.6 & 25.3 / 11.6 & 19.3 / 11.7 \\
 
\bottomrule
\end{tabular}}
\caption{Morphological productivity 3-shot \textbf{ID} / \textbf{OOD} \textbf{accuracy} results for \textbf{Turkish} in \textbf{English} template for all examined models.}
\label{tab:results-tr-en-by-affix-gen-acc-s3}
\end{table*}

%% file: results/tr/results-tr-en-by-affix/results-tr-en-by-affix_disc-f1_s3.tex
\begin{table*}[h]
\footnotesize
\centering
\scalebox{1.0}{
\centering
\begin{tabular}{lrrrrrrr}
\toprule 
   \multirow{2}{*}{\textbf{Models}}
 & \multicolumn{7}{c}{{\textbf{Number of morphemes (excl. root)}}} \\
 
 & \multicolumn{1}{c}{1} & \multicolumn{1}{c}{2} 
 & \multicolumn{1}{c}{3} & \multicolumn{1}{c}{4} 
 & \multicolumn{1}{c}{5} & \multicolumn{1}{c}{6} & \multicolumn{1}{c}{7} \\
 \midrule

\textbf{majority} & 33.3 / 33.3 & 33.1 / 33.1 & 44.3 / 44.3 & 44.4 / 44.4 & 44.4 / 44.4 & 44.4 / 44.4 & 44.4 / 44.4 \\
\textbf{random} & 40.0 / 45.3 & 40.2 / 43.1 & 44.4 / 41.5 & 40.2 / 43.4 & 43.9 / 40.3 & 43.9 / 41.8 & 39.8 / 41.0 \\
\midrule
\textbf{aya-23-8b} & 75.3 / 51.8 & 68.0 / 43.3 & 64.5 / 34.2 & 60.3 / 44.2 & 64.3 / 52.7 & 55.6 / 55.3 & 64.0 / 63.3 \\
\textbf{aya-23-35b} & 74.9 / 57.3 & 83.6 / 68.2 & 86.2 / 78.1 & 79.5 / 75.3 & 77.3 / 77.3 & 78.8 / 72.8 & 80.3 / 68.2 \\
\textbf{qwen-2.5-7b} & 60.7 / 56.9 & 75.3 / 62.9 & 76.9 / 72.2 & 78.4 / 67.9 & 74.1 / 72.3 & 74.9 / 67.3 & 74.9 / 68.4 \\
\textbf{qwen-2.5-32b} & 76.4 / 53.8 & 74.0 / 60.2 & 87.3 / 75.6 & 88.6 / 78.0 & 91.0 / 76.6 & 89.0 / 76.2 & 86.5 / 78.7 \\
\textbf{gemini-1.5-flash} & 86.0 / 45.3 & 81.6 / 50.4 & 79.1 / 55.0 & 71.2 / 53.9 & 81.2 / 55.7 & 79.1 / 55.1 & 87.6 / 54.1 \\
\textbf{gpt-4} & 89.6 / 59.1 & 81.8 / 62.9 & 92.5 / 84.9 & 94.9 / 85.7 & 92.2 / 84.7 & 88.0 / 81.8 & 92.4 / 84.7 \\

\bottomrule
\end{tabular}}
\caption{Morphological systematicity 3-shot \textbf{ID} / \textbf{OOD} \textbf{macro-F1} results for \textbf{Turkish} in \textbf{English} template for all examined models.}
\label{tab:results-tr-en-by-affix-disc-f1-s3}
\end{table*}

%% file: results/tr/results-tr-en-by-affix/results-tr-en-by-affix_disc-coh_s3.tex
\begin{table*}[h]
\footnotesize
\centering
\scalebox{1.0}{
\centering
\begin{tabular}{lrrrrrrr}
\toprule 
   \multirow{2}{*}{\textbf{Models}}
 & \multicolumn{7}{c}{{\textbf{Number of morphemes (excl. root)}}} \\
 
 & \multicolumn{1}{c}{1} & \multicolumn{1}{c}{2} 
 & \multicolumn{1}{c}{3} & \multicolumn{1}{c}{4} 
 & \multicolumn{1}{c}{5} & \multicolumn{1}{c}{6} & \multicolumn{1}{c}{7} \\
 \midrule

\textbf{majority} & 0.0 / 0.0 & 0.0 / 0.0 & 0.0 / 0.0 & 0.0 / 0.0 & 0.0 / 0.0 & 0.0 / 0.0 & 0.0 / 0.0 \\
\textbf{random} & 25.3 / 29.3 & 23.3 / 24.7 & 2.7 / 2.7 & 0.7 / 2.7 & 3.3 / 2.0 & 6.0 / 1.3 & 2.0 / 4.7 \\
\midrule
\textbf{aya-23-8b} & 64.7 / 34.0 & 52.7 / 19.3 & 22.0 / 2.0 & 22.7 / 6.7 & 24.0 / 12.0 & 13.3 / 12.7 & 20.1 / 23.5 \\
\textbf{aya-23-35b} & 64.0 / 39.3 & 75.3 / 54.7 & 56.7 / 43.3 & 41.3 / 34.7 & 39.3 / 41.3 & 43.3 / 36.0 & 48.3 / 30.2 \\
\textbf{qwen-2.5-7b} & 44.7 / 41.3 & 64.7 / 46.7 & 39.3 / 34.0 & 45.3 / 26.0 & 36.0 / 32.7 & 40.0 / 28.7 & 40.3 / 28.2 \\
\textbf{qwen-2.5-32b} & 65.3 / 32.7 & 61.1 / 40.9 & 71.3 / 49.3 & 65.3 / 52.0 & 72.0 / 51.3 & 66.7 / 46.0 & 62.4 / 47.7 \\
\textbf{gemini-1.5-flash} & 79.3 / 20.0 & 72.7 / 26.0 & 58.0 / 18.7 & 45.3 / 16.0 & 62.0 / 18.7 & 56.7 / 18.0 & 71.1 / 17.4 \\
\textbf{gpt-4} & 84.7 / 40.0 & 72.7 / 45.3 & 84.0 / 65.3 & 84.0 / 70.0 & 74.7 / 61.3 & 63.3 / 52.7 & 74.5 / 56.4 \\
 
\bottomrule
\end{tabular}}
\caption{Morphological systematicity 3-shot \textbf{ID} / \textbf{OOD} \textbf{coherence} results for \textbf{Turkish} in \textbf{English} template for all examined models.}
\label{tab:results-tr-en-by-affix-disc-coh-s3}
\end{table*}

%% file: results/tr/results-tr-en-by-affix/results-tr-en-by-affix_gen-acc_s5.tex
\begin{table*}[h]
\footnotesize
\centering
\scalebox{1.0}{
\centering
\begin{tabular}{lrrrrrrr}
\toprule 
   \multirow{2}{*}{\textbf{Models}}
 & \multicolumn{7}{c}{{\textbf{Number of morphemes (excl. root)}}} \\
 
 & \multicolumn{1}{c}{1} & \multicolumn{1}{c}{2} 
 & \multicolumn{1}{c}{3} & \multicolumn{1}{c}{4} 
 & \multicolumn{1}{c}{5} & \multicolumn{1}{c}{6} & \multicolumn{1}{c}{7} \\
 \midrule

\textbf{majority} & 0.0 / 0.0 & 0.0 / 0.0 & 0.0 / 0.0 & 0.0 / 0.0 & 0.0 / 0.0 & 0.0 / 0.0 & 0.0 / 0.0 \\
\textbf{random} & 100.0 / 100.0 & 44.0 / 48.7 & 22.7 / 15.3 & 7.3 / 4.7 & 0.7 / 0.7 & 0.0 / 0.0 & 0.0 / 0.0 \\
\midrule
\textbf{aya-23-8b} & 60.0 / 64.0 & 26.7 / 19.3 & 3.3 / 2.7 & 0.7 / 0.0 & 1.3 / 0.0 & 0.7 / 0.0 & 0.7 / 0.0 \\
\textbf{aya-23-35b} & 76.0 / 87.3 & 46.0 / 34.7 & 14.7 / 10.0 & 5.3 / 3.3 & 1.3 / 0.0 & 2.0 / 0.0 & 1.3 / 0.0 \\
\textbf{qwen-2.5-7b} & 70.7 / 65.3 & 27.3 / 16.7 & 8.0 / 6.7 & 3.3 / 0.7 & 0.7 / 0.7 & 0.0 / 0.0 & 0.7 / 0.0 \\
\textbf{qwen-2.5-32b} & 83.3 / 86.7 & 48.3 / 39.6 & 20.7 / 14.7 & 12.0 / 10.0 & 2.7 / 2.0 & 0.0 / 0.0 & 2.0 / 0.0 \\
\textbf{gemini-1.5-flash} & 91.3 / 79.3 & 56.7 / 51.3 & 39.3 / 29.3 & 12.0 / 8.7 & 10.0 / 6.0 & 2.7 / 1.3 & 2.7 / 0.0 \\
\textbf{gpt-4} & 96.0 / 96.7 & 85.3 / 72.0 & 66.0 / 55.5 & 43.7 / 37.3 & 40.0 / 23.9 & 28.0 / 11.6 & 20.6 / 10.1 \\
\midrule
\textbf{human} & 100.0 / 100.0 & 100.0 / 95.0 & 100.0 / 95.0 & 100.0 / 100.0 & 100.0 / 95.0 & 90.0 / 100.0 & 90.0 / 80.0 \\
 
\bottomrule
\end{tabular}}
\caption{Morphological productivity 5-shot \textbf{ID} / \textbf{OOD} \textbf{accuracy} results for \textbf{Turkish} in \textbf{English} template for all examined models. 1-shot and 3-shot results can be found in Tables \ref{tab:results-tr-en-by-affix-gen-acc-s1}, \ref{tab:results-tr-en-by-affix-gen-acc-s3} respectively. }
\label{tab:results-tr-en-by-affix-gen-acc-s5}
\end{table*}

%% file: results/tr/results-tr-en-by-affix/results-tr-en-by-affix_disc-f1_s5.tex
\begin{table*}[h]
\footnotesize
\centering
\scalebox{0.95}{
\centering
\begin{tabular}{lrrrrrrr}
\toprule 
   \multirow{2}{*}{\textbf{Models}}
 & \multicolumn{7}{c}{{\textbf{Number of morphemes (excl. root)}}} \\
 
 & \multicolumn{1}{c}{1} & \multicolumn{1}{c}{2} 
 & \multicolumn{1}{c}{3} & \multicolumn{1}{c}{4} 
 & \multicolumn{1}{c}{5} & \multicolumn{1}{c}{6} & \multicolumn{1}{c}{7} \\
 \midrule

\textbf{majority} & 33.3 / 33.3 & 33.1 / 33.1 & 44.3 / 44.3 & 44.4 / 44.4 & 44.4 / 44.4 & 44.4 / 44.4 & 44.4 / 44.4 \\
\textbf{random} & 42.9 / 46.4 & 43.8 / 37.6 & 43.6 / 42.2 & 40.4 / 40.4 & 44.5 / 43.9 & 43.1 / 44.0 & 46.0 / 41.4 \\
\midrule
\textbf{aya-23-8b} & 74.0 / 49.6 & 71.3 / 49.8 & 67.7 / 44.3 & 65.6 / 52.8 & 69.7 / 56.7 & 61.3 / 48.0 & 63.3 / 59.4 \\
\textbf{aya-23-35b} & 77.3 / 63.6 & 87.1 / 68.2 & 85.7 / 79.4 & 82.2 / 75.2 & 82.1 / 76.7 & 80.5 / 70.5 & 77.8 / 71.2 \\
\textbf{qwen-2.5-7b} & 66.0 / 58.0 & 76.2 / 62.7 & 76.2 / 71.1 & 78.6 / 70.4 & 74.4 / 68.6 & 76.6 / 67.1 & 74.5 / 63.8 \\
\textbf{qwen-2.5-32b} & 80.4 / 58.4 & 78.1 / 68.9 & 89.5 / 79.1 & 90.3 / 80.8 & 89.7 / 80.8 & 88.1 / 81.0 & 85.5 / 77.9 \\
\textbf{gemini-1.5-flash} & 86.9 / 53.3 & 82.9 / 55.8 & 83.4 / 66.3 & 85.5 / 60.0 & 84.8 / 60.5 & 85.5 / 66.3 & 89.2 / 72.3 \\
\textbf{gpt-4} & 92.0 / 52.7 & 90.9 / 82.9 & 94.4 / 86.1 & 93.9 / 84.7 & 90.8 / 80.6 & 89.1 / 82.9 & 90.2 / 81.4 \\

\midrule
\textbf{human} & 100.0 / 100.0 & 100.0 / 96.7 & 100.0 / 100.0 & 97.2 / 100.0 & 100.0 / 100.0 & 98.8 / 97.6 & 95.2 / 100.0 \\
 
\bottomrule
\end{tabular}}
\caption{Morphological systematicity 5-shot \textbf{ID} / \textbf{OOD} \textbf{macro-F1} results for \textbf{Turkish} in \textbf{English} template for all examined models. 1-shot and 3-shot results can be found in Tables \ref{tab:results-tr-en-by-affix-disc-f1-s1}, \ref{tab:results-tr-en-by-affix-disc-f1-s3} respectively. }
\label{tab:results-tr-en-by-affix-disc-f1-s5}
\end{table*}

%% file: results/tr/results-tr-en-by-affix/results-tr-en-by-affix_disc-coh_s5.tex
\begin{table*}[h]
\footnotesize
\centering
\scalebox{0.95}{
\centering
\begin{tabular}{lrrrrrrr}
\toprule 
   \multirow{2}{*}{\textbf{Models}}
 & \multicolumn{7}{c}{{\textbf{Number of morphemes (excl. root)}}} \\
 
 & \multicolumn{1}{c}{1} & \multicolumn{1}{c}{2} 
 & \multicolumn{1}{c}{3} & \multicolumn{1}{c}{4} 
 & \multicolumn{1}{c}{5} & \multicolumn{1}{c}{6} & \multicolumn{1}{c}{7} \\
 \midrule

\textbf{majority} & 0.0 / 0.0 & 0.0 / 0.0 & 0.0 / 0.0 & 0.0 / 0.0 & 0.0 / 0.0 & 0.0 / 0.0 & 0.0 / 0.0 \\
\textbf{random} & 24.7 / 30.0 & 26.0 / 20.0 & 2.0 / 1.3 & 4.7 / 1.3 & 2.0 / 4.7 & 4.0 / 4.0 & 2.7 / 2.0 \\
\midrule
\textbf{aya-23-8b} & 62.7 / 32.0 & 58.7 / 30.7 & 24.7 / 7.3 & 30.0 / 14.7 & 34.0 / 15.3 & 18.0 / 8.0 & 24.2 / 20.8 \\
\textbf{aya-23-35b} & 68.0 / 50.7 & 80.7 / 53.3 & 56.0 / 49.3 & 47.3 / 36.0 & 49.3 / 38.7 & 46.0 / 30.0 & 43.0 / 34.9 \\
\textbf{qwen-2.5-7b} & 51.3 / 40.7 & 66.7 / 46.0 & 34.0 / 30.7 & 43.3 / 30.7 & 37.3 / 31.3 & 39.3 / 27.3 & 43.6 / 24.8 \\
\textbf{qwen-2.5-32b} & 71.3 / 40.7 & 67.1 / 54.4 & 72.0 / 48.0 & 72.0 / 50.0 & 67.3 / 53.3 & 62.0 / 49.3 & 55.7 / 42.3 \\
\textbf{gemini-1.5-flash} & 80.7 / 31.3 & 75.3 / 34.7 & 65.3 / 34.7 & 69.3 / 26.7 & 64.7 / 27.3 & 68.0 / 34.7 & 71.8 / 43.6 \\
\textbf{gpt-4} & 88.0 / 31.3 & 86.7 / 74.7 & 86.0 / 57.3 & 78.0 / 56.7 & 68.7 / 46.0 & 64.7 / 50.7 & 64.4 / 43.0 \\
\midrule
\textbf{human} & 100.0 / 100.0 & 100.0 / 95.0 & 100.0 / 100.0 & 95.0 / 100.0 & 100.0 / 100.0 & 95.0 / 90.0 & 80.0 / 100.0 \\
 
\bottomrule
\end{tabular}}
\caption{Morphological systematicity 5-shot \textbf{ID} / \textbf{OOD} \textbf{coherence} results for \textbf{Turkish} in \textbf{English} template for all examined models. 1-shot and 3-shot results can be found in Tables \ref{tab:results-tr-en-by-affix-disc-coh-s1}, \ref{tab:results-tr-en-by-affix-disc-coh-s3} respectively. }
\label{tab:results-tr-en-by-affix-disc-coh-s5}
\end{table*}

%% file: results/fi/results-fi-en-by-affix/results-fi-en-by-affix_gen-acc_s1.tex
\begin{table*}[h]
\footnotesize
\centering
\scalebox{1.0}{
\centering
\begin{tabular}{lrrrrrr}
\toprule 
   \multirow{2}{*}{\textbf{Models}}
 & \multicolumn{6}{c}{{\textbf{Number of morphemes (excl. root)}}} \\
 
 & \multicolumn{1}{c}{1} & \multicolumn{1}{c}{2} 
 & \multicolumn{1}{c}{3} & \multicolumn{1}{c}{4} 
 & \multicolumn{1}{c}{5} & \multicolumn{1}{c}{6} \\
 \midrule

\textbf{majority} & 0.0 / 0.0 & 0.0 / 0.0 & 0.0 / 0.0 & 0.0 / 0.0 & 0.0 / 0.0 & 0.0 / 0.0 \\
\textbf{random} & 100.0 / 100.0 & 43.8 / 67.5 & 18.8 / 10.0 & 7.5 / 10.0 & 5.0 / 5.0 & 1.2 / 3.8 \\
\midrule
\textbf{qwen-2.5-7b} & 55.0 / 45.0 & 22.5 / 15.0 & 2.5 / 1.2 & 1.2 / 0.0 & 0.0 / 0.0 & 0.0 / 0.0 \\
\textbf{qwen-2.5-32b} & 72.5 / 71.2 & 41.2 / 32.5 & 18.8 / 8.8 & 2.5 / 2.5 & 0.0 / 0.0 & 0.0 / 0.0 \\
\textbf{gemini-1.5-flash} & 75.0 / 73.8 & 38.8 / 36.2 & 12.5 / 8.8 & 6.2 / 1.2 & 2.5 / 3.8 & 0.0 / 0.0 \\
\textbf{gpt-4} & 81.2 / 86.2 & 76.2 / 58.8 & 26.2 / 21.2 & 15.0 / 6.2 & 18.8 / 8.8 & 8.8 / 7.5 \\
 
\bottomrule
\end{tabular}}
\caption{Morphological productivity 1-shot \textbf{ID} / \textbf{OOD} \textbf{accuracy} results for \textbf{Finnish} in \textbf{English} template for all examined models.}
\label{tab:results-fi-en-by-affix-gen-acc-s1}
\end{table*}

%% file: results/fi/results-fi-en-by-affix/results-fi-en-by-affix_disc-f1_s1.tex
\begin{table*}[h]
\footnotesize
\centering
\scalebox{1.0}{
\centering
\begin{tabular}{lrrrrrr}
\toprule 
   \multirow{2}{*}{\textbf{Models}}
 & \multicolumn{6}{c}{{\textbf{Number of morphemes (excl. root)}}} \\
 
 & \multicolumn{1}{c}{1} & \multicolumn{1}{c}{2} 
 & \multicolumn{1}{c}{3} & \multicolumn{1}{c}{4} 
 & \multicolumn{1}{c}{5} & \multicolumn{1}{c}{6} \\
 \midrule

\textbf{majority} & 33.3 / 33.3 & 33.3 / 33.3 & 44.4 / 44.4 & 44.4 / 44.4 & 44.4 / 44.4 & 44.4 / 44.4 \\
\textbf{random} & 41.7 / 37.1 & 42.1 / 46.7 & 42.3 / 42.7 & 40.7 / 41.6 & 42.5 / 42.2 & 45.0 / 44.4 \\
\midrule
\textbf{qwen-2.5-7b} & 65.8 / 57.1 & 77.5 / 71.2 & 61.5 / 57.2 & 53.3 / 47.7 & 53.5 / 47.2 & 56.0 / 47.2 \\
\textbf{qwen-2.5-32b} & 39.2 / 34.2 & 58.8 / 41.7 & 56.6 / 49.3 & 55.4 / 46.5 & 52.2 / 45.1 & 49.7 / 45.1 \\
\textbf{gemini-1.5-flash} & 50.8 / 33.3 & 45.0 / 33.3 & 54.7 / 44.4 & 49.7 / 44.4 & 46.5 / 44.4 & 49.7 / 44.4 \\
\textbf{gpt-4} & 79.2 / 34.2 & 73.3 / 36.7 & 73.9 / 47.9 & 64.4 / 45.8 & 71.2 / 44.4 & 58.2 / 44.4 \\
 
\bottomrule
\end{tabular}}
\caption{Morphological systematicity 1-shot \textbf{ID} / \textbf{OOD} \textbf{macro-F1} results for \textbf{Finnish} in \textbf{English} template for all examined models.}
\label{tab:results-fi-en-by-affix-disc-f1-s1}
\end{table*}

%% file: results/fi/results-fi-en-by-affix/results-fi-en-by-affix_disc-coh_s1.tex
\begin{table*}[h]
\footnotesize
\centering
\scalebox{1.0}{
\centering
\begin{tabular}{lrrrrrr}
\toprule 
   \multirow{2}{*}{\textbf{Models}}
 & \multicolumn{6}{c}{{\textbf{Number of morphemes (excl. root)}}} \\
 
 & \multicolumn{1}{c}{1} & \multicolumn{1}{c}{2} 
 & \multicolumn{1}{c}{3} & \multicolumn{1}{c}{4} 
 & \multicolumn{1}{c}{5} & \multicolumn{1}{c}{6} \\
 \midrule

\textbf{majority} & 0.0 / 0.0 & 0.0 / 0.0 & 0.0 / 0.0 & 0.0 / 0.0 & 0.0 / 0.0 & 0.0 / 0.0 \\
\textbf{random} & 23.8 / 21.2 & 25.0 / 30.0 & 3.8 / 1.2 & 0.0 / 5.0 & 6.2 / 0.0 & 2.5 / 7.5 \\
\midrule
\textbf{qwen-2.5-7b} & 50.0 / 41.2 & 67.5 / 57.5 & 26.2 / 17.5 & 12.5 / 5.0 & 15.0 / 5.0 & 16.2 / 5.0 \\
\textbf{qwen-2.5-32b} & 11.2 / 3.8 & 38.8 / 12.5 & 22.5 / 8.8 & 18.8 / 3.8 & 13.8 / 1.2 & 8.8 / 1.2 \\
\textbf{gemini-1.5-flash} & 28.7 / 0.0 & 17.5 / 0.0 & 16.2 / 0.0 & 10.0 / 0.0 & 3.8 / 0.0 & 8.8 / 0.0 \\
\textbf{gpt-4} & 68.8 / 2.5 & 60.0 / 5.0 & 52.5 / 6.2 & 33.8 / 2.5 & 45.0 / 0.0 & 25.0 / 0.0 \\
 
\bottomrule
\end{tabular}}
\caption{Morphological systematicity 1-shot \textbf{ID} / \textbf{OOD} \textbf{coherence} results for \textbf{Finnish} in \textbf{English} template for all examined models.}
\label{tab:results-fi-en-by-affix-disc-coh-s1}
\end{table*}

%% file: results/fi/results-fi-en-by-affix/results-fi-en-by-affix_gen-acc_s3.tex
\begin{table*}[h]
\footnotesize
\centering
\scalebox{1.0}{
\centering
\begin{tabular}{lrrrrrr}
\toprule 
   \multirow{2}{*}{\textbf{Models}}
 & \multicolumn{6}{c}{{\textbf{Number of morphemes (excl. root)}}} \\
 
 & \multicolumn{1}{c}{1} & \multicolumn{1}{c}{2} 
 & \multicolumn{1}{c}{3} & \multicolumn{1}{c}{4} 
 & \multicolumn{1}{c}{5} & \multicolumn{1}{c}{6} \\
 \midrule

\textbf{majority} & 0.0 / 0.0 & 0.0 / 0.0 & 0.0 / 0.0 & 0.0 / 0.0 & 0.0 / 0.0 & 0.0 / 0.0 \\
\textbf{random} & 100.0 / 100.0 & 56.2 / 46.2 & 12.5 / 17.5 & 7.5 / 12.5 & 13.8 / 3.8 & 3.8 / 1.2 \\
\midrule
\textbf{qwen-2.5-7b} & 58.8 / 55.0 & 15.0 / 11.2 & 5.0 / 2.5 & 1.2 / 0.0 & 1.2 / 1.2 & 0.0 / 0.0 \\
\textbf{qwen-2.5-32b} & 70.0 / 67.5 & 41.2 / 36.2 & 12.5 / 10.0 & 3.8 / 2.5 & 1.2 / 2.5 & 2.5 / 0.0 \\
\textbf{gemini-1.5-flash} & 80.0 / 76.2 & 51.2 / 42.5 & 13.8 / 7.5 & 5.0 / 1.2 & 6.2 / 6.2 & 5.0 / 3.8 \\
\textbf{gpt-4} & 81.2 / 90.0 & 73.8 / 62.5 & 32.5 / 20.0 & 20.0 / 13.8 & 26.2 / 11.2 & 10.0 / 12.5 \\
 
\bottomrule
\end{tabular}}
\caption{Morphological productivity 3-shot \textbf{ID} / \textbf{OOD} \textbf{accuracy} results for \textbf{Finnish} in \textbf{English} template for all examined models.}
\label{tab:results-fi-en-by-affix-gen-acc-s3}
\end{table*}

%% file: results/fi/results-fi-en-by-affix/results-fi-en-by-affix_disc-f1_s3.tex
\begin{table*}[h]
\footnotesize
\centering
\scalebox{1.0}{
\centering
\begin{tabular}{lrrrrrr}
\toprule 
   \multirow{2}{*}{\textbf{Models}}
 & \multicolumn{6}{c}{{\textbf{Number of morphemes (excl. root)}}} \\
 
 & \multicolumn{1}{c}{1} & \multicolumn{1}{c}{2} 
 & \multicolumn{1}{c}{3} & \multicolumn{1}{c}{4} 
 & \multicolumn{1}{c}{5} & \multicolumn{1}{c}{6} \\
 \midrule

\textbf{majority} & 33.3 / 33.3 & 33.3 / 33.3 & 44.4 / 44.4 & 44.4 / 44.4 & 44.4 / 44.4 & 44.4 / 44.4 \\
\textbf{random} & 40.0 / 43.8 & 43.8 / 37.9 & 45.9 / 48.5 & 46.1 / 42.9 & 44.7 / 45.3 & 39.3 / 44.0 \\
\midrule
\textbf{qwen-2.5-7b} & 62.9 / 56.7 & 81.2 / 77.1 & 67.2 / 55.4 & 62.8 / 52.5 & 57.7 / 52.0 & 60.4 / 50.4 \\
\textbf{qwen-2.5-32b} & 52.9 / 37.5 & 70.0 / 57.5 & 71.6 / 59.4 & 66.3 / 54.7 & 65.7 / 56.1 & 68.7 / 62.8 \\
\textbf{gemini-1.5-flash} & 84.2 / 50.8 & 70.4 / 52.1 & 72.1 / 55.1 & 64.9 / 48.8 & 66.8 / 46.4 & 69.0 / 48.6 \\
\textbf{gpt-4} & 81.7 / 52.5 & 87.9 / 65.8 & 84.4 / 67.7 & 83.0 / 68.9 & 78.4 / 69.7 & 83.1 / 69.2 \\
 
\bottomrule
\end{tabular}}
\caption{Morphological systematicity 3-shot \textbf{ID} / \textbf{OOD} \textbf{macro-F1} results for \textbf{Finnish} in \textbf{English} template for all examined models.}
\label{tab:results-fi-en-by-affix-disc-f1-s3}
\end{table*}

%% file: results/fi/results-fi-en-by-affix/results-fi-en-by-affix_disc-coh_s3.tex
\begin{table*}[h]
\footnotesize
\centering
\scalebox{1.0}{
\centering
\begin{tabular}{lrrrrrr}
\toprule 
   \multirow{2}{*}{\textbf{Models}}
 & \multicolumn{6}{c}{{\textbf{Number of morphemes (excl. root)}}} \\
 
 & \multicolumn{1}{c}{1} & \multicolumn{1}{c}{2} 
 & \multicolumn{1}{c}{3} & \multicolumn{1}{c}{4} 
 & \multicolumn{1}{c}{5} & \multicolumn{1}{c}{6} \\
 \midrule

\textbf{majority} & 0.0 / 0.0 & 0.0 / 0.0 & 0.0 / 0.0 & 0.0 / 0.0 & 0.0 / 0.0 & 0.0 / 0.0 \\
\textbf{random} & 26.2 / 26.2 & 27.5 / 18.8 & 5.0 / 8.8 & 5.0 / 2.5 & 3.8 / 6.2 & 2.5 / 2.5 \\
\midrule
\textbf{qwen-2.5-7b} & 45.0 / 41.2 & 72.5 / 66.2 & 30.0 / 15.0 & 27.5 / 11.2 & 22.5 / 12.5 & 17.5 / 8.8 \\
\textbf{qwen-2.5-32b} & 31.2 / 8.8 & 56.2 / 36.2 & 45.0 / 25.0 & 35.0 / 16.2 & 31.2 / 17.5 & 40.0 / 28.7 \\
\textbf{gemini-1.5-flash} & 76.2 / 28.7 & 56.2 / 28.7 & 43.8 / 16.2 & 36.2 / 7.5 & 37.5 / 3.8 & 38.8 / 7.5 \\
\textbf{gpt-4} & 72.5 / 32.5 & 82.5 / 48.8 & 63.7 / 38.8 & 61.3 / 38.8 & 53.8 / 41.2 & 58.8 / 38.8 \\
 
\bottomrule
\end{tabular}}
\caption{Morphological systematicity 3-shot \textbf{ID} / \textbf{OOD} \textbf{coherence} results for \textbf{Finnish} in \textbf{English} template for all examined models.}
\label{tab:results-fi-en-by-affix-disc-coh-s3}
\end{table*}

%% file: results/fi/results-fi-en-by-affix/results-fi-en-by-affix_gen-acc_s5.tex
\begin{table*}[h]
\footnotesize
\centering
\scalebox{1.0}{
\centering
\begin{tabular}{lrrrrrr}
\toprule 
   \multirow{2}{*}{\textbf{Models}}
 & \multicolumn{6}{c}{{\textbf{Number of morphemes (excl. root)}}} \\
 
 & \multicolumn{1}{c}{1} & \multicolumn{1}{c}{2} 
 & \multicolumn{1}{c}{3} & \multicolumn{1}{c}{4} 
 & \multicolumn{1}{c}{5} & \multicolumn{1}{c}{6} \\
 \midrule

\textbf{majority} & 0.0 / 0.0 & 0.0 / 0.0 & 0.0 / 0.0 & 0.0 / 0.0 & 0.0 / 0.0 & 0.0 / 0.0 \\
\textbf{random} & 100.0 / 100.0 & 50.0 / 46.2 & 13.8 / 17.5 & 7.5 / 11.2 & 6.2 / 2.5 & 1.2 / 3.8 \\
\midrule
\textbf{qwen-2.5-7b} & 62.5 / 60.0 & 26.2 / 17.5 & 3.8 / 6.2 & 1.2 / 1.2 & 1.2 / 1.2 & 1.2 / 0.0 \\
\textbf{qwen-2.5-32b} & 72.5 / 77.5 & 43.8 / 33.8 & 10.0 / 11.2 & 3.8 / 2.5 & 1.2 / 1.2 & 2.5 / 1.2 \\
\textbf{gemini-1.5-flash} & 81.2 / 73.8 & 51.2 / 45.0 & 17.5 / 10.0 & 5.0 / 3.8 & 8.8 / 7.5 & 5.0 / 3.8 \\
\textbf{gpt-4} & 83.8 / 90.0 & 73.8 / 66.2 & 40.0 / 23.8 & 22.5 / 5.0 & 30.0 / 12.5 & 15.0 / 8.8 \\

\midrule
\textbf{human} & 90.0 / 95.0 & 90.0 / 90.0 & 100.0 / 95.0 & 80.0 / 80.0 & 60.0 / 80.0 & 90.0 / 60.0 \\

\bottomrule
\end{tabular}}
\caption{Morphological productivity 5-shot \textbf{ID} / \textbf{OOD} \textbf{accuracy} results for \textbf{Finnish} in \textbf{English} template for all examined models. 1-shot and 3-shot results can be found in Tables \ref{tab:results-fi-en-by-affix-gen-acc-s1}, \ref{tab:results-fi-en-by-affix-gen-acc-s3} respectively.}
\label{tab:results-fi-en-by-affix-gen-acc-s5}
\end{table*}

%% file: results/fi/results-fi-en-by-affix/results-fi-en-by-affix_disc-f1_s5.tex
\begin{table*}[h]
\footnotesize
\centering
\scalebox{1.0}{
\centering
\begin{tabular}{lrrrrrr}
\toprule 
   \multirow{2}{*}{\textbf{Models}}
 & \multicolumn{6}{c}{{\textbf{Number of morphemes (excl. root)}}} \\
 
 & \multicolumn{1}{c}{1} & \multicolumn{1}{c}{2} 
 & \multicolumn{1}{c}{3} & \multicolumn{1}{c}{4} 
 & \multicolumn{1}{c}{5} & \multicolumn{1}{c}{6} \\
 \midrule

\textbf{majority} & 33.3 / 33.3 & 33.3 / 33.3 & 44.4 / 44.4 & 44.4 / 44.4 & 44.4 / 44.4 & 44.4 / 44.4 \\
\textbf{random} & 43.8 / 42.5 & 40.4 / 42.9 & 43.7 / 39.8 & 41.6 / 47.3 & 39.4 / 42.4 & 45.7 / 40.0 \\
\midrule
\textbf{qwen-2.5-7b} & 65.8 / 55.4 & 86.7 / 75.4 & 65.0 / 58.1 & 66.2 / 54.2 & 62.5 / 57.7 & 63.5 / 55.9 \\
\textbf{qwen-2.5-32b} & 57.1 / 40.0 & 72.9 / 66.7 & 71.7 / 69.8 & 72.8 / 63.6 & 70.7 / 71.1 & 68.6 / 62.3 \\
\textbf{gemini-1.5-flash} & 84.2 / 49.6 & 80.8 / 65.4 & 79.6 / 62.0 & 75.4 / 52.8 & 74.9 / 53.5 & 71.4 / 57.7 \\
\textbf{gpt-4} & 84.2 / 61.3 & 89.2 / 78.8 & 84.2 / 83.8 & 88.0 / 76.7 & 83.0 / 76.5 & 82.7 / 71.6 \\

 \midrule
\textbf{human} & 73.3 / 86.7 & 93.3 / 100.0 & 93.2 / 97.6 & 95.5 / 90.9 & 91.2 / 90.5 & 89.7 / 84.4 \\

\bottomrule
\end{tabular}}
\caption{Morphological systematicity 5-shot \textbf{ID} / \textbf{OOD} \textbf{macro-F1} results for \textbf{Finnish} in \textbf{English} template for all examined models. 1-shot and 3-shot results can be found in Tables \ref{tab:results-fi-en-by-affix-disc-f1-s1}, \ref{tab:results-fi-en-by-affix-disc-f1-s3} respectively.}
\label{tab:results-fi-en-by-affix-disc-f1-s5}
\end{table*}

%% file: results/fi/results-fi-en-by-affix/results-fi-en-by-affix_disc-coh_s5.tex
\begin{table*}[h]
\footnotesize
\centering
\scalebox{1.0}{
\centering
\begin{tabular}{lrrrrrr}
\toprule 
   \multirow{2}{*}{\textbf{Models}}
 & \multicolumn{6}{c}{{\textbf{Number of morphemes (excl. root)}}} \\
 
 & \multicolumn{1}{c}{1} & \multicolumn{1}{c}{2} 
 & \multicolumn{1}{c}{3} & \multicolumn{1}{c}{4} 
 & \multicolumn{1}{c}{5} & \multicolumn{1}{c}{6} \\
 \midrule

\textbf{majority} & 0.0 / 0.0 & 0.0 / 0.0 & 0.0 / 0.0 & 0.0 / 0.0 & 0.0 / 0.0 & 0.0 / 0.0 \\
\textbf{random} & 26.2 / 26.2 & 25.0 / 23.8 & 3.8 / 0.0 & 1.2 / 5.0 & 2.5 / 1.2 & 6.2 / 1.2 \\
\midrule
\textbf{qwen-2.5-7b} & 48.8 / 36.2 & 81.2 / 65.0 & 28.7 / 16.2 & 31.2 / 15.0 & 22.5 / 18.8 & 22.5 / 15.0 \\
\textbf{qwen-2.5-32b} & 37.5 / 13.8 & 61.3 / 51.2 & 41.2 / 38.8 & 46.2 / 28.7 & 33.8 / 41.2 & 35.0 / 25.0 \\
\textbf{gemini-1.5-flash} & 76.2 / 28.7 & 71.2 / 48.8 & 47.5 / 26.2 & 43.8 / 12.5 & 42.5 / 12.5 & 32.5 / 21.2 \\
\textbf{gpt-4} & 76.2 / 45.0 & 83.8 / 68.8 & 56.2 / 61.3 & 68.8 / 48.8 & 57.5 / 40.0 & 55.0 / 40.0 \\
 \midrule
\textbf{human} & 60.0 / 80.0 & 90.0 / 100.0 & 75.0 / 90.0 & 85.0 / 75.0 & 70.0 / 60.0 & 70.0 / 50.0 \\

\bottomrule
\end{tabular}}
\caption{Morphological systematicity 5-shot \textbf{ID} / \textbf{OOD} \textbf{coherence} results for \textbf{Finnish} in \textbf{English} template for all examined models. 1-shot and 3-shot results can be found in Tables \ref{tab:results-fi-en-by-affix-disc-coh-s1}, \ref{tab:results-fi-en-by-affix-disc-coh-s3} respectively.}
\label{tab:results-fi-en-by-affix-disc-coh-s5}
\end{table*}

%% file: results/tr/results-tr-by-lang/results-tr-by-lang_s1.tex
\begin{table*}[h]
\footnotesize
\centering
\scalebox{0.9}{
\centering
\begin{tabular}{lcccccc}
\toprule 
   \multirow{2}{*}{\textbf{Models}}
 & \multicolumn{2}{c}{{\textbf{Morph. Productivity (accuracy)}}} 
 & \multicolumn{2}{c}{{\textbf{Morph. Systematicity (macro-F1)}}}
 & \multicolumn{2}{c}{{\textbf{Morph. Systematicity (coherence)}}} \\
 
 & \multicolumn{1}{c}{ID} & \multicolumn{1}{c}{OOD} 
 & \multicolumn{1}{c}{ID} & \multicolumn{1}{c}{OOD} 
 & \multicolumn{1}{c}{ID} & \multicolumn{1}{c}{OOD} \\
 \midrule

\textbf{aya-23-8b} & 12.8 / 8.4 & 8.8 / 5.3 & 62.0 / 62.4 & 53.9 / 44.3 & 27.9 / 31.9 & 19.1 / 4.5 \\
\textbf{aya-23-35b} & 17.4 / 14.6 & 14.6 / 11.7 & 69.9 / 79.6 & 64.6 / 48.9 & 36.8 / 59.2 & 29.2 / 12.2 \\
\textbf{qwen-2.5-7b} & 15.0 / 9.3 & 13.2 / 10.0 & 71.1 / 69.9 & 65.7 / 59.1 & 40.5 / 38.1 & 33.5 / 23.6 \\
\textbf{qwen-2.5-32b} & 22.6 / 22.9 & 21.7 / 20.4 & 77.3 / 78.5 & 53.1 / 44.5 & 56.7 / 60.0 & 18.5 / 5.7 \\
\textbf{gemini-1.5-flash} & 28.8 / 21.8 & 24.9 / 19.3 & 60.8 / 45.1 & 41.4 / 41.2 & 32.2 / 5.9 & 0.4 / 0.0 \\
\textbf{gpt-4} & 49.0 / 48.5 & 36.7 / 38.1 & 85.5 / 76.3 & 61.9 / 50.4 & 71.4 / 58.6 & 33.5 / 15.7 \\
 
\bottomrule
\end{tabular}}
\caption{1-shot \textbf{English} / \textbf{Turkish} template results for \textbf{Turkish} for all examined models across tasks.}
\label{tab:results-tr-by-lang-s1}
\end{table*}

%% file: results/tr/results-tr-by-lang/results-tr-by-lang_s3.tex
\begin{table*}[h]
\footnotesize
\centering
\scalebox{0.9}{
\centering
\begin{tabular}{lcccccc}
\toprule 
   \multirow{2}{*}{\textbf{Models}}
 & \multicolumn{2}{c}{{\textbf{Morph. Productivity (accuracy)}}} 
 & \multicolumn{2}{c}{{\textbf{Morph. Systematicity (macro-F1)}}}
 & \multicolumn{2}{c}{{\textbf{Morph. Systematicity (coherence)}}} \\
 
 & \multicolumn{1}{c}{ID} & \multicolumn{1}{c}{OOD} 
 & \multicolumn{1}{c}{ID} & \multicolumn{1}{c}{OOD} 
 & \multicolumn{1}{c}{ID} & \multicolumn{1}{c}{OOD} \\
 \midrule

\textbf{aya-23-8b} & 13.7 / 7.0 & 11.5 / 9.6 & 64.6 / 69.0 & 49.3 / 46.4 & 31.4 / 38.7 & 15.7 / 8.0 \\
\textbf{aya-23-35b} & 19.8 / 17.1 & 17.7 / 16.9 & 80.1 / 81.6 & 71.0 / 57.1 & 52.6 / 59.6 & 39.9 / 24.7 \\
\textbf{qwen-2.5-7b} & 14.9 / 11.2 & 12.9 / 11.4 & 73.6 / 73.7 & 66.8 / 61.4 & 44.3 / 44.8 & 33.9 / 28.2 \\
\textbf{qwen-2.5-32b} & 23.7 / 20.7 & 21.8 / 19.9 & 84.7 / 86.1 & 71.3 / 65.2 & 66.3 / 70.5 & 45.7 / 36.8 \\
\textbf{gemini-1.5-flash} & 30.5 / 23.0 & 25.7 / 20.7 & 80.8 / 56.9 & 52.8 / 41.2 & 63.6 / 25.3 & 19.3 / 0.0 \\
\textbf{gpt-4} & 52.1 / 51.5 & 40.5 / 40.6 & 90.2 / 88.6 & 77.7 / 66.3 & 76.8 / 76.3 & 55.9 / 39.9 \\
 
\bottomrule
\end{tabular}}
\caption{3-shot \textbf{English} / \textbf{Turkish} template results for \textbf{Turkish} for all examined models across tasks.}
\label{tab:results-tr-by-lang-s3}
\end{table*}

%% file: results/tr/results-tr-by-lang/results-tr-by-lang_s5.tex
\begin{table*}[h]
\footnotesize
\centering
\scalebox{0.9}{
\centering
\begin{tabular}{lcccccc}
\toprule 
   \multirow{2}{*}{\textbf{Models}}
 & \multicolumn{2}{c}{{\textbf{Morph. Productivity (accuracy)}}} 
 & \multicolumn{2}{c}{{\textbf{Morph. Systematicity (macro-F1)}}}
 & \multicolumn{2}{c}{{\textbf{Morph. Systematicity (coherence)}}} \\
 
 & \multicolumn{1}{c}{ID} & \multicolumn{1}{c}{OOD} 
 & \multicolumn{1}{c}{ID} & \multicolumn{1}{c}{OOD} 
 & \multicolumn{1}{c}{ID} & \multicolumn{1}{c}{OOD} \\
 \midrule

\textbf{aya-23-8b} & 13.3 / 8.6 & 12.3 / 8.1 & 67.5 / 62.0 & 51.5 / 47.9 & 36.0 / 30.6 & 18.4 / 15.8 \\
\textbf{aya-23-35b} & 21.0 / 17.1 & 19.3 / 18.0 & 81.8 / 81.7 & 72.1 / 64.3 & 55.8 / 57.7 & 41.8 / 30.4 \\
\textbf{qwen-2.5-7b} & 15.8 / 13.0 & 12.9 / 12.3 & 74.6 / 73.5 & 66.0 / 65.2 & 45.1 / 42.4 & 33.1 / 32.3 \\
\textbf{qwen-2.5-32b} & 24.1 / 23.3 & 21.8 / 21.9 & 85.9 / 86.9 & 75.3 / 70.7 & 66.8 / 70.5 & 48.3 / 44.2 \\
\textbf{gemini-1.5-flash} & 30.7 / 25.8 & 25.1 / 22.1 & 85.4 / 61.6 & 62.1 / 41.6 & 70.7 / 32.9 & 33.3 / 0.7 \\
\textbf{gpt-4} & 54.2 / 53.1 & 43.9 / 40.7 & 91.6 / 92.7 & 78.8 / 76.3 & 76.6 / 82.1 & 51.4 / 51.9 \\
 
\bottomrule
\end{tabular}}
\caption{5-shot \textbf{English} / \textbf{Turkish} template results for \textbf{Turkish} for all examined models across tasks. Results for 1-shot and 3-shot can be found in Tables \ref{tab:results-tr-by-lang-s1} and \ref{tab:results-tr-by-lang-s3}.}
\label{tab:results-tr-by-lang-s5}
\end{table*}

%% file: results/fi/results-fi-by-lang/results-fi-by-lang_s1.tex
\begin{table*}[h]
\footnotesize
\centering
\scalebox{0.9}{
\centering
\begin{tabular}{lcccccc}
\toprule 
   \multirow{2}{*}{\textbf{Models}}
 & \multicolumn{2}{c}{{\textbf{Morph. Productivity (accuracy)}}} 
 & \multicolumn{2}{c}{{\textbf{Morph. Systematicity (macro-F1)}}}
 & \multicolumn{2}{c}{{\textbf{Morph. Systematicity (coherence)}}} \\
 
 & \multicolumn{1}{c}{ID} & \multicolumn{1}{c}{OOD} 
 & \multicolumn{1}{c}{ID} & \multicolumn{1}{c}{OOD} 
 & \multicolumn{1}{c}{ID} & \multicolumn{1}{c}{OOD} \\
 \midrule

\textbf{qwen-2.5-7b} & 13.5 / 7.9 & 10.2 / 10.8 & 61.3 / 56.8 & 54.6 / 41.2 & 31.2 / 23.8 & 21.9 / 2.1 \\
\textbf{qwen-2.5-32b} & 22.5 / 16.9 & 19.2 / 11.9 & 52.0 / 54.5 & 43.6 / 44.4 & 19.0 / 21.9 & 5.2 / 5.8 \\
\textbf{gemini-1.5-flash} & 22.5 / 25.2 & 20.6 / 21.2 & 49.4 / 50.2 & 40.7 / 40.9 & 14.2 / 15.6 & 0.0 / 0.2 \\
\textbf{gpt-4} & 37.7 / 37.3 & 31.5 / 27.7 & 70.0 / 59.6 & 42.2 / 41.6 & 47.5 / 30.4 & 2.7 / 1.2 \\
 
\bottomrule
\end{tabular}}
\caption{1-shot \textbf{English} / \textbf{Finnish} template results for \textbf{Finnish} for all examined models across tasks.}
\label{tab:results-fi-by-lang-s1}
\end{table*}

%% file: results/fi/results-fi-by-lang/results-fi-by-lang_s3.tex
\begin{table*}[h]
\footnotesize
\centering
\scalebox{0.9}{
\centering
\begin{tabular}{lcccccc}
\toprule 
   \multirow{2}{*}{\textbf{Models}}
 & \multicolumn{2}{c}{{\textbf{Morph. Productivity (accuracy)}}} 
 & \multicolumn{2}{c}{{\textbf{Morph. Systematicity (macro-F1)}}}
 & \multicolumn{2}{c}{{\textbf{Morph. Systematicity (coherence)}}} \\
 
 & \multicolumn{1}{c}{ID} & \multicolumn{1}{c}{OOD} 
 & \multicolumn{1}{c}{ID} & \multicolumn{1}{c}{OOD} 
 & \multicolumn{1}{c}{ID} & \multicolumn{1}{c}{OOD} \\
 \midrule

\textbf{qwen-2.5-7b} & 13.5 / 10.4 & 11.7 / 10.2 & 65.4 / 65.0 & 57.3 / 55.7 & 35.8 / 32.5 & 25.8 / 22.5 \\
\textbf{qwen-2.5-32b} & 21.9 / 21.0 & 19.8 / 17.3 & 65.9 / 60.9 & 54.7 / 56.4 & 39.8 / 31.7 & 22.1 / 24.8 \\
\textbf{gemini-1.5-flash} & 26.9 / 24.6 & 22.9 / 19.2 & 71.2 / 74.7 & 50.3 / 48.0 & 48.1 / 51.7 & 15.4 / 11.9 \\
\textbf{gpt-4} & 40.6 / 40.8 & 35.0 / 32.7 & 83.1 / 72.0 & 65.6 / 45.9 & 65.4 / 51.0 & 39.8 / 8.5 \\
 
\bottomrule
\end{tabular}}
\caption{3-shot \textbf{English} / \textbf{Finnish} template results for \textbf{Finnish} for all examined models across tasks.}
\label{tab:results-fi-by-lang-s3}
\end{table*}

%% file: results/fi/results-fi-by-lang/results-fi-by-lang_s5.tex
\begin{table*}[h]
\footnotesize
\centering
\scalebox{0.9}{
\centering
\begin{tabular}{lcccccc}
\toprule 
   \multirow{2}{*}{\textbf{Models}}
 & \multicolumn{2}{c}{{\textbf{Morph. Productivity (accuracy)}}} 
 & \multicolumn{2}{c}{{\textbf{Morph. Systematicity (macro-F1)}}}
 & \multicolumn{2}{c}{{\textbf{Morph. Systematicity (coherence)}}} \\
 
 & \multicolumn{1}{c}{ID} & \multicolumn{1}{c}{OOD} 
 & \multicolumn{1}{c}{ID} & \multicolumn{1}{c}{OOD} 
 & \multicolumn{1}{c}{ID} & \multicolumn{1}{c}{OOD} \\
 \midrule

\textbf{qwen-2.5-7b} & 16.0 / 13.1 & 14.4 / 13.3 & 68.3 / 66.6 & 59.4 / 60.1 & 39.2 / 35.2 & 27.7 / 28.1 \\
\textbf{qwen-2.5-32b} & 22.3 / 20.6 & 21.3 / 20.6 & 69.0 / 66.5 & 62.2 / 59.1 & 42.5 / 39.4 & 33.1 / 29.0 \\
\textbf{gemini-1.5-flash} & 28.1 / 28.1 & 24.0 / 21.7 & 77.7 / 76.2 & 56.8 / 59.2 & 52.3 / 51.9 & 25.0 / 28.1 \\
\textbf{gpt-4} & 44.2 / 42.9 & 34.4 / 34.4 & 85.2 / 81.1 & 74.8 / 57.2 & 66.2 / 65.0 & 50.6 / 26.2 \\
 
\bottomrule
\end{tabular}}
\caption{5-shot \textbf{English} / \textbf{Finnish} template results for \textbf{Finnish} for all examined models across tasks. 1-shot and 3-shot results can be found in Tables \ref{tab:results-fi-by-lang-s1} and \ref{tab:results-fi-by-lang-s3}.}
\label{tab:results-fi-by-lang-s5}
\end{table*}

%% file: results/tr/results-tr-en-by-context/results-tr-en-by-context_s1.tex
\begin{table*}[h]
\footnotesize
\centering
\scalebox{0.9}{
\centering
\begin{tabular}{lcccccc}
\toprule 
   \multirow{2}{*}{\textbf{Models}}
 & \multicolumn{2}{c}{{\textbf{Morph. Productivity (accuracy)}}} 
 & \multicolumn{2}{c}{{\textbf{Morph. Systematicity (macro-F1)}}}
 & \multicolumn{2}{c}{{\textbf{Morph. Systematicity (coherence)}}} \\
 
 & \multicolumn{1}{c}{ID} & \multicolumn{1}{c}{OOD} 
 & \multicolumn{1}{c}{ID} & \multicolumn{1}{c}{OOD} 
 & \multicolumn{1}{c}{ID} & \multicolumn{1}{c}{OOD} \\
 \midrule

\textbf{aya-23-8b} & 12.8 / 13.1 & 8.8 / 6.3 & 62.0 / 49.8 & 53.9 / 40.3 & 27.9 / 16.8 & 19.1 / 7.5 \\
\textbf{aya-23-35b} & 17.4 / 21.0 & 14.6 / 13.1 & 69.9 / 66.3 & 64.6 / 55.8 & 36.8 / 37.3 & 29.2 / 22.0 \\
\textbf{qwen-2.5-7b} & 15.0 / 13.0 & 13.2 / 9.8 & 71.1 / 61.9 & 65.7 / 53.6 & 40.5 / 29.8 & 33.5 / 19.7 \\
\textbf{qwen-2.5-32b} & 22.6 / 23.7 & 21.7 / 19.3 & 77.3 / 44.0 & 53.1 / 41.4 & 56.7 / 4.7 & 18.5 / 0.3 \\
\textbf{gemini-1.5-flash} & 28.8 / 36.1 & 24.9 / 24.1 & 60.8 / 57.4 & 41.4 / 43.5 & 32.2 / 26.2 & 0.4 / 3.8 \\
\textbf{gpt-4} & 49.0 / 59.6 & 36.7 / 39.9 & 85.5 / 71.0 & 61.9 / 54.5 & 71.4 / 49.6 & 33.5 / 21.4 \\
 
\bottomrule
\end{tabular}}
\caption{1-shot \textbf{No context} / \textbf{With context} results for \textbf{Turkish} in \textbf{English} template for all examined models across tasks.}
\label{tab:results-tr-en-by-context-s1}
\end{table*}

%% file: results/tr/results-tr-en-by-context/results-tr-en-by-context_s3.tex
\begin{table*}[h]
\footnotesize
\centering
\scalebox{0.9}{
\centering
\begin{tabular}{lcccccc}
\toprule 
   \multirow{2}{*}{\textbf{Models}}
 & \multicolumn{2}{c}{{\textbf{Morph. Productivity (accuracy)}}} 
 & \multicolumn{2}{c}{{\textbf{Morph. Systematicity (macro-F1)}}}
 & \multicolumn{2}{c}{{\textbf{Morph. Systematicity (coherence)}}} \\
 
 & \multicolumn{1}{c}{ID} & \multicolumn{1}{c}{OOD} 
 & \multicolumn{1}{c}{ID} & \multicolumn{1}{c}{OOD} 
 & \multicolumn{1}{c}{ID} & \multicolumn{1}{c}{OOD} \\
 \midrule

\textbf{aya-23-8b} & 13.7 / 13.6 & 11.5 / 9.6 & 64.6 / 50.3 & 49.3 / 42.5 & 31.4 / 16.8 & 15.7 / 9.6 \\
\textbf{aya-23-35b} & 19.8 / 24.3 & 17.7 / 17.8 & 80.1 / 66.3 & 71.0 / 51.0 & 52.6 / 34.9 & 39.9 / 16.1 \\
\textbf{qwen-2.5-7b} & 14.9 / 13.5 & 12.9 / 9.8 & 73.6 / 63.4 & 66.8 / 59.3 & 44.3 / 30.3 & 33.9 / 25.2 \\
\textbf{qwen-2.5-32b} & 23.7 / 24.2 & 21.8 / 19.3 & 84.7 / 61.0 & 71.3 / 47.3 & 66.3 / 33.2 & 45.7 / 10.2 \\
\textbf{gemini-1.5-flash} & 30.5 / 38.4 & 25.7 / 26.5 & 80.8 / 75.6 & 52.8 / 60.1 & 63.6 / 49.7 & 19.3 / 27.2 \\
\textbf{gpt-4} & 52.1 / 59.8 & 40.5 / 45.6 & 90.2 / 85.2 & 77.7 / 67.1 & 76.8 / 71.9 & 55.9 / 41.5 \\
 
\bottomrule
\end{tabular}}
\caption{3-shot \textbf{No context} / \textbf{With context} results for \textbf{Turkish} in \textbf{English} template for all examined models across tasks.}
\label{tab:results-tr-en-by-context-s3}
\end{table*}

%% file: results/tr/results-tr-en-by-context/results-tr-en-by-context_s5.tex
\begin{table*}[h]
\footnotesize
\centering
\scalebox{0.9}{
\centering
\begin{tabular}{lcccccc}
\toprule 
   \multirow{2}{*}{\textbf{Models}}
 & \multicolumn{2}{c}{{\textbf{Morph. Productivity (accuracy)}}} 
 & \multicolumn{2}{c}{{\textbf{Morph. Systematicity (macro-F1)}}}
 & \multicolumn{2}{c}{{\textbf{Morph. Systematicity (coherence)}}} \\
 
 & \multicolumn{1}{c}{ID} & \multicolumn{1}{c}{OOD} 
 & \multicolumn{1}{c}{ID} & \multicolumn{1}{c}{OOD} 
 & \multicolumn{1}{c}{ID} & \multicolumn{1}{c}{OOD} \\
 \midrule

\textbf{aya-23-8b} & 13.3 / 14.8 & 12.3 / 9.9 & 67.5 / 51.4 & 51.5 / 42.9 & 36.0 / 19.3 & 18.4 / 10.6 \\
\textbf{aya-23-35b} & 21.0 / 26.5 & 19.3 / 18.7 & 81.8 / 66.2 & 72.1 / 47.6 & 55.8 / 35.4 & 41.8 / 14.3 \\
\textbf{qwen-2.5-7b} & 15.8 / 15.0 & 12.9 / 11.4 & 74.6 / 63.9 & 66.0 / 58.8 & 45.1 / 30.3 & 33.1 / 23.1 \\
\textbf{qwen-2.5-32b} & 24.1 / 26.2 & 21.8 / 21.4 & 85.9 / 68.4 & 75.3 / 52.0 & 66.8 / 44.8 & 48.3 / 18.1 \\
\textbf{gemini-1.5-flash} & 30.7 / 41.7 & 25.1 / 28.7 & 85.4 / 77.6 & 62.1 / 65.3 & 70.7 / 51.6 & 33.3 / 32.2 \\
\textbf{gpt-4} & 54.2 / 60.0 & 43.9 / 46.3 & 91.6 / 88.4 & 78.8 / 72.1 & 76.6 / 77.2 & 51.4 / 48.4 \\
 
\bottomrule
\end{tabular}}
\caption{5-shot \textbf{No context} / \textbf{With context} results for \textbf{Turkish} in \textbf{English} template for all examined models across tasks. 1-shot and 3-shot results can be found in Tables \ref{tab:results-tr-en-by-context-s1} and \ref{tab:results-tr-en-by-context-s3}.}
\label{tab:results-tr-en-by-context-s5}
\end{table*}

%% file: results/fi/results-fi-en-by-context/results-fi-en-by-context_s1.tex
\begin{table*}[h]
\footnotesize
\centering
\scalebox{0.9}{
\centering
\begin{tabular}{lcccccc}
\toprule 
   \multirow{2}{*}{\textbf{Models}}
 & \multicolumn{2}{c}{{\textbf{Morph. Productivity (accuracy)}}} 
 & \multicolumn{2}{c}{{\textbf{Morph. Systematicity (macro-F1)}}}
 & \multicolumn{2}{c}{{\textbf{Morph. Systematicity (coherence)}}} \\
 
 & \multicolumn{1}{c}{ID} & \multicolumn{1}{c}{OOD} 
 & \multicolumn{1}{c}{ID} & \multicolumn{1}{c}{OOD} 
 & \multicolumn{1}{c}{ID} & \multicolumn{1}{c}{OOD} \\
 \midrule

\textbf{qwen-2.5-7b} & 13.5 / 10.6 & 10.2 / 9.6 & 61.3 / 61.5 & 54.6 / 53.2 & 31.2 / 30.4 & 21.9 / 19.0 \\
\textbf{qwen-2.5-32b} & 22.5 / 22.3 & 19.2 / 17.3 & 52.0 / 43.6 & 43.6 / 40.7 & 19.0 / 4.8 & 5.2 / 0.0 \\
\textbf{gemini-1.5-flash} & 22.5 / 26.7 & 20.6 / 21.9 & 49.4 / 54.7 & 40.7 / 41.1 & 14.2 / 22.1 & 0.0 / 0.6 \\
\textbf{gpt-4} & 37.7 / 46.7 & 31.5 / 31.7 & 70.0 / 74.6 & 42.2 / 49.6 & 47.5 / 53.1 & 2.7 / 14.4 \\
 
\bottomrule
\end{tabular}}
\caption{1-shot \textbf{No context} / \textbf{With context} results for \textbf{Finnish} in \textbf{English} template for all examined models across tasks.}
\label{tab:results-fi-en-by-context-s1}
\end{table*}

%% file: results/fi/results-fi-en-by-context/results-fi-en-by-context_s3.tex
\begin{table*}[h]
\footnotesize
\centering
\scalebox{0.9}{
\centering
\begin{tabular}{lcccccc}
\toprule 
   \multirow{2}{*}{\textbf{Models}}
 & \multicolumn{2}{c}{{\textbf{Morph. Productivity (accuracy)}}} 
 & \multicolumn{2}{c}{{\textbf{Morph. Systematicity (macro-F1)}}}
 & \multicolumn{2}{c}{{\textbf{Morph. Systematicity (coherence)}}} \\
 
 & \multicolumn{1}{c}{ID} & \multicolumn{1}{c}{OOD} 
 & \multicolumn{1}{c}{ID} & \multicolumn{1}{c}{OOD} 
 & \multicolumn{1}{c}{ID} & \multicolumn{1}{c}{OOD} \\
 \midrule

\textbf{qwen-2.5-7b} & 13.5 / 11.0 & 11.7 / 10.8 & 65.4 / 63.5 & 57.3 / 55.3 & 35.8 / 34.0 & 25.8 / 21.5 \\
\textbf{qwen-2.5-32b} & 21.9 / 22.5 & 19.8 / 16.0 & 65.9 / 62.0 & 54.7 / 47.3 & 39.8 / 33.8 & 22.1 / 10.6 \\
\textbf{gemini-1.5-flash} & 26.9 / 32.3 & 22.9 / 23.3 & 71.2 / 72.3 & 50.3 / 57.3 & 48.1 / 43.3 & 15.4 / 23.3 \\
\textbf{gpt-4} & 40.6 / 52.3 & 35.0 / 33.5 & 83.1 / 83.8 & 65.6 / 65.2 & 65.4 / 67.9 & 39.8 / 39.2 \\
 
\bottomrule
\end{tabular}}
\caption{3-shot \textbf{No context} / \textbf{With context} results for \textbf{Finnish} in \textbf{English} template for all examined models across tasks.}
\label{tab:results-fi-en-by-context-s3}
\end{table*}

%% file: results/fi/results-fi-en-by-context/results-fi-en-by-context_s5.tex
\begin{table*}[h]
\footnotesize
\centering
\scalebox{0.9}{
\centering
\begin{tabular}{lcccccc}
\toprule 
   \multirow{2}{*}{\textbf{Models}}
 & \multicolumn{2}{c}{{\textbf{Morph. Productivity (accuracy)}}} 
 & \multicolumn{2}{c}{{\textbf{Morph. Systematicity (macro-F1)}}}
 & \multicolumn{2}{c}{{\textbf{Morph. Systematicity (coherence)}}} \\
 
 & \multicolumn{1}{c}{ID} & \multicolumn{1}{c}{OOD} 
 & \multicolumn{1}{c}{ID} & \multicolumn{1}{c}{OOD} 
 & \multicolumn{1}{c}{ID} & \multicolumn{1}{c}{OOD} \\
 \midrule

\textbf{qwen-2.5-7b} & 16.0 / 13.5 & 14.4 / 12.7 & 68.3 / 63.0 & 59.4 / 54.5 & 39.2 / 32.3 & 27.7 / 21.9 \\
\textbf{qwen-2.5-32b} & 22.3 / 23.8 & 21.3 / 20.0 & 69.0 / 65.5 & 62.2 / 54.5 & 42.5 / 37.9 & 33.1 / 20.6 \\
\textbf{gemini-1.5-flash} & 28.1 / 32.7 & 24.0 / 24.4 & 77.7 / 72.4 & 56.8 / 53.8 & 52.3 / 42.7 & 25.0 / 19.8 \\
\textbf{gpt-4} & 44.2 / 53.1 & 34.4 / 32.3 & 85.2 / 85.8 & 74.8 / 68.5 & 66.2 / 68.1 & 50.6 / 41.5 \\
 
\bottomrule
\end{tabular}}
\caption{5-shot \textbf{No context} / \textbf{With context} results for \textbf{Finnish} in \textbf{English} template for all examined models across tasks. 1-shot and 3-shot results can be found in Tables \ref{tab:results-fi-en-by-context-s1} and \ref{tab:results-fi-en-by-context-s3}.}
\label{tab:results-fi-en-by-context-s5}
\end{table*}

%% file: results/tr/results-tr-en-by-comp/results-tr-en-by-comp_gen-acc_s1_id.tex
\begin{table*}[h]
\footnotesize
\centering
\scalebox{1.0}{
\centering
\begin{tabular}{lrrrrrrr}
\toprule 
   \multirow{2}{*}{\textbf{Models}}
 & \multicolumn{7}{c}{{\textbf{Number of morphemes (excl. root)}}} \\
 
 & \multicolumn{1}{c}{1} & \multicolumn{1}{c}{2} 
 & \multicolumn{1}{c}{3} & \multicolumn{1}{c}{4} 
 & \multicolumn{1}{c}{5} & \multicolumn{1}{c}{6} & \multicolumn{1}{c}{7} \\
 \midrule

\textbf{gpt-4} & 95.3 / 84.0 & 80.7 / 67.8 & 62.7 / 52.7 & 43.8 / 42.1 & 27.3 / 32.0 & 19.3 / 26.5 & 13.8 / 13.9 \\

\bottomrule
\end{tabular}}
\caption{GPT-4 morphological productivity 1-shot \textbf{morphologically aligned} / \textbf{tokenizer aligned} \textbf{accuracy} results on the ID test set for \textbf{Turkish} in \textbf{English} template.}
\label{tab:results-tr-en-by-comp-gen-acc-s1}
\end{table*}

%% file: results/tr/results-tr-en-by-comp/results-tr-en-by-comp_gen-acc_s3_id.tex
\begin{table*}[h]
\footnotesize
\centering
\scalebox{1.0}{
\centering
\begin{tabular}{lrrrrrrr}
\toprule 
   \multirow{2}{*}{\textbf{Models}}
 & \multicolumn{7}{c}{{\textbf{Number of morphemes (excl. root)}}} \\
 
 & \multicolumn{1}{c}{1} & \multicolumn{1}{c}{2} 
 & \multicolumn{1}{c}{3} & \multicolumn{1}{c}{4} 
 & \multicolumn{1}{c}{5} & \multicolumn{1}{c}{6} & \multicolumn{1}{c}{7} \\
 \midrule

\textbf{gpt-4} & 94.7 / 84.7 & 81.3 / 72.5 & 64.0 / 58.0 & 49.3 / 48.2 & 30.7 / 39.5 & 25.3 / 36.8 & 19.3 / 23.8 \\

\bottomrule
\end{tabular}}
\caption{GPT-4 morphological productivity 3-shot \textbf{morphologically aligned} / \textbf{tokenizer aligned} \textbf{accuracy} results on the ID test set for \textbf{Turkish} in \textbf{English} template.}
\label{tab:results-tr-en-by-comp-gen-acc-s3}
\end{table*}

%% file: results/tr/results-tr-en-by-comp/results-tr-en-by-comp_gen-acc_s5_id.tex
\begin{table*}[h]
\footnotesize
\centering
\scalebox{1.0}{
\centering
\begin{tabular}{lrrrrrrr}
\toprule 
   \multirow{2}{*}{\textbf{Models}}
 & \multicolumn{7}{c}{{\textbf{Number of morphemes (excl. root)}}} \\
 
 & \multicolumn{1}{c}{1} & \multicolumn{1}{c}{2} 
 & \multicolumn{1}{c}{3} & \multicolumn{1}{c}{4} 
 & \multicolumn{1}{c}{5} & \multicolumn{1}{c}{6} & \multicolumn{1}{c}{7} \\
 \midrule

\textbf{gpt-4} & 96.0 / 88.0 & 85.3 / 69.8 & 66.0 / 64.0 & 43.7 / 44.8 & 40.0 / 42.2 & 28.0 / 32.4 & 20.6 / 25.7 \\

\bottomrule
\end{tabular}}
\caption{GPT-4 morphological productivity 5-shot \textbf{morphologically aligned} / \textbf{tokenizer aligned} \textbf{accuracy} results on the ID test set for \textbf{Turkish} in \textbf{English} template. 1-shot and 3-shot results can be found in Tables \ref{tab:results-tr-en-by-comp-gen-acc-s1} and \ref{tab:results-tr-en-by-comp-gen-acc-s3}.}
\label{tab:results-tr-en-by-comp-gen-acc-s5}
\end{table*}

%% file: results/tr/results-tr-en-by-cot/results-tr-en-by-cot_s505.tex
\begin{table*}[h]
\footnotesize
\centering
\scalebox{0.9}{
\centering
\begin{tabular}{lcccccc}
\toprule 
   \multirow{2}{*}{\textbf{Models}}
 & \multicolumn{2}{c}{{\textbf{Morph. Productivity (accuracy)}}} 
 & \multicolumn{2}{c}{{\textbf{Morph. Systematicity (macro-F1)}}}
 & \multicolumn{2}{c}{{\textbf{Morph. Systematicity (coherence)}}} \\
 
 & \multicolumn{1}{c}{ID} & \multicolumn{1}{c}{OOD} 
 & \multicolumn{1}{c}{ID} & \multicolumn{1}{c}{OOD} 
 & \multicolumn{1}{c}{ID} & \multicolumn{1}{c}{OOD} \\
 \midrule

\textbf{gpt-4} & 54.2 / 36.4 / 46.8 & 43.9 / 31.2 / 45.8 & 91.6 / 85.1 / 88.8 & 78.8 / 70.3 / 83.0 & 76.6 / 63.8 / 72.5 & 51.4 / 38.1 / 61.1 \\
 
\bottomrule
\end{tabular}}
\caption{GPT-4 \textbf{5-shot} / \textbf{0-shot-cot} / \textbf{5-shot-cot} results for \textbf{Turkish} in \textbf{English} template across tasks.}
\label{tab:results-tr-en-by-cot-s505}
\end{table*}

%% file: results/tr/results-tr-en-by-order/results-tr-en-by-order_s1.tex
\begin{table*}[h]
\footnotesize
\centering
\scalebox{0.9}{
\centering
\begin{tabular}{lcccccc}
\toprule 
   \multirow{2}{*}{\textbf{Models}}
 & \multicolumn{2}{c}{{\textbf{Morph. Productivity (accuracy)}}} 
 & \multicolumn{2}{c}{{\textbf{Morph. Systematicity (macro-F1)}}}
 & \multicolumn{2}{c}{{\textbf{Morph. Systematicity (coherence)}}} \\
 
 & \multicolumn{1}{c}{ID} & \multicolumn{1}{c}{OOD} 
 & \multicolumn{1}{c}{ID} & \multicolumn{1}{c}{OOD} 
 & \multicolumn{1}{c}{ID} & \multicolumn{1}{c}{OOD} \\
 \midrule

\textbf{aya-23-8b} & 12.8 / 14.2 & 8.8 / 9.4 & 62.0 / 62.0 & 53.9 / 53.5 & 27.9 / 28.0 & 19.1 / 19.4 \\
\textbf{aya-23-35b} & 17.4 / 22.3 & 14.6 / 17.9 & 69.9 / 69.4 & 64.6 / 64.3 & 36.8 / 36.2 & 29.2 / 28.2 \\
\textbf{qwen-2.5-7b} & 15.0 / 16.7 & 13.2 / 14.9 & 71.1 / 73.1 & 65.7 / 67.8 & 40.5 / 42.7 & 33.5 / 36.5 \\
\textbf{qwen-2.5-32b} & 22.6 / 30.0 & 21.7 / 29.2 & 77.3 / 80.4 & 53.1 / 58.1 & 56.7 / 61.0 & 18.5 / 25.9 \\
\textbf{gemini-1.5-flash} & 28.8 / 37.2 & 24.9 / 32.1 & 60.8 / 65.4 & 41.4 / 41.5 & 32.2 / 39.6 & 0.4 / 0.7 \\
\textbf{gpt-4} & 49.0 / 63.0 & 36.7 / 54.5 & 85.5 / 88.6 & 61.9 / 68.3 & 71.4 / 77.5 & 33.5 / 43.7 \\
 
\bottomrule
\end{tabular}}
\caption{1-shot \textbf{Shuffled} / \textbf{Correct} morpheme order results for \textbf{Turkish} in \textbf{English} template for all examined models across tasks.}
\label{tab:results-tr-en-by-order-s1}
\end{table*}

%% file: results/tr/results-tr-en-by-order/results-tr-en-by-order_s3.tex
\begin{table*}[h]
\footnotesize
\centering
\scalebox{0.9}{
\centering
\begin{tabular}{lcccccc}
\toprule 
   \multirow{2}{*}{\textbf{Models}}
 & \multicolumn{2}{c}{{\textbf{Morph. Productivity (accuracy)}}} 
 & \multicolumn{2}{c}{{\textbf{Morph. Systematicity (macro-F1)}}}
 & \multicolumn{2}{c}{{\textbf{Morph. Systematicity (coherence)}}} \\
 
 & \multicolumn{1}{c}{ID} & \multicolumn{1}{c}{OOD} 
 & \multicolumn{1}{c}{ID} & \multicolumn{1}{c}{OOD} 
 & \multicolumn{1}{c}{ID} & \multicolumn{1}{c}{OOD} \\
 \midrule

\textbf{aya-23-8b} & 13.7 / 14.9 & 11.5 / 12.5 & 64.6 / 63.8 & 49.3 / 49.1 & 31.4 / 30.7 & 15.7 / 16.4 \\
\textbf{aya-23-35b} & 19.8 / 25.7 & 17.7 / 23.7 & 80.1 / 80.2 & 71.0 / 73.0 & 52.6 / 52.2 & 39.9 / 41.9 \\
\textbf{qwen-2.5-7b} & 14.9 / 17.8 & 12.9 / 15.5 & 73.6 / 76.1 & 66.8 / 69.4 & 44.3 / 47.9 & 33.9 / 38.0 \\
\textbf{qwen-2.5-32b} & 23.7 / 30.7 & 21.8 / 32.1 & 84.7 / 86.9 & 71.3 / 75.7 & 66.3 / 70.3 & 45.7 / 53.0 \\
\textbf{gemini-1.5-flash} & 30.5 / 39.6 & 25.7 / 33.4 & 80.8 / 85.4 & 52.8 / 58.0 & 63.6 / 71.6 & 19.3 / 27.4 \\
\textbf{gpt-4} & 52.1 / 70.3 & 40.5 / 63.5 & 90.2 / 92.9 & 77.7 / 81.0 & 76.8 / 82.0 & 55.9 / 59.6 \\
 
\bottomrule
\end{tabular}}
\caption{3-shot \textbf{Shuffled} / \textbf{Correct} morpheme order results for \textbf{Turkish} in \textbf{English} template for all examined models across tasks.}
\label{tab:results-tr-en-by-order-s3}
\end{table*}

%% file: results/tr/results-tr-en-by-order/results-tr-en-by-order_s5.tex
\begin{table*}[h]
\footnotesize
\centering
\scalebox{0.9}{
\centering
\begin{tabular}{lcccccc}
\toprule 
   \multirow{2}{*}{\textbf{Models}}
 & \multicolumn{2}{c}{{\textbf{Morph. Productivity (accuracy)}}} 
 & \multicolumn{2}{c}{{\textbf{Morph. Systematicity (macro-F1)}}}
 & \multicolumn{2}{c}{{\textbf{Morph. Systematicity (coherence)}}} \\
 
 & \multicolumn{1}{c}{ID} & \multicolumn{1}{c}{OOD} 
 & \multicolumn{1}{c}{ID} & \multicolumn{1}{c}{OOD} 
 & \multicolumn{1}{c}{ID} & \multicolumn{1}{c}{OOD} \\
 \midrule

\textbf{aya-23-8b} & 13.3 / 15.0 & 12.3 / 13.0 & 67.5 / 66.4 & 51.5 / 51.8 & 36.0 / 34.7 & 18.4 / 18.1 \\
\textbf{aya-23-35b} & 21.0 / 28.5 & 19.3 / 25.8 & 81.8 / 81.3 & 72.1 / 72.3 & 55.8 / 55.2 & 41.8 / 42.3 \\
\textbf{qwen-2.5-7b} & 15.8 / 19.2 & 12.9 / 16.9 & 74.6 / 76.4 & 66.0 / 68.5 & 45.1 / 47.0 & 33.1 / 35.7 \\
\textbf{qwen-2.5-32b} & 24.1 / 32.3 & 21.8 / 36.4 & 85.9 / 87.5 & 75.3 / 78.2 & 66.8 / 69.6 & 48.3 / 52.2 \\
\textbf{gemini-1.5-flash} & 30.7 / 43.3 & 25.1 / 35.1 & 85.4 / 88.6 & 62.1 / 66.1 & 70.7 / 74.5 & 33.3 / 39.5 \\
\textbf{gpt-4} & 54.2 / 73.0 & 43.9 / 66.7 & 91.6 / 93.7 & 78.8 / 82.6 & 76.6 / 82.2 & 51.4 / 58.3 \\
 
\bottomrule
\end{tabular}}
\caption{5-shot \textbf{Shuffled} / \textbf{Correct} morpheme order results for \textbf{Turkish} in \textbf{English} template for all examined models across tasks. 1-shot and 3-shot results can be found in Tables \ref{tab:results-tr-en-by-order-s1} and \ref{tab:results-tr-en-by-order-s3}.}
\label{tab:results-tr-en-by-order-s5}
\end{table*}

%% file: results/fi/results-fi-en-by-order/results-fi-en-by-order_s1.tex
\begin{table*}[h]
\footnotesize
\centering
\scalebox{0.9}{
\centering
\begin{tabular}{lcccccc}
\toprule 
   \multirow{2}{*}{\textbf{Models}}
 & \multicolumn{2}{c}{{\textbf{Morph. Productivity (accuracy)}}} 
 & \multicolumn{2}{c}{{\textbf{Morph. Systematicity (macro-F1)}}}
 & \multicolumn{2}{c}{{\textbf{Morph. Systematicity (coherence)}}} \\
 
 & \multicolumn{1}{c}{ID} & \multicolumn{1}{c}{OOD} 
 & \multicolumn{1}{c}{ID} & \multicolumn{1}{c}{OOD} 
 & \multicolumn{1}{c}{ID} & \multicolumn{1}{c}{OOD} \\
 \midrule

\textbf{qwen-2.5-7b} & 13.5 / 15.6 & 10.2 / 12.3 & 61.3 / 62.9 & 54.6 / 55.2 & 31.2 / 34.2 & 21.9 / 22.5 \\
\textbf{qwen-2.5-32b} & 22.5 / 24.8 & 19.2 / 23.1 & 52.0 / 53.1 & 43.6 / 44.0 & 19.0 / 20.4 & 5.2 / 5.8 \\
\textbf{gemini-1.5-flash} & 22.5 / 26.7 & 20.6 / 25.6 & 49.4 / 51.2 & 40.7 / 40.9 & 14.2 / 17.5 & 0.0 / 0.2 \\
\textbf{gpt-4} & 37.7 / 46.0 & 31.5 / 39.6 & 70.0 / 70.7 & 42.2 / 43.7 & 47.5 / 48.3 & 2.7 / 5.0 \\
 
\bottomrule
\end{tabular}}
\caption{1-shot \textbf{Shuffled} / \textbf{Correct} morpheme order results for \textbf{Finnish} in \textbf{English} template for all examined models across tasks.}
\label{tab:results-fi-en-by-order-s1}
\end{table*}

%% file: results/fi/results-fi-en-by-order/results-fi-en-by-order_s3.tex
\begin{table*}[h]
\footnotesize
\centering
\scalebox{0.9}{
\centering
\begin{tabular}{lcccccc}
\toprule 
   \multirow{2}{*}{\textbf{Models}}
 & \multicolumn{2}{c}{{\textbf{Morph. Productivity (accuracy)}}} 
 & \multicolumn{2}{c}{{\textbf{Morph. Systematicity (macro-F1)}}}
 & \multicolumn{2}{c}{{\textbf{Morph. Systematicity (coherence)}}} \\
 
 & \multicolumn{1}{c}{ID} & \multicolumn{1}{c}{OOD} 
 & \multicolumn{1}{c}{ID} & \multicolumn{1}{c}{OOD} 
 & \multicolumn{1}{c}{ID} & \multicolumn{1}{c}{OOD} \\
 \midrule

\textbf{qwen-2.5-7b} & 13.5 / 16.5 & 11.7 / 15.2 & 65.4 / 67.5 & 57.3 / 59.6 & 35.8 / 39.8 & 25.8 / 29.0 \\
\textbf{qwen-2.5-32b} & 21.9 / 25.2 & 19.8 / 25.0 & 65.9 / 66.6 & 54.7 / 55.4 & 39.8 / 41.5 & 22.1 / 23.1 \\
\textbf{gemini-1.5-flash} & 26.9 / 35.0 & 22.9 / 28.7 & 71.2 / 73.2 & 50.3 / 51.1 & 48.1 / 50.6 & 15.4 / 16.5 \\
\textbf{gpt-4} & 40.6 / 56.0 & 35.0 / 50.4 & 83.1 / 83.2 & 65.6 / 67.4 & 65.4 / 67.3 & 39.8 / 42.7 \\
 
\bottomrule
\end{tabular}}
\caption{3-shot \textbf{Shuffled} / \textbf{Correct} morpheme order results for \textbf{Finnish} in \textbf{English} template for all examined models across tasks.}
\label{tab:results-fi-en-by-order-s3}
\end{table*}

%% file: results/fi/results-fi-en-by-order/results-fi-en-by-order_s5.tex
\begin{table*}[h]
\footnotesize
\centering
\scalebox{0.9}{
\centering
\begin{tabular}{lcccccc}
\toprule 
   \multirow{2}{*}{\textbf{Models}}
 & \multicolumn{2}{c}{{\textbf{Morph. Productivity (accuracy)}}} 
 & \multicolumn{2}{c}{{\textbf{Morph. Systematicity (macro-F1)}}}
 & \multicolumn{2}{c}{{\textbf{Morph. Systematicity (coherence)}}} \\
 
 & \multicolumn{1}{c}{ID} & \multicolumn{1}{c}{OOD} 
 & \multicolumn{1}{c}{ID} & \multicolumn{1}{c}{OOD} 
 & \multicolumn{1}{c}{ID} & \multicolumn{1}{c}{OOD} \\
 \midrule

\textbf{qwen-2.5-7b} & 16.0 / 18.8 & 14.4 / 17.1 & 68.3 / 68.7 & 59.4 / 61.6 & 39.2 / 40.2 & 27.7 / 29.4 \\
\textbf{qwen-2.5-32b} & 22.3 / 27.9 & 21.3 / 31.7 & 69.0 / 71.3 & 62.2 / 64.7 & 42.5 / 46.7 & 33.1 / 37.3 \\
\textbf{gemini-1.5-flash} & 28.1 / 35.0 & 24.0 / 30.6 & 77.7 / 79.9 & 56.8 / 57.7 & 52.3 / 58.5 & 25.0 / 26.9 \\
\textbf{gpt-4} & 44.2 / 59.6 & 34.4 / 52.1 & 85.2 / 86.9 & 74.8 / 78.4 & 66.2 / 70.0 & 50.6 / 57.1 \\
 
\bottomrule
\end{tabular}}
\caption{5-shot \textbf{Shuffled} / \textbf{Correct} morpheme order results for \textbf{Finnish} in \textbf{English} template for all examined models across tasks. 1-shot and 3-shot results can be found in Tables \ref{tab:results-fi-en-by-order-s1} and \ref{tab:results-fi-en-by-order-s3}.}
\label{tab:results-fi-en-by-order-s5}
\end{table*}

%% file: results/tr/results-tr-en-by-negsel/results-tr-en-by-negsel_s1.tex
\begin{table*}[h]
\footnotesize
\centering
\scalebox{0.9}{
\centering
\begin{tabular}{lcccc}
\toprule 
   \multirow{2}{*}{\textbf{Models}}
 & \multicolumn{2}{c}{{\textbf{Morph. Systematicity (macro-F1)}}}
 & \multicolumn{2}{c}{{\textbf{Morph. Systematicity (coherence)}}} \\
 
 & \multicolumn{1}{c}{ID} & \multicolumn{1}{c}{OOD} 
 & \multicolumn{1}{c}{ID} & \multicolumn{1}{c}{OOD} \\
 \midrule

\textbf{aya-23-8b} & 74.8 / 62.0 / 60.2 & 59.7 / 53.9 / 51.1 & 46.4 / 27.9 / 25.4 & 26.9 / 19.1 / 16.1 \\
\textbf{aya-23-35b} & 83.5 / 69.9 / 67.9 & 74.2 / 64.6 / 61.6 & 59.5 / 36.8 / 33.3 & 43.3 / 29.2 / 26.0 \\
\textbf{qwen-2.5-7b} & 81.7 / 71.1 / 68.6 & 75.1 / 65.7 / 63.7 & 64.1 / 40.5 / 36.3 & 54.1 / 33.5 / 29.4 \\
\textbf{qwen-2.5-32b} & 79.7 / 77.3 / 76.6 & 54.0 / 53.1 / 52.8 & 65.5 / 56.7 / 53.8 & 22.0 / 18.5 / 17.7 \\
\textbf{gemini-1.5-flash} & 62.3 / 60.8 / 60.6 & 41.5 / 41.4 / 41.3 & 35.0 / 32.2 / 31.6 & 0.6 / 0.4 / 0.4 \\
\textbf{gpt-4} & 85.8 / 85.5 / 83.2 & 63.3 / 61.9 / 60.9 & 75.7 / 71.4 / 66.9 & 38.5 / 33.5 / 30.4 \\
 
\bottomrule
\end{tabular}}
\caption{1-shot \textbf{Random} / \textbf{Language-agnostic} / \textbf{Language-specific} negative sample selection results for \textbf{Turkish} in \textbf{English} template for all examined models across tasks.}
\label{tab:results-tr-en-by-negsel-s1}
\end{table*}

%% file: results/tr/results-tr-en-by-negsel/results-tr-en-by-negsel_s3.tex
\begin{table*}[h]
\footnotesize
\centering
\scalebox{0.9}{
\centering
\begin{tabular}{lcccc}
\toprule 
   \multirow{2}{*}{\textbf{Models}}
 & \multicolumn{2}{c}{{\textbf{Morph. Systematicity (macro-F1)}}}
 & \multicolumn{2}{c}{{\textbf{Morph. Systematicity (coherence)}}} \\
 
 & \multicolumn{1}{c}{ID} & \multicolumn{1}{c}{OOD} 
 & \multicolumn{1}{c}{ID} & \multicolumn{1}{c}{OOD} \\
 \midrule

\textbf{aya-23-8b} & 74.4 / 64.6 / 61.9 & 53.0 / 49.3 / 47.1 & 45.6 / 31.4 / 28.1 & 21.5 / 15.7 / 13.4 \\
\textbf{aya-23-35b} & 88.2 / 80.1 / 78.8 & 80.4 / 71.0 / 71.0 & 72.9 / 52.6 / 48.9 & 60.5 / 39.9 / 38.1 \\
\textbf{qwen-2.5-7b} & 81.2 / 73.6 / 71.6 & 75.3 / 66.8 / 65.6 & 63.8 / 44.3 / 38.8 & 53.5 / 33.9 / 32.2 \\
\textbf{qwen-2.5-32b} & 88.3 / 84.7 / 83.4 & 74.5 / 71.3 / 69.8 & 78.3 / 66.3 / 63.6 & 55.4 / 45.7 / 42.4 \\
\textbf{gemini-1.5-flash} & 80.2 / 80.8 / 79.9 & 51.7 / 52.8 / 51.5 & 65.3 / 63.6 / 60.8 & 17.8 / 19.3 / 17.3 \\
\textbf{gpt-4} & 93.7 / 90.2 / 89.1 & 82.4 / 77.7 / 74.1 & 88.1 / 76.8 / 72.7 & 66.8 / 55.9 / 45.8 \\
 
\bottomrule
\end{tabular}}
\caption{3-shot \textbf{Random} / \textbf{Language-agnostic} / \textbf{Language-specific} negative sample selection results for \textbf{Turkish} in \textbf{English} template for all examined models across tasks.}
\label{tab:results-tr-en-by-negsel-s3}
\end{table*}

%% file: results/tr/results-tr-en-by-negsel/results-tr-en-by-negsel_s5.tex
\begin{table*}[h]
\footnotesize
\centering
\scalebox{0.9}{
\centering
\begin{tabular}{lcccc}
\toprule 
   \multirow{2}{*}{\textbf{Models}}
 & \multicolumn{2}{c}{{\textbf{Morph. Systematicity (macro-F1)}}}
 & \multicolumn{2}{c}{{\textbf{Morph. Systematicity (coherence)}}} \\
 
 & \multicolumn{1}{c}{ID} & \multicolumn{1}{c}{OOD} 
 & \multicolumn{1}{c}{ID} & \multicolumn{1}{c}{OOD} \\
 \midrule

\textbf{aya-23-8b} & 77.4 / 67.5 / 66.6 & 58.1 / 51.5 / 50.6 & 51.0 / 36.0 / 33.5 & 25.5 / 18.4 / 17.3 \\
\textbf{aya-23-35b} & 89.6 / 81.8 / 80.5 & 80.7 / 72.1 / 70.9 & 75.8 / 55.8 / 51.8 & 60.5 / 41.8 / 38.8 \\
\textbf{qwen-2.5-7b} & 83.1 / 74.6 / 72.1 & 76.0 / 66.0 / 65.0 & 64.3 / 45.1 / 39.7 & 52.9 / 33.1 / 30.3 \\
\textbf{qwen-2.5-32b} & 90.1 / 85.9 / 83.9 & 81.3 / 75.3 / 73.6 & 80.3 / 66.8 / 61.8 & 64.7 / 48.3 / 45.8 \\
\textbf{gemini-1.5-flash} & 87.8 / 85.4 / 85.3 & 61.6 / 62.1 / 59.4 & 78.3 / 70.7 / 68.6 & 34.5 / 33.3 / 29.1 \\
\textbf{gpt-4} & 95.4 / 91.6 / 89.4 & 83.7 / 78.8 / 72.1 & 89.2 / 76.6 / 70.8 & 64.7 / 51.4 / 38.7 \\
 
\bottomrule
\end{tabular}}
\caption{5-shot \textbf{Random} / \textbf{Language-agnostic} / \textbf{Language-specific} negative sample selection results for \textbf{Turkish} in \textbf{English} template for all examined models across tasks.}
\label{tab:results-tr-en-by-negsel-s5}
\end{table*}

%% file: 111_prompts.tex
\section{Prompts}
\label{sec:prompts}
This section lists the instruction prompts for all tasks and language templates. We present examples in one-shot setting, templates for different shots are the same with more examples. For the English language template, we provide examples in Turkish, the templates are the same for Finnish with examples in Finnish.

\subsection{Templates in English}

\begin{highlight}
\textbf{Productivity task prompt [ID root]}\\
You are given a word root and a list of affixes (separated by comma) in Turkish and your task is to generate a grammatically correct word from this root using all the given affixes. You are allowed to use only the given affixes and each affix only once. Answer with only the generated word.\\
Example 1:\\
Word root: bulaş\\
Affixes: ma, sa, tır, ydı, k\\
Answer: bulaştırmasaydık\\

Example 2:\\
Word root: bekle\\
Affixes: me, di, z, n, e\\
Answer:
\end{highlight}

\begin{highlight}
\textbf{Productivity task prompt [OOD root]}\\
You are given a novel word root with its definition and a list of affixes (separated by comma) in Turkish and your task is to generate a grammatically correct word from this root using all the given affixes. You are allowed to use only the given affixes and each affix only once. Answer with only the generated word.\\
Example 1:\\
Word root: lıdış\\
Definition: lıdış means karış in Turkish.\\
Affixes: sa, ydı, k, ma\\
Answer: lıdışmasaydık\\

Example 2:\\
Word root: ihek\\
Definition: ihek means emek in Turkish.\\
Affixes: in, imiz, ler, çi\\
Answer:
\end{highlight}

\begin{highlightgreen}
\textbf{Systematicity task prompt [ID root]}\\
You are given a word root, a list of affixes (separated by comma) and a word in Turkish that is derived from the given word root using the given affixes. Your task is to determine whether the derived word is grammatically correct. Answer only with Yes or No.\\
Example 1:\\
Word root: küçük\\
Affixes: ümüz, lüğ, den\\
Derived word: küçüklüğümüzden\\
Answer: Yes\\

Example 2:\\
Word root: evren\\
Affixes: sel, e, liğ\\
Derived word: evreneselliğ\\
Answer:
\end{highlightgreen}

\begin{highlightgreen}
\textbf{Systematicity task prompt [OOD root]}\\
You are given a novel word root with its definition, a list of affixes (separated by comma) and a word in Turkish that is derived from the given word root using the given affixes. Your task is to determine whether the derived word is grammatically correct. Answer only with Yes or No.\\
Example 1:\\
Word root: eneşilvöte\\
Definition: eneşilvöte means üniversite in Turkish.\\
Affixes: niz, yse, de\\
Derived word: eneşilvötedeyseniz\\
Answer: Yes\\

Example 2:\\
Word root: yivek\\
Definition: yivek means yürek in Turkish.\\
Affixes: den, ler, iniz\\
Derived word: yiveklerdeniniz\\
Answer:
\end{highlightgreen}

\begin{highlight}
\textbf{Productivity task prompt [ID root] (with context)}\\
You are given a word root, a list of affixes (separated by comma) and a sentence with a blank (\_\_\_) in Turkish and your task is to fill in the blank by generating a grammatically correct word from this root using all the given affixes. You are allowed to use only the given affixes and each affix only once. Answer with only the generated word.\\
Example 1:\\
Word root: kal\\
Affixes: an, lar\\
Sentence: giden geminin yokluğuna bir türlü inandıramaz kendilerini limanda \_\_\_\\
Answer: kalanlar\\

Example 2:\\
Word root: kurtar\\
Affixes: ecek, abil\\
Sentence: göç ettikten sonra diğer hemşerileri gibi mal, mülk peşinde olsa belki annesini parasızlıktan \_\_\_ belki de kızı bir fabrika köşesinde çalışmak zorunda kalmayıp daha uzun yaşayabilecekti\\
Answer:   
\end{highlight}

\begin{highlightgreen}
\textbf{Systematicity task prompt [ID root] (with context)}
You are given a word root, a list of affixes (separated by comma), a sentence with a blank (\_\_\_) and a word in Turkish that is derived from the given word root using the given affixes. Your task is to determine whether the derived word is the correct option to fill in the blank. Answer only with Yes or No.\\

Example 1:\\
Word root: küçük\\
Affixes: ümüz, den, lüğ\\
Sentence: \_\_\_ kalma bir oyuna dönüştürdük hayatımızı\\
Derived word: küçüklüğümüzden\\
Answer: Yes\\

Example 2:\\
Word root: akıl\\
Affixes: lan, ız, acağ\\
Sentence: bir şeyler yaşadıktan sonra mı \_\_\_ hep\\
Derived word: akılacağızlan\\
Answer:
\end{highlightgreen}

\begin{highlight}
\textbf{Productivity task prompt [ID root] (CoT)}\\
You are given a word root and a list of affixes (separated by comma) in Turkish. Your task is to construct a grammatically correct word by appending the given affixes to the root. Use each affix exactly once. After forming a word, list each affix used in the construction of that word to verify adherence to the rules. Check the following: Ensure no affix is used more than once, confirm that all provided affixes are used, verify that no extra affixes outside the provided list are included. Think step by step and then provide your final answer within the tags <Answer>correctword</Answer>.\\

Example 1:\\
Word root: kuru\\
Affixes: t, muş\\
Answer: First, let's append the affixes to the root "kuru" in a grammatically correct order:\\
...<explaining the correct order of morphemes>...

Example 2:\\
Word root: mana\\
Affixes: sız, dır\\
Answer:
\end{highlight}

\begin{highlight}
\textbf{Productivity task prompt [OOD root] (CoT)}\\
You are provided with a novel word root with its definition, and a list of affixes (separated by comma) in Turkish. Your task is to construct a grammatically correct word by appending the given affixes to the root. Use each affix exactly once. After forming a word, list each affix used in the construction of that word to verify adherence to the rules. Check the following: Ensure no affix is used more than once, confirm that all provided affixes are used, verify that no extra affixes outside the provided list are included. Think step by step and then provide your final answer within the tags <Answer>correctword</Answer>.\\

Example 1:\\
Word root: doru\\
Definition: doru means kuru in Turkish.\\
Affixes: t, muş\\
Answer: ...<explanation>...

Example 2:\\
Word root: çokan\\
Definition: çokan means yalan in Turkish.\\
Affixes: la, lar\\
Answer:
\end{highlight}

\begin{highlightgreen}
\textbf{Systematicity task prompt [ID root] (CoT)}\\
You are given a word root, a list of affixes (separated by comma) and a word in Turkish that is derived from the given word root using the given affixes. Your task is to determine whether the derived word is grammatically correct. First, analyze how the affixes interact with the word root. Then, assess the order in which the affixes are applied and verify that this order adheres to the language's rules. Think step by step and then provide your final answer within the tags <Answer>Yes/No</Answer>.\\
Example 1:\\
Word root: kuru\\
Affixes: t, muş\\
Derived word: kurutmuş\\
Answer: To analyze the derived word "kurutmuş," we need to look at the affixes and how they interact with the word root "kuru."\\
...<explaining the correct order of morphemes>...

Example 2:\\
Word root: etki\\
Affixes: yici, le\\
Derived word: etkileyici\\
Answer:
\end{highlightgreen}

\begin{highlightgreen}
\textbf{Systematicity task prompt [OOD root] (CoT)}\\
You are given a novel word root with its definition, a list of affixes (separated by comma) and a word in Turkish that is derived from the given word root using the given affixes. Your task is to determine whether the derived word is grammatically correct. First, analyze how the affixes interact with the word root. Then, assess the order in which the affixes are applied and verify that this order adheres to the language's rules. Think step by step and then provide your final answer within the tags <Answer>Yes/No</Answer>.\\
Example 1:\\
Word root: doru\\
Definition: doru means kuru in Turkish.\\
Affixes: t, muş\\
Derived word: dorutmuş\\
Answer: ...<explain the correct order of morphemes based on the definition>...

Example 2:\\
Word root: imli\\
Definition: imli means etki in Turkish.\\
Affixes: yici, le\\
Derived word: imlileyici\\
Answer:
\end{highlightgreen}


\begin{highlight}
\textbf{Productivity task prompt [ID root][paraphrased]}\\
You are provided with a word root and a set of affixes (comma-separated) in {language}. Your task is to create a grammatically correct word using this root and all the provided affixes. You must use only the given affixes, and each affix can be used only once. Respond with the final word only.\\
Example 1:\\
Word root: bulaş\\
Affixes: ma, sa, tır, ydı, k\\
Answer: bulaştırmasaydık\\

Example 2:\\
Word root: bekle\\
Affixes: me, di, z, n, e\\
Answer:
\end{highlight}

\begin{highlight}
\textbf{Productivity task prompt [OOD root][paraphrased]}\\
You are given a new word root along with its definition, and a set of affixes (comma-separated) in {language}. Assuming that the new word root is a valid {language} word, your task is to form a grammatically correct word using this root and all the provided affixes. You must use only the given affixes, and each one can be used just once. Provide only the generated word as your answer.\\

Example 1:\\
Word root: lıdış\\
Definition: lıdış means karış in Turkish.\\
Affixes: sa, ydı, k, ma\\
Answer: lıdışmasaydık\\

Example 2:\\
Word root: ihek\\
Definition: ihek means emek in Turkish.\\
Affixes: in, imiz, ler, çi\\
Answer:
\end{highlight}

\begin{highlightgreen}
\textbf{Systematicity task prompt [ID root][paraphrased]}\\
You are provided with a word root, a set of affixes (comma-separated), and a word in {language} that is derived from the given root using the provided affixes. Your task is to verify whether the derived word is grammatically correct. Respond with only Yes or No.\\

Example 1:\\
Word root: küçük\\
Affixes: ümüz, lüğ, den\\
Derived word: küçüklüğümüzden\\
Answer: Yes\\

Example 2:\\
Word root: evren\\
Affixes: sel, e, liğ\\
Derived word: evreneselliğ\\
Answer:
\end{highlightgreen}

\begin{highlightgreen}
\textbf{Systematicity task prompt [OOD root][paraphrased]}\\
You are provided with a new word root along with its definition, a set of affixes (comma-separated), and a word in {language} that is derived from the given root using the provided affixes. Assuming that the new word root is a valid {language} word, your task is to verify whether the derived word is grammatically correct. Respond with only Yes or No.\\

Example 1:\\
Word root: eneşilvöte\\
Definition: eneşilvöte means üniversite in Turkish.\\
Affixes: niz, yse, de\\
Derived word: eneşilvötedeyseniz\\
Answer: Yes\\

Example 2:\\
Word root: yivek\\
Definition: yivek means yürek in Turkish.\\
Affixes: den, ler, iniz\\
Derived word: yiveklerdeniniz\\
Answer:
\end{highlightgreen}

\subsection{Templates in Turkish}

\begin{highlight}
\textbf{Productivity task prompt [ID root]}\\
Size Türkçe bir kök ve bir ek listesi (virgülle ayrılmış) verilecek ve sizden bu kökten verilen tüm ekleri kullanarak dilbilgisel olarak doğru bir kelime üretmeniz istenecek. Sadece verilen ekleri kullanabilirsiniz ve her bir ek sadece bir kez kullanılabilir. Sadece üretilen kelimeyi çıktı olarak verin.\\

Örnek 1:\\
Kök: küçük\\
Ekler: ümüz, lüğ, den\\
Cevap: küçüklüğümüzden\\

Örnek 2:\\
Kök: sevgi\\
Ekler: in, li, m\\
Cevap:
\end{highlight}

\begin{highlight}
\textbf{Productivity task prompt [OOD root]}\\
Size Türkçe yeni bir kök, onun tanımlaması ve bir ek listesi (virgülle ayrılmış) verilecek ve sizden bu kökten verilen tüm ekleri kullanarak dilbilgisel olarak doğru bir kelime üretmeniz istenecek. Sadece verilen ekleri kullanabilirsiniz ve her bir ek sadece bir kez kullanılabilir. Sadece üretilen kelimeyi çıktı olarak verin.\\

Örnek 1:\\
Kök: nıtal\\
Tanım: nıtal Türkçe kal anlamına gelir.\\
Ekler: lar, an\\
Cevap: nıtalanlar\\

Örnek 2:\\
Kök: rarcu\\
Tanım: rarcu Türkçe vurgu anlamına gelir.\\
Ekler: la, mış\\
Cevap:
\end{highlight}

\begin{highlightgreen}
\textbf{Systematicity task prompt [ID root]}\\
Size Türkçe bir kök, bir ek listesi (virgülle ayrılmış) ve bu ekleri kullanarak türetilmiş bir kelime verilecek. Sizden bu kelimenin dilbilgisel olarak doğru olup olmadığını belirlemeniz istenecek. Sadece Evet veya Hayır ile cevap verin.\\

Örnek 1:\\
Kök: küçük\\
Ekler: ümüz, lüğ, den\\
Türetilmiş kelime: küçüklüğümüzden\\
Cevap: Evet\\

Örnek 2:\\
Kök: sahip\\
Ekler: iniz, diğ, len\\
Türetilmiş kelime: sahipdiğinizlen\\
Cevap:
\end{highlightgreen}

\begin{highlightgreen}
\textbf{Systematicity task prompt [OOD root]}\\
Size Türkçe yeni bir kök, onun tanımlaması, bir ek listesi (virgülle ayrılmış) ve bu ekleri kullanarak türetilmiş bir kelime verilecek. Sizden bu kelimenin dilbilgisel olarak doğru olup olmadığını belirlemeniz istenecek. Sadece Evet veya Hayır ile cevap verin.\\

Örnek 1:\\
Kök: yivük\\
Tanım: yivük Türkçe küçük anlamına gelir.\\
Ekler: den, lüğ, ümüz\\
Türetilmiş kelime: yivüklüğümüzden\\
Cevap: Evet\\

Örnek 2:\\
Kök: minlek\\
Tanım: minlek Türkçe gerçek anlamına gelir.\\
Ekler: leş, di, me\\
Türetilmiş kelime: minlekleşmedi\\
Cevap:
\end{highlightgreen}

\begin{highlight}
\textbf{Productivity task prompt [ID root] (with context)}\\
Size Türkçe bir kök, bir ek listesi (virgülle ayrılmış) ve boşluklu (\_\_\_) bir cümle verilecek ve sizden boşluğu doldurmak için bu kökten verilen tüm ekleri kullanarak dilbilgisel olarak doğru bir kelime üretmeniz istenecek. Sadece verilen ekleri kullanabilirsiniz ve her bir ek sadece bir kez kullanılabilir. Sadece üretilen kelimeyi çıktı olarak verin.\\

Örnek 1:\\
Kök: küçük\\
Ekler: den, ümüz, lüğ\\
Cümle: \_\_\_ kalma bir oyuna dönüştürdük hayatımızı\\
Cevap: küçüklüğümüzden\\

Örnek 2:\\
Kök: ilkokul\\
Ekler: da, m, ydı\\
Cümle: Ilk kez onun bir şiirini okuyabilme fırsatı bulduğumda, henüz daha \_\_\_ ve bu kadar farklı bir tarzla karşılaşmak beni oldukça heyecanlandırmıştı\\
Cevap:
\end{highlight}

\begin{highlightgreen}
\textbf{Systematicity task prompt [ID root] (with context)}\\
Size Türkçe bir kök, bir ek listesi (virgülle ayrılmış), boşluklu (\_\_\_) bir cümle ve bu ekleri kullanarak türetilmiş bir kelime verilecek. Sizden boşluğu doldurmak için bu kelimenin dilbilgisel olarak doğru olup olmadığını belirlemeniz istenecek. Sadece Evet veya Hayır ile cevap verin.\\

Örnek 1:\\
Kök: karış\\
Ekler: ma, sa, k, ydı\\
Cümle: gerçek şu ki anlayamadığımız şeylere mucize deyip \_\_\_, bugünlere belki de hiç ulaşamayacaktık\\
Türetilmiş kelime: karışmasaydık\\
Cevap: Evet\\

Örnek 2:\\
Kök: sanat\\
Ekler: ı, çı, lar, ndan\\
Cümle: tüm bu deneyimlerime ev sahipliği yapan ülke ise dünyanın en ünlü ve en çok beğenilen \_\_\_ biri olan van gogh’un doğup büyüdüğü hollanda’dan başka bir yer değil\\
Türetilmiş kelime: sanatçılarndanı\\
Cevap:
\end{highlightgreen}

\subsection{Templates in Finnish}

\begin{highlight}
\textbf{Productivity task prompt [ID root]}\\
Sinulle annetaan sanan sananvartalo ja luettelo pilkulla erotettuja päätteitä kielellä suomi. Tehtäväsi on luoda tästä juuresta kieliopillisesti oikea sana käyttämällä kaikkia annettuja päätteitä. Voit käyttää vain annettuja päätteitä ja kutakin päätettä vain kerran. Vastaa vain luodulla sanalla.\\

Esimerkki 1:\\
Sananvartalo: markiise\\
Päätteet: j, a\\
Vastaus: markiiseja\\

Esimerkki 2:\\
Sananvartalo: kasvattamis\\
Päätteet: si, ta\\
Vastaus:
\end{highlight}

\begin{highlight}
\textbf{Productivity task prompt [OOD root]}\\
Sinulle annetaan uusi sananvartalo, sen määritelmä sekä pilkulla eroteltu luettelo päätteitä kielellä suomi. Tehtäväsi on luoda juuresta kieliopillisesti oikea sana käyttämällä kaikkia annettuja päätteitä. Käyttä vain annettuja päätteitä ja kutakin päätettä vain kerran. Vastaa vain luodulla sanalla.\\

Esimerkki 1:\\
Sananvartalo: seloks\\
Määritelmä: seloks tarkoittaa petoks kielellä suomi.\\
Päätteet: ne, en, i\\
Vastaus: seloksineen\\

Esimerkki 2:\\
Sananvartalo: osivma\\
Määritelmä: osivma tarkoittaa ohitta kielellä suomi.\\
Päätteet: han, ko, a\\
Vastaus:
\end{highlight}

\begin{highlightgreen}
\textbf{Systematicity task prompt [ID root]}\\
Sinulle annetaan sananvartalo, pilkulla eroteltu luettelo päätteistä sekä annettuja päätteitä käyttämällä vartalosta johdettu sana kielellä suomi. Tehtäväsi on selvittää, onko johdettu sana kieliopillisesti oikein. Vastaa vain Kyllä tai Ei.\\

Esimerkki 1:\\
Sananvartalo: palauttaminen\\
Päätteet: n, mi, elee\\
Johdettu sana: mieleenpalauttaminen\\
Vastaus: Kyllä\\

Esimerkki 2:\\
Sananvartalo: näkyv\\
Päätteet: imp, in, i\\
Johdettu sana: näkyvimpiin\\
Vastaus:
\end{highlightgreen}

\begin{highlightgreen}
\textbf{Systematicity task prompt [OOD root]}\\
Sinulle annetaan uusi sananvartalo, sen määritelmä sekä pilkulla eroteltu luettelo päätteistä sekä uusi sana kielellä suomi, joka on johdettu annetusta sananvartalosta annettujen päätteiden avulla. Tehtäväsi on selvittää, onko johdettu sana kieliopillisesti oikein. Vastaa vain Kyllä tai Ei.\\

Esimerkki 1:\\
Sananvartalo: sätletjimsä\\
Määritelmä: sätletjimsä tarkoittaa järjestelmä kielellä suomi.\\
Päätteet: laadu, hallinta, n, n\\
Johdettu sana: laadunhallintasätletjimsän\\
Vastaus: Kyllä\\

Esimerkki 2:\\
Sananvartalo: olanajke\\
Määritelmä: olanajke tarkoittaa olosuhte kielellä suomi.\\
Päätteet: i, kuvaus, an, lta\\
Johdettu sana: kuvausolanajkeanilta\\
Vastaus:
\end{highlightgreen}

\begin{highlight}
\textbf{Productivity task prompt [ID root] (with context)}\\
Allaolevassa lauseessa (kirjoitettu kielellä suomi) on tyhjä kohta (\_\_\_) joka tulee täyttää kieliopillisesti oikealla sanalla. Alla on myös sananvartalo sekä pilkulla eroteltu luettelo päätteistä. Tehtäväsi on käyttää vartaloa sekä päätteitä ja johtaa niistä kieliopillisesti oikein taivutetu sana joka sopii tyhjään kohtaan lausessaa asiayhteys/konteksti huomioonottaen. Käytä jokaista päätettä vain kerran. Vastaa vain generoidulla sanalla, älä sano mitään muuta.\\

Esimerkki 1:\\
Sananvartalo: markiise\\
Päätteet: a, j\\
Lause: \_\_\_ saatavana yksivärisinä, raidallisina ja voit myös valita haluatko markiisisi veivi- vai sähkökäyttöisenä.\\
Vastaus: markiiseja\\

Esimerkki 2:\\
Sananvartalo: suhteutet\\
Päätteet: na, tu\\
Lause: \_\_\_ väkilukuun, suomessa on enemmän metsää kuin missään muussa euroopan maassa.\\
Vastaus:
\end{highlight}

\begin{highlightgreen}
\textbf{Systematicity task prompt [ID root] (with context)}\\
Allaolevassa lauseessa on tyhjä kohta (\_\_\_) joka tulee täyttää kieliopillisesti oikealla sanalla. Alla on myös sananvartalo, pilkulla eroteltu luettelo päätteistä sekä niitä käyttäen annetusta vartalosta johdettu sana kielellä suomi. Tehtäväsi on päätellä, onko johdettu sana kieliopillisesti oikein, jos sen asettaa lauseen tyhjään kohtaan eli onko sana kieliopillisesti oikein taivutetu asiayhteys/konteksti huomioonottaen. Vastaa joko Kyllä tai Ei.\\

Esimerkki 1:\\
Sananvartalo: petoks\\
Päätteet: ne, en, i\\
Lause: hän paljasti koko korruptoituneen järjestelmän \_\_\_.\\
Johdettu sana: petoksineen\\
Vastaus: Kyllä

Esimerkki 2:\\
Sananvartalo: kannatta\\
Päätteet: isi, han, ko\\
Lause: \_\_\_ minun opiskella suomea?\\
Johdettu sana: kannattakoisihan\\
Vastaus:
\end{highlightgreen}

%% file: results/tr/results-tr-en-by-decoding/results-tr-en-by-temp.tex
\begin{table*}[h]
\footnotesize
\centering
\scalebox{0.9}{
\centering
\begin{tabular}{lcccccc}
\toprule 
   \multirow{2}{*}{\textbf{Models}}
 & \multicolumn{2}{c}{{\textbf{Morph. Productivity (accuracy)}}} 
 & \multicolumn{2}{c}{{\textbf{Morph. Systematicity (macro-F1)}}}
 & \multicolumn{2}{c}{{\textbf{Morph. Systematicity (coherence)}}} \\
 
 & \multicolumn{1}{c}{ID} & \multicolumn{1}{c}{OOD} 
 & \multicolumn{1}{c}{ID} & \multicolumn{1}{c}{OOD} 
 & \multicolumn{1}{c}{ID} & \multicolumn{1}{c}{OOD} \\
 \midrule

\textbf{gpt-4 (temp=0)$^*$} & 54.0 & 44.0 & 92.0 & 79.0 & 77.0 & 51.0 \\
\textbf{gpt-4 (temp=0.3)} & 53.0 & 43.0 & 92.0 & 79.0 & 80.0 & 53.0 \\
\textbf{gpt-4 (temp=0.5)} & 55.0 & 43.0 & 92.0 & 80.0 & 80.0 & 53.0 \\
\textbf{gpt-4 (temp=0.7)} & 53.0 & 43.0 & 92.0 & 80.0 & 80.0 & 53.0 \\
\textbf{gpt-4 (temp=0.9)} & 53.0 & 42.0 & 92.0 & 79.0 & 80.0 & 51.0 \\

\bottomrule
\end{tabular}}
\caption{5-shot results for \textbf{Turkish} in \textbf{English} template for GPT-4 across tasks and different temperature values. $^*$Corresponds to default decoding setting for main results.}
\label{tab:results-tr-en-by-decoding-temp}
\end{table*}

%% file: results/tr/results-tr-en-by-decoding/results-tr-en-by-topp.tex
\begin{table*}[h]
\footnotesize
\centering
\scalebox{0.9}{
\centering
\begin{tabular}{lcccccc}
\toprule 
   \multirow{2}{*}{\textbf{Models}}
 & \multicolumn{2}{c}{{\textbf{Morph. Productivity (accuracy)}}} 
 & \multicolumn{2}{c}{{\textbf{Morph. Systematicity (macro-F1)}}}
 & \multicolumn{2}{c}{{\textbf{Morph. Systematicity (coherence)}}} \\
 
 & \multicolumn{1}{c}{ID} & \multicolumn{1}{c}{OOD} 
 & \multicolumn{1}{c}{ID} & \multicolumn{1}{c}{OOD} 
 & \multicolumn{1}{c}{ID} & \multicolumn{1}{c}{OOD} \\
 \midrule

\textbf{gpt-4 (top\_p=1)$^*$} & 54.0 & 44.0 & 92.0 & 79.0 & 77.0 & 51.0 \\
\textbf{gpt-4 (top\_p=0.95)} & 53.0 & 43.0 & 92.0 & 78.0 & 79.0 & 51.0 \\
\textbf{gpt-4 (top\_p=0.9)} & 54.0 & 42.0 & 92.0 & 78.0 & 80.0 & 52.0 \\

\bottomrule
\end{tabular}}
\caption{5-shot results for \textbf{Turkish} in \textbf{English} template for GPT-4 across tasks and different top\_p values. $^*$Corresponds to default decoding setting for main results.}
\label{tab:results-tr-en-by-decoding-topp}
\end{table*}

%% file: results/fi/results-fi-en-by-decoding/results-fi-en-by-temp.tex
\begin{table*}[h]
\footnotesize
\centering
\scalebox{0.9}{
\centering
\begin{tabular}{lcccccc}
\toprule 
   \multirow{2}{*}{\textbf{Models}}
 & \multicolumn{2}{c}{{\textbf{Morph. Productivity (accuracy)}}} 
 & \multicolumn{2}{c}{{\textbf{Morph. Systematicity (macro-F1)}}}
 & \multicolumn{2}{c}{{\textbf{Morph. Systematicity (coherence)}}} \\
 
 & \multicolumn{1}{c}{ID} & \multicolumn{1}{c}{OOD} 
 & \multicolumn{1}{c}{ID} & \multicolumn{1}{c}{OOD} 
 & \multicolumn{1}{c}{ID} & \multicolumn{1}{c}{OOD} \\
 \midrule

\textbf{gpt-4 (temp=0)$^*$} & 44.0 & 34.0 & 85.0 & 75.0 & 66.0 & 51.0 \\
\textbf{gpt-4 (temp=0.3)} & 45.0 & 36.0 & 85.0 & 74.0 & 64.0 & 48.0 \\
\textbf{gpt-4 (temp=0.5)} & 45.0 & 34.0 & 85.0 & 73.0 & 64.0 & 45.0 \\
\textbf{gpt-4 (temp=0.7)} & 44.0 & 36.0 & 86.0 & 73.0 & 65.0 & 46.0 \\
\textbf{gpt-4 (temp=0.9)} & 44.0 & 33.0 & 84.0 & 71.0 & 63.0 & 42.0 \\

\bottomrule
\end{tabular}}
\caption{5-shot results for \textbf{Finnish} in \textbf{English} template for GPT-4 across tasks and different temperature values. $^*$Corresponds to default decoding setting for main results.}
\label{tab:results-fi-en-by-decoding-temp}
\end{table*}

%% file: results/fi/results-fi-en-by-decoding/results-fi-en-by-topp.tex
\begin{table*}[h]
\footnotesize
\centering
\scalebox{0.9}{
\centering
\begin{tabular}{lcccccc}
\toprule 
   \multirow{2}{*}{\textbf{Models}}
 & \multicolumn{2}{c}{{\textbf{Morph. Productivity (accuracy)}}} 
 & \multicolumn{2}{c}{{\textbf{Morph. Systematicity (macro-F1)}}}
 & \multicolumn{2}{c}{{\textbf{Morph. Systematicity (coherence)}}} \\
 
 & \multicolumn{1}{c}{ID} & \multicolumn{1}{c}{OOD} 
 & \multicolumn{1}{c}{ID} & \multicolumn{1}{c}{OOD} 
 & \multicolumn{1}{c}{ID} & \multicolumn{1}{c}{OOD} \\
 \midrule

\textbf{gpt-4 (top\_p=1)$^*$} & 44.0 & 34.0 & 85.0 & 75.0 & 66.0 & 51.0 \\
\textbf{gpt-4 (top\_p=0.95)} & 43.0 & 34.0 & 86.0 & 73.0 & 65.0 & 46.0 \\
\textbf{gpt-4 (top\_p=0.9)} & 43.0 & 34.0 & 85.0 & 79.0 & 64.0 & 44.0 \\

\bottomrule
\end{tabular}}
\caption{5-shot results for \textbf{Finnish} in \textbf{English} template for GPT-4 across tasks and different top\_p values. $^*$Corresponds to default decoding setting for main results.}
\label{tab:results-fi-en-by-decoding-topp}
\end{table*}

%% file: results/tr/results-tr-en-by-prompt.tex
\begin{table*}[h]
\footnotesize
\centering
\scalebox{0.9}{
\centering
\begin{tabular}{lcccccc}
\toprule 
   \multirow{2}{*}{\textbf{Models}}
 & \multicolumn{2}{c}{{\textbf{Morph. Productivity (accuracy)}}} 
 & \multicolumn{2}{c}{{\textbf{Morph. Systematicity (macro-F1)}}}
 & \multicolumn{2}{c}{{\textbf{Morph. Systematicity (coherence)}}} \\
 
 & \multicolumn{1}{c}{ID} & \multicolumn{1}{c}{OOD} 
 & \multicolumn{1}{c}{ID} & \multicolumn{1}{c}{OOD} 
 & \multicolumn{1}{c}{ID} & \multicolumn{1}{c}{OOD} \\
 \midrule

\textbf{gpt-4 (original)$^*$} & 54.0 & 44.0 & 92.0 & 79.0 & 77.0 & 51.0 \\
\textbf{gpt-4 (paraphrased)} & 56.0 & 46.0 & 93.0 & 80.0 & 81.0 & 54.0 \\

\bottomrule
\end{tabular}}
\caption{5-shot results for \textbf{Turkish} in \textbf{English} template for GPT-4 across tasks and different prompt instructions. $^*$Corresponds to default prompt instructions for main results.}
\label{tab:results-tr-en-by-prompt}
\end{table*}

%% file: results/fi/results-fi-en-by-prompt.tex
\begin{table*}[h]
\footnotesize
\centering
\scalebox{0.9}{
\centering
\begin{tabular}{lcccccc}
\toprule 
   \multirow{2}{*}{\textbf{Models}}
 & \multicolumn{2}{c}{{\textbf{Morph. Productivity (accuracy)}}} 
 & \multicolumn{2}{c}{{\textbf{Morph. Systematicity (macro-F1)}}}
 & \multicolumn{2}{c}{{\textbf{Morph. Systematicity (coherence)}}} \\
 
 & \multicolumn{1}{c}{ID} & \multicolumn{1}{c}{OOD} 
 & \multicolumn{1}{c}{ID} & \multicolumn{1}{c}{OOD} 
 & \multicolumn{1}{c}{ID} & \multicolumn{1}{c}{OOD} \\
 \midrule

\textbf{gpt-4 (original)$^*$} & 44.0 & 34.0 & 85.0 & 75.0 & 66.0 & 51.0 \\
\textbf{gpt-4 (paraphrased)} & 46.0 & 37.0 & 84.0 & 73.0 & 62.0 & 45.0 \\

\bottomrule
\end{tabular}}
\caption{5-shot results for \textbf{Finnish} in \textbf{English} template for GPT-4 across tasks and different prompt instructions. $^*$Corresponds to default prompt instructions for main results.}
\label{tab:results-fi-en-by-prompt}
\end{table*}